\documentclass[3p, times, 10pt, twocolumn]{elsarticle}

\usepackage{amssymb}
\usepackage{amsmath}
\usepackage{booktabs}
\usepackage{array}
\usepackage{tikz}
\usepackage{float}
\usepackage{xurl}
\newcolumntype{C}[1]{>{\centering\arraybackslash}p{#1}}
\newcommand{\myimage}[3]{%
  \begin{minipage}[c]{\linewidth}
    \centering
    \begin{tikzpicture}
        \node[inner sep=0] (main) {\includegraphics[width=\linewidth]{#1}};
        \node[anchor=south east, inner sep=2pt] at (main.south east) {%
            \includegraphics[width=0.35\linewidth, height=0.35\linewidth, keepaspectratio=true, trim=#3, clip]{#2}%
        };
    \end{tikzpicture}
  \end{minipage}%
}

\journal{Pattern Recognition}

\begin{document}

\begin{frontmatter}



\title{Rethinking Monocular Depth Embedding for Generalized Stereo Matching}

\author[1]{Libo Lin}
\author[1]{Shuangli Du\corref{cor1}}
\author[1]{Minghua Zhao}
\author[1]{Zhenzhen You}
\author[2]{Shun Lv}
\author[2]{Yiguang Liu}

\affiliation[1]{organization={Shaanxi Key Laboratory for Network Computing and Security Technology, Xi'an University of Technology},
            addressline={No. 5, Jinhua South Road},
            city={Xi'an},
            postcode={710048},
            state={Shaanxi},
            country={China}}

\affiliation[2]{organization={College of Computer Science, Sichuan University},
            addressline={},
            city={Chengdu},
            postcode={610065},
            state={Sichuan},
            country={China}}

\cortext[cor1]{Corresponding author. Email: dusl@xaut.edu.cn}

\begin{abstract}
Generally, monocular methods capture rich contextual priors but lack geometric precision, whereas stereo methods are geometrically accurate yet struggle in textureless and occluded regions. Several approaches attempt to combine their strengths to enhance the generalization of stereo matching (SM) by aligning monocular depth with stereo information. However, establishing a stable and generalizable alignment is challenging, and unreliable monocular cues can substantially degrade performance. This paper rethinks monocular depth embedding. First, to prevent shortcut learning, we reduce branch coupling instead of expanding network width. Second, we construct soft constraints instead of hard ones from monocular depth to improve tolerance to monocular depth errors. Based on the principles, we integrate monocular information into both feature extraction and GRU iterations. Specifically, the monocular depth map is fused with the RGB image to sharpen depth boundary perception and suppress matching ambiguities. The fused image is then used for feature extraction, allowing the contextual features to encode global geometric information. Furthermore, the monocular depth gradient feature is employed to guide disparity updates, helping to escape local oscillations. Finally, to address the boundary blurring of supervised disparity caused by data augmentation, we propose an edge confidence estimation method and an edge-aware loss function. Our method achieves state-of-the-art (SOTA) performance on multiple standard benchmarks, demonstrating excellent generalization while improving accuracy. The code is available at https://github.com/linliboabc-maker/stereo-matching-digital.
\end{abstract}




\end{frontmatter}

\section{Introduction}\label{sec:intro}
{S}{tereo} matching (SM) aims to estimate dense, pixel-wise disparity maps from a pair of rectified stereo images \cite{SGM, belief_propagation}. The resulting disparity maps can subsequently be converted into metric depth for various downstream applications, including autonomous driving, robotics, augmented reality, etc. These applications require that SM methods generalize effectively across diverse real-world scenes \cite{Cross-domain}.

A typical SM pipeline \cite{chang2018pyramid, SGM} includes four stages: feature extraction, cost volume generation, cost aggregation, and disparity regression. After embedding these steps into an end-to-end deep network \cite{mayer2016large}, many deep learning-based SM methods emerged with significantly improved performance. Early deep SM primarily relied on filtering-based techniques to refine the cost volume and subsequently regress accurate disparities. These approaches typically begin by constructing a 3D or 4D cost volume, which is then filtered using 2D or 3D convolutional neural networks (CNNs)\cite{chang2018pyramid, GwcNet, 10890914, 11495076}. However, the single-stage regression makes the learned mapping highly dependent on training data. Then, any domain shift (e.g., changes in lighting, texture, or noise) at test time will cause severe performance degradation.

 Later, following the introduction of RAFT-Stereo \cite{Raft-stereo}, the iterative optimization paradigm become highly popular. Such methods \cite{wang2024selective, Feng2024MCStereoML, SHI2025111737, QI2024110410, ZHAO2025114062} first construct an all-pairs correlation volume and then index a local cost to extract motion features. These features subsequently guide recurrent units (ConvGRUs) [9] to iteratively refine the disparity map. By learning iterative update rules rather than a fixed mapping, these methods achieve much better generalization and accuracy. However, accurate disparity estimation remains challenging in inherently ambiguous regions, such as repetitive textures, textureless areas, occlusions, and blurred edges.

\begin{figure*}[htp]
\centering
\begin{minipage}{0.3\textwidth}
    \centering
    \includegraphics[width=\linewidth]{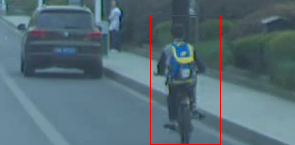}
\end{minipage}
\hspace{0.02\textwidth}
\begin{minipage}{0.3\textwidth}
    \centering
    \includegraphics[width=\linewidth]{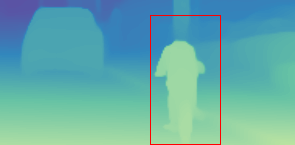}
\end{minipage}
\hspace{0.02\textwidth}
\begin{minipage}{0.3\textwidth}
    \centering
    \includegraphics[width=\linewidth]{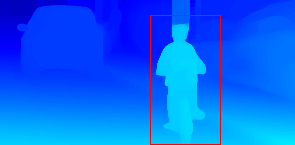}
\end{minipage}
\\[1ex]
\begin{minipage}{0.3\textwidth}
    \centering
    \includegraphics[width=\linewidth]{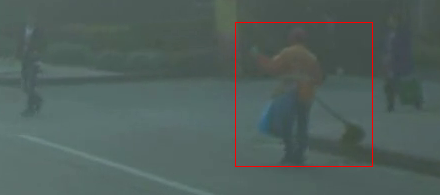}
\end{minipage}
\hspace{0.02\textwidth}
\begin{minipage}{0.3\textwidth}
    \centering
    \includegraphics[width=\linewidth]{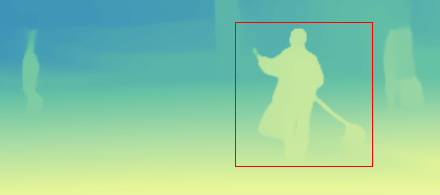}
\end{minipage}
\hspace{0.02\textwidth}
\begin{minipage}{0.3\textwidth}
    \centering
    \includegraphics[width=\linewidth]{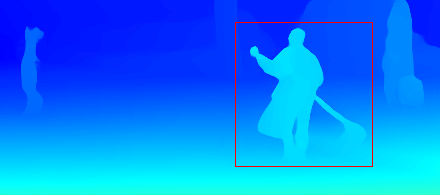}
\end{minipage}
\\[1ex]
\begin{minipage}{0.3\textwidth}
    \centering
    \includegraphics[width=\linewidth]{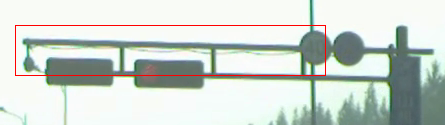}
\end{minipage}
\hspace{0.02\textwidth}
\begin{minipage}{0.3\textwidth}
    \centering
    \includegraphics[width=\linewidth]{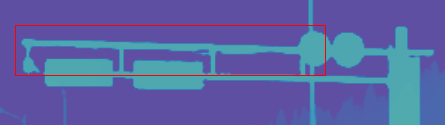}
\end{minipage}
\hspace{0.02\textwidth}
\begin{minipage}{0.3\textwidth}
    \centering
    \includegraphics[width=\linewidth]{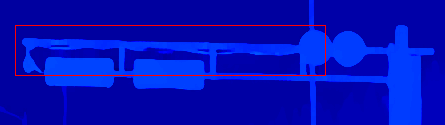}
\end{minipage}
\\[1ex]
\begin{minipage}{0.3\textwidth}
    \centering
    \includegraphics[width=\linewidth]{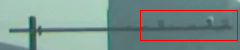}
    \small (a) RGB Image
\end{minipage}
\hspace{0.02\textwidth}
\begin{minipage}{0.3\textwidth}
    \centering
    \includegraphics[width=\linewidth]{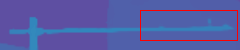}
    \small (b) Depth
\end{minipage}
\hspace{0.02\textwidth}
\begin{minipage}{0.3\textwidth}
    \centering
    \includegraphics[width=\linewidth]{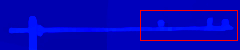}
    \small (c) Dispariry
\end{minipage}
\caption{A comparison between Depth Anything V2 depth maps and our disparity maps. The results show that in local blurred regions, our model effectively refines the disparity to recover clearer geometric structures.}
\label{fig: mono-vs-disp}
\end{figure*}

 Recent works leverage foundational vision models to enhance SM generalization. Features from SAM \cite{SAM10378323} and DINO \cite{DINO} have been used to improve matching robustness. Separately, monocular depth models (e.g., Depth Anything \cite{DepthAnything}) offer complementary strengths to SM. Monocular priors provide rich context but lack geometric precision, while stereo methods are geometrically accurate yet struggle in the ambiguous regions such as texture-less or occluded areas. To leverage their respective strengths, some researchers align monocular and binocular outputs in scale space. Although these methods yield notable improvements, achieving stable and generalizable alignment is difficult, and unreliable monocular cues can severely degrade performance. Moreover, how to effectively integrate monocular priors into SM remains an open question requiring deeper investigation.

This paper revisits how monocular depth information is embedded into SM. Motivated by the impact of architectural design on generalization, we observe that increasing network width tends to strengthen inter-branch coupling and may lead to shortcut learning, as discussed in the 'Insights for SM Network Architecture'. Therefore, we adopt the iterative optimization framework of RAFT-Stereo as our baseline. Furthermore, instead of imposing monocular cues as hard constraints, we extract general and lightweight features from monocular depth and incorporate them as soft constraints, thereby improving robustness to monocular depth errors, as illustrated in Fig. \ref{fig: mono-vs-disp}
 
 Based on these principles, we propose a monocular depth-guided iterative SM method. First, we embed the estimated monocular depth map into the original RGB stereo images. Since the depth map provides clear depth boundaries, this fusion constructs a new feature space for matching, enabling more accurate localization of depth discontinuities. Such boundary cues are particularly beneficial for coarse-to-fine disparity refinement, as they help constrain the valid disparity search range for each pixel. Second, although monocular depth is relative, the ordinal relationships between a pixel and its neighboring pixels are often reliable. We therefore introduce monocular gradient feature to guide disparity updates. Finally, random scaling is commonly used as data augmentation to improve the model's robustness to variations in disparity scale. However, it may also lead to blurring at depth edges. To mitigate this issue, we propose an edge detection method combined with an edge-aware loss. To summarize, our main contributions are:
\begin{itemize}
  \item  This paper presents a simple yet effective method that revisits monocular depth embedding for generalized SM. By fusing the monocular depth map with the original image for feature extraction and leveraging depth gradients to guide disparity updates, the proposed approach clearly improves matching accuracy in ill-posed regions.
  \item To address edge blurring in disparity maps caused by spatial-domain data augmentation techniques, we propose an edge detection method coupled with an edge-aware loss. This enables the model to localize disparity edge more accurately.
  \item We propose a SM algorithm with good generalization ability, achieving state-of-the-art (SOTA) performance on multiple datasets, including KITTI,  Middlebury-H, DrivingStereo, and  ETH3D.
\end{itemize}

\section{Related Work}\label{sec:related}
In this section, we first review deep SM methods without considering Generalization. Then, we discuss the Generalized SM models.

\subsection{Deep SM Ignoring Generalization}

 After Mayer \cite{mayer2016large} embedding all the core steps of SM (i.e., feature extraction, cost volume generation, cost aggregation, and disparity regression) into an end-to-end deep network, the filtering-based pipeline has been widely adopted. Based on the cost volume construction, these methods can be divided into correlation-based and concatenation-based approaches. Correlation-based methods (e.g., SegStereo \cite{SegStereo}, EdgeStereo \cite{song2020edgestereo} and MAFNet \cite{Xu2025MAFNetMultifrequencyAF}) compute cross-view feature correlations to build the cost volume, enabling efficient 2D convolution-based disparity estimation. However, they suffer from loss of semantic and structural information in feature representation due to the correlation operation. Concatenation-based approaches build the cost volume by concatenating the left and the right features and aggregate the cost with 3D convolutions, such as PSMNet \cite{chang2018pyramid} and GwcNet \cite{GwcNet}. For more effective cost aggregation, GA-Net \cite{Ganet} introduced content-aware aggregation layers. Bangunharcana \emph{et al.} \cite{Correlate_and_Excite} excited cost volume channels via weights derived from reference image features. While theses methods can achieve superior performance with 3D-CNN, they fail to generalize to unseen environments without fine-tuning due to domain shift between the test and train domain.

\subsection{Generalized SM Models}
To bridge the gap between the test and train domains, some generalized SM models have been proposed. They can be categorized into three types: robust feature extraction-based, iterative optimization paradigm-based, and Vision Foundation Model-guided. A detailed introduction is given below.

\textbf{Robust Feature Extraction-based:} They aim to learn a feature extraction module invariant to imaging variations like illumination and color. DSMNet \cite{Domain_Invariant} introduced domain normalization to stabilize features across
domains. GraftNet \cite{graftnet} leveraged the broad-spectrum feature trained on large-scale datasets to deal with the domain shift since it has seen various styles of images. ITSA \cite{chuah2022itsa} employed an information theoretic strategy to minimize feature sensitivity to input perturbations. HVT \cite{chang2023domain} proposed a hierarchical data augmentation method to enforce domain-invariant feature extraction, including global, local and pixel-wise levels. Masked-stereo \cite{rao2023masked} jointly trains SM and image reconstruction in a pseudo-multi-task framework, promoting models to learn structure information.  HODC \cite{Hierarchical_object_aware} enforces object- and pixel-level consistency via dual-level contrastive learning, improving robustness to scene-level changes.

\textbf{Iterative Optimization Paradigm-based:} The first iterative SM approach, namely RAFT-Stereo \cite{Raft-stereo}, adapts RAFT's optical flow framework \cite{RAFT} by introducing an all-pairs cost volume pyramid. It extracts local correlation features from the pyramid, iteratively updates disparity via GRU-based operators. Later, some its variants are proposed.
To handle both edges and smooth regions effectively, Selective-Stereo \cite{wang2024selective} fused hidden disparity cues extracted by convolutions with varying receptive fields. MC-Stereo \cite{Feng2024MCStereoML} replaced the fixed search range in RAFT-Stereo with a multi-peak lookup strategy for local correlation indexing. IGEV-Stereo \cite{iterative_Geometry} estimated an initial disparity to accelerate GRU convergence and leverages global geometry from the 4D cost volume to improve accuracy. To build a more robust cost volume, GREAT-Stereo \cite{Global_Regulation_Excitation} refined matching features with spatial attention and extracts epipolar global context via matching attention, while DiffuVolume \cite{diffuvolume} employed a diffusion model as a recurrent filter to remove redundant information from the cost volume. Zeng \emph{et al.} \cite{Parameterized_Cost_Volume} accelerate RAFT-Stereo by 4--15 times using a multi-Gaussian parameterized cost volume. Differently, Wu \emph{et al.} \cite{Few-shot} formulated SM as dynamic optimization within a solution space, and using an adaptive recurrent 3D CNN to determine the optimal solution. Unlike filter-based methods, iterative SM methods generalize better by learning optimization rules. However, their generalization performance remains limited.

\begin{figure*}[htp]
\centering
\begin{minipage}{0.3\textwidth}
    \centering
    \includegraphics[width=\linewidth]{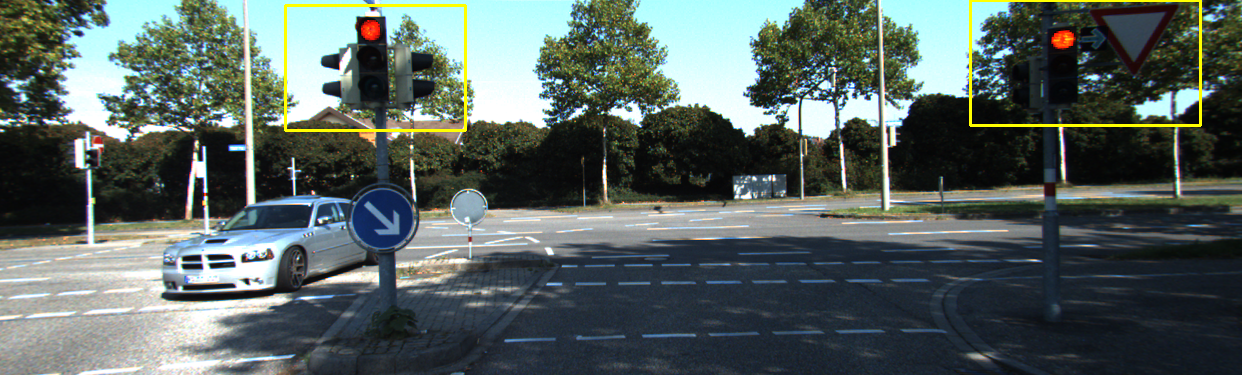}
\end{minipage}
\hspace{0.02\textwidth}
\begin{minipage}{0.3\textwidth}
    \centering
    \includegraphics[width=\linewidth]{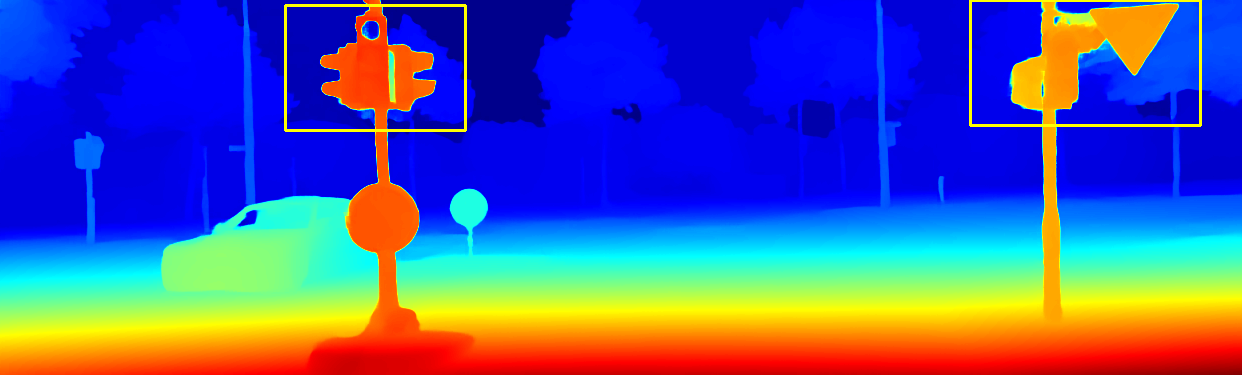}
\end{minipage}
\hspace{0.02\textwidth}
\begin{minipage}{0.3\textwidth}
    \centering
    \includegraphics[width=\linewidth]{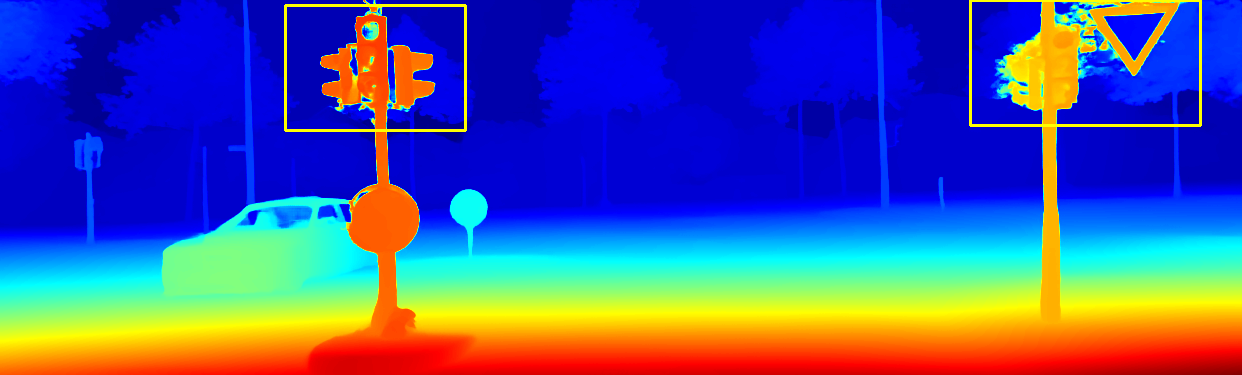}
\end{minipage}
\\[1ex]
\begin{minipage}{0.3\textwidth}
    \centering
    \includegraphics[width=\linewidth]{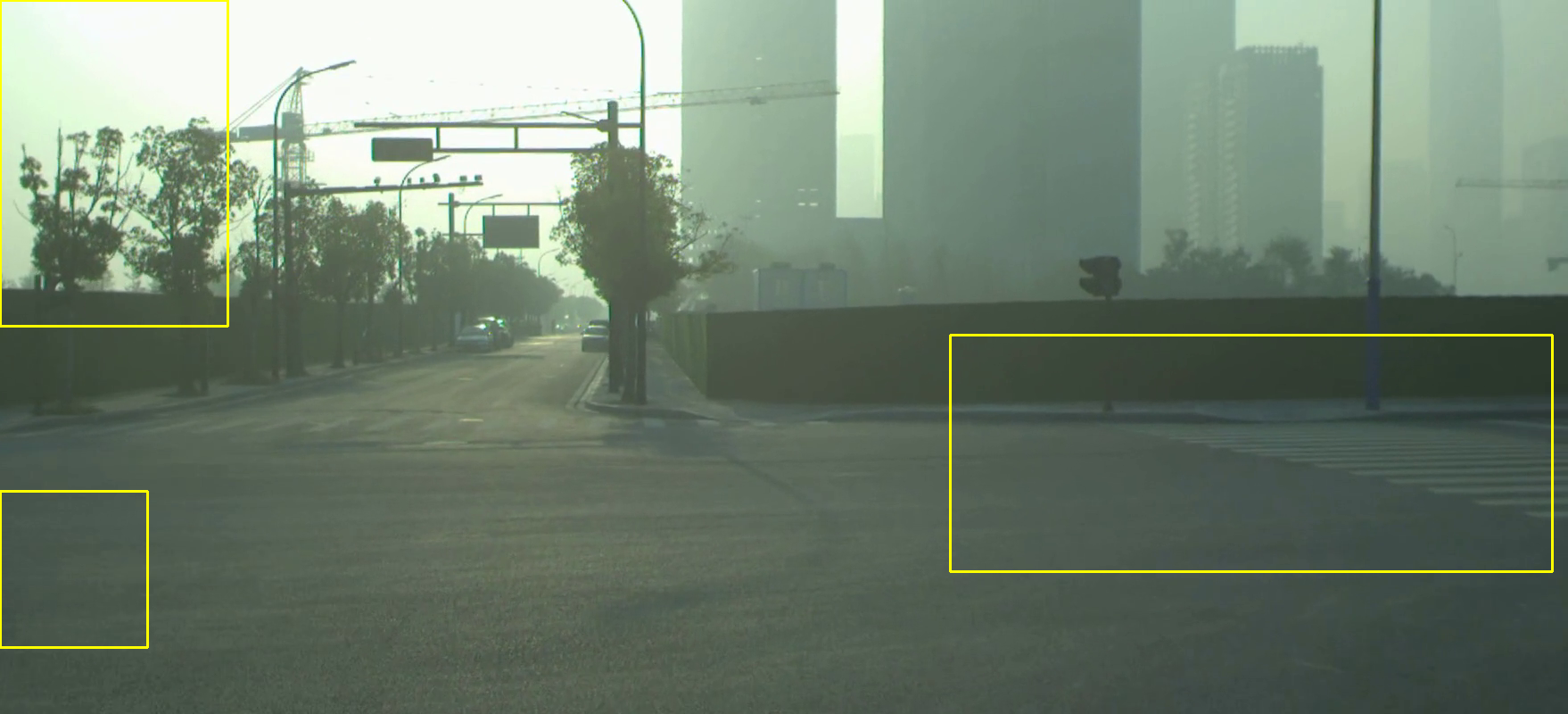}
    \small (a) RGB Image
\end{minipage}
\hspace{0.02\textwidth}
\begin{minipage}{0.3\textwidth}
    \centering
    \includegraphics[width=\linewidth]{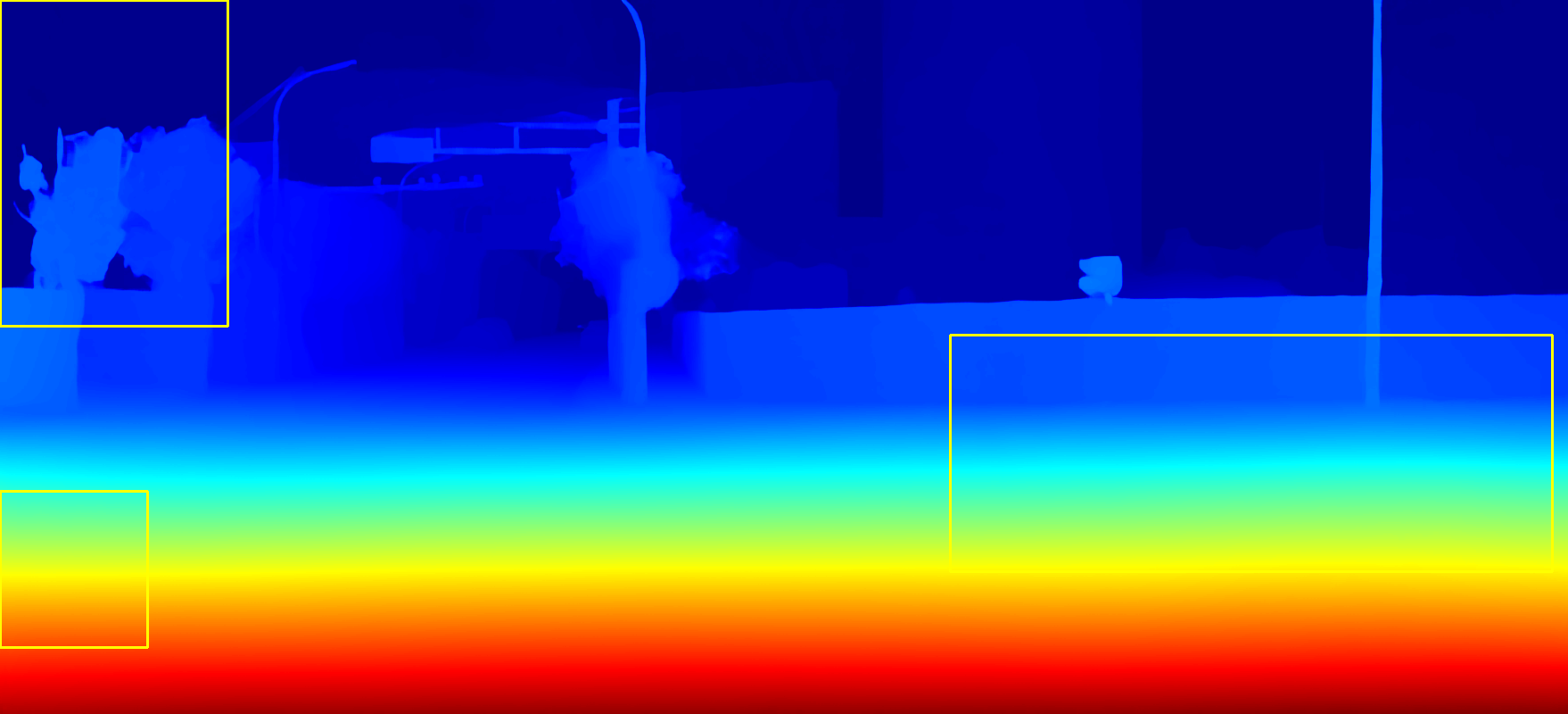}
    \small (b) RAFT-Stereo
\end{minipage}
\hspace{0.02\textwidth}
\begin{minipage}{0.3\textwidth}
    \centering
    \includegraphics[width=\linewidth]{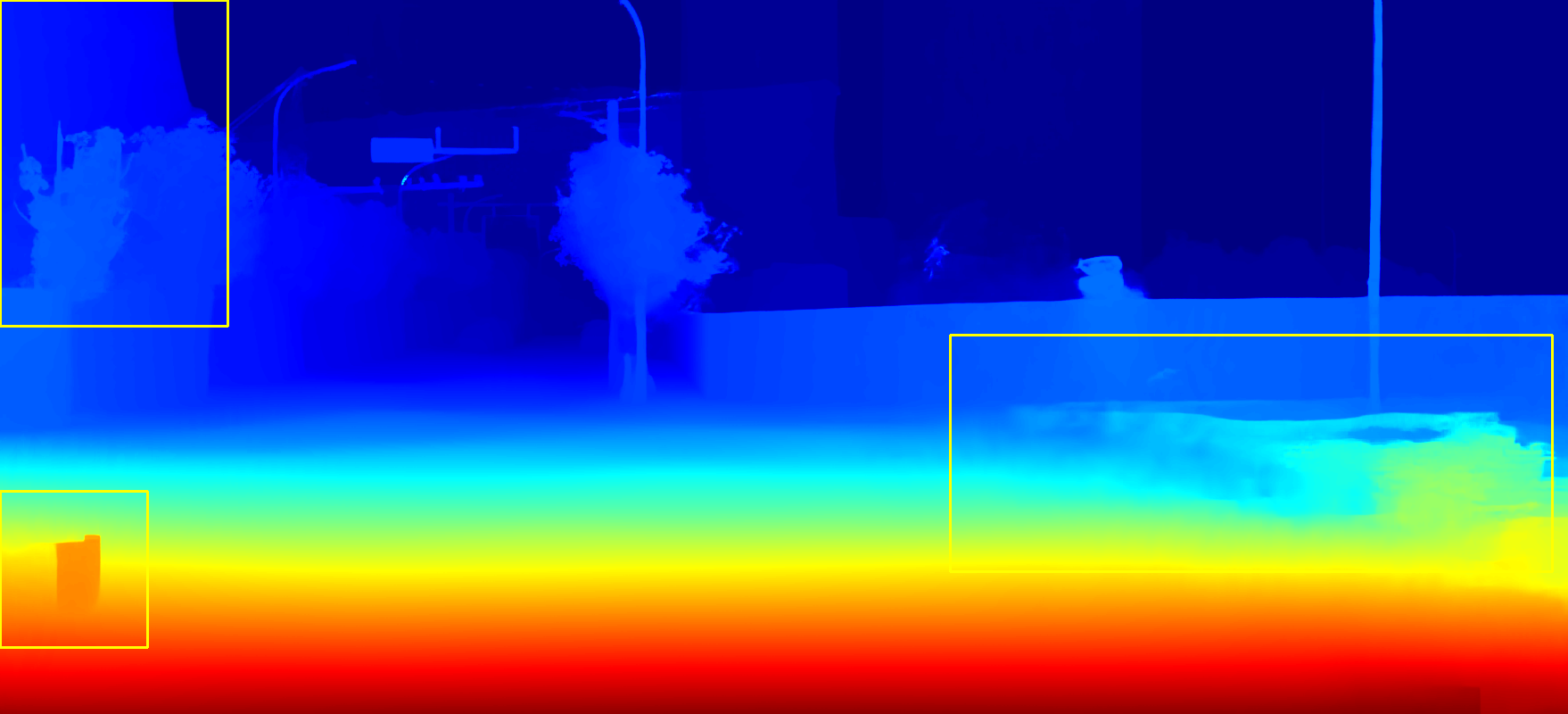}
    \small (c) Selective-Stereo
\end{minipage}
\caption{RAFT-Stereo is compared to Selective-Stereo without fine-tuning. While Selective-Stereo excels after fine-tuning on new scenes, it struggles in real-world scenarios when used without fine-tuning.}
\label{fig: raft-vs-selective}
\end{figure*}

\textbf{Vision Foundation Model-based:}
 The commonly used vision foundation models (VFM) include SAM \cite{SAM10378323}, DINO\cite{DINO} and Depth Anything series \cite{DepthAnything}. AIO-Stereo introduced a dual-level selective knowledge transfer module to leverage the strengths of these VFMs. However, running multiple VFMs significantly increases memory usage and inference time. To avoid this, many recent works rely solely on DepthAnything V2  \cite{Stereo_Anywhere, Defom, Monster, Los, yao2025diving, FoundationStereo, generalizedgeometryencoding, 11121379}. FoundationStereo \cite{FoundationStereo} constructed a large-scale synthetic dataset and bridges the sim-to-real gap by adapting frozen DepthAnythingV2 priors within its feature backbone. GGEV \cite{generalizedgeometryencoding} incorporated depth structural priors to enhance the fragile matching in the initial cost volume. To align relative monocular depth with absolute disparity depth, Yao \emph{et al.} \cite{yao2025diving} proposed a binary local ordering map that converts depth into a unified binary relative format to guide fusion. MonSter \cite{Monster} adopted a dual-branch architecture where monocular depth and stereo matching iteratively refine each other. LoS \cite{Los} first extracted local structure information from monocular depth to capture scene geometry, and used it to enhance disparity initialization, optimization, and refinement.
 DEFOM-Stereo \cite{Defom} augmented plain CNN encoders with DepthAnythingV2 for more powerful feature encoding. Bartolomei \emph{et al.} \cite{Stereo_Anywhere} integrated monocular depth cues into SM through iterative fusion of RGB-based and depth-based cost volumes.
ViTAStereo \cite{11121379} leveraged the monocular depth prior from Depth Anything V2 to self-supervise disparity map training through a composite loss function, which consists of a Local Depth Ranking (LDR) loss and a Dual Disparity Smoothness (DDS) loss.

 Monocular-assisted methods have significantly improved SM generalization but often depend on intricate alignment and fusion of monocular and binocular information. Such methods either introduce significant computational overhead or exhibit high sensitivity to errors in monocular priors. To address this challenge, we aim to explore fundamental features from monocular depth to effectively guide SM.

\section{Insights for SM Network Architecture}\label{sec:insightsm}

As we know, iterative optimization-based SM methods, such as RAFT-Stereo, demonstrate superior generalization compared to filter-based algorithms. The process of RAFT-Stereo can be summarized as follows:
\begin{equation}
\begin{split}
\label{raft-stereo}
    &\mathbf{d}_{k} = \mathbf{d}_{k-1} + \Delta\mathbf{d}_k  \\
    & \Delta\mathbf{d}_k = \text{Conv}(\mathbf{h}_k)\\
\end{split}
\end{equation}

where $\mathbf{h}_k$ and $\mathbf{d}_k$ are the hidden state and disparity at the $k$-th GRU. Instead of learning a one-step mapping function, it learns to progressively refine disparity rules. Several RAFT-Stereo variants, such as Selective-Stereo \cite{wang2024selective} and IGEV-Stereo \cite{iterative_Geometry} are proposed. They indeed outperform RAFT-Stereo when fine-tuned on new scenes. However,  we find their generalization ability drops when directly applied to real-world scenarios without fine-tuning, as demonstrated in Table \ref{tab:zero_comparison}. Fig. \ref{fig: raft-vs-selective} shows two examples. Below, we take Selective-Stereo as an example to analyze the reasons.

 In addition to the core state variables $\mathbf{h}_k$ and $\mathbf{d}_k$, Selective-Stereo introduces an attention map $\mathbf{A}$ , which regulates GRU information as follows:
\begin{equation}
\label{selective-stereo}
    \mathbf{h}_k = \mathbf{A} \odot \mathbf{h}_k^s + (1 - \mathbf{A}) \odot \mathbf{h}_k^l
\end{equation}
where $\mathbf{h}_k^s$ and $\mathbf{h}_k^l$ denote the hidden state obtained with small and large kernel sizes.

\begin{figure*}[htp]
    \centering
    \includegraphics[trim=25 25 40 20, clip, width=\textwidth, page=1]{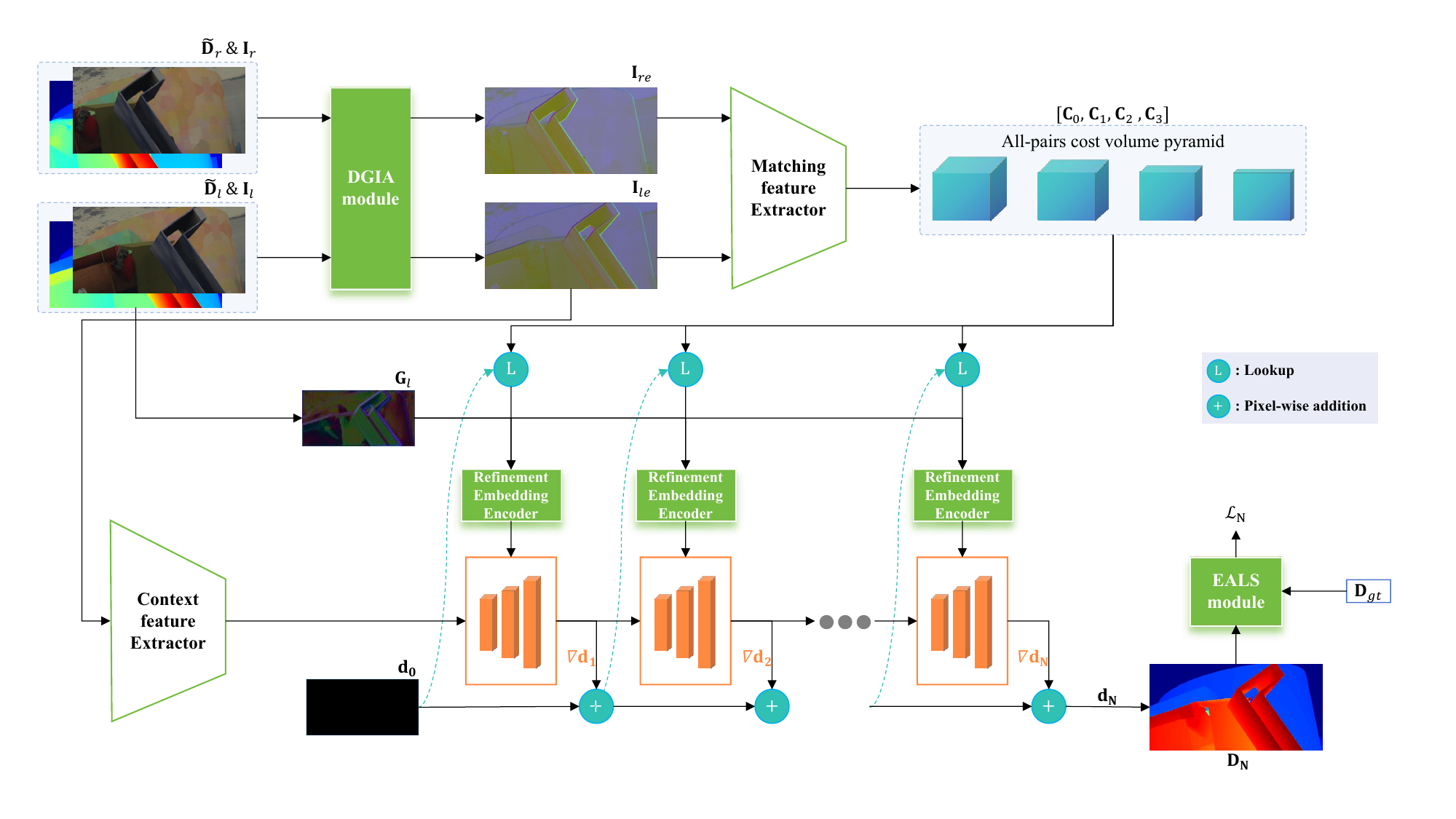}
    \caption{Architecture of the proposed SM framework. The pipeline consists of three main components: depth-guided image augmentation (DGIA), gradient-guided GRU refinement (gradient-GRU), and edge-aware loss suppression (EALS).}
    \label{fig:overview_framework}
\end{figure*}

$\mathbf{A}$ is an independent module for analyzing image content, which is optimized alongside the GRU module (i.e., $\mathbf{h}_k$ and $\mathbf{d}_k$) with disparity supervision. $\mathbf{A}$ and GRU are tightly coupled, which easily induces spurious correlations in the training set, resulting in shortcut learning. In the test phase, accurate A and GRU hidden state are both required; otherwise, performance degrades sharply.
In contrast, RAFT-Stereo optimizes only its multiple cascaded GRUs, enabling it to learn an essential coarse-to-fine disparity update rule. This insight motivates us to avoid highly coupled parallel branches when designing generalized SM models. To this end, this paper integrates monocular information without altering RAFT-Stereo main architecture, leveraging monocular depth cues to enhance both feature extraction and GRU updates.


\section{Methodology}\label{sec:method}
The overall architecture of the proposed SM framework is illustrated in Fig. \ref{fig:overview_framework}. It comprises three key components: 1) depth-guided image augmentation (DGIA), 2) gradient-guided GRU refinement (Gradient-GRU), and 3) an edge-aware loss. Specifically, the DGIA module augments the input RGB images with DepthAnything V2 depth to enrich geometric representations. The Gradient-GRU module derives gradient maps from the depth prior and exploits them to guide GRU-based refinement. Furthermore, an edge-aware loss is introduced to alleviate the boundary blurring of supervised disparity signals caused by random scaling. The details of these components are described in the following subsections.

\subsection{Depth-Guided Image Augmentation}

Reliable pixel matching and precise disparity edge localization are challenging in ambiguous regions, particularly in areas with weak texture, repetitive patterns, and occlusions. However, monocular depth maps provide complementary geometric cues for RGB images, clearly indicating depth (disparity) edges and continuity. Motivated by this, we fuse the RGB image and depth map for feature extraction. The fused representation not only suppresses ambiguous RGB textures that hinder SM, but also enhances responses around disparity edges. The edge cues are crucial for the coarse-to-fine disparity refinement process, as they directly constrain and narrow the valid disparity search range for each pixel.

\begin{figure*}[htp]
    \centering
    \includegraphics[trim=30 150 10 70, clip, width=\linewidth, page=2]{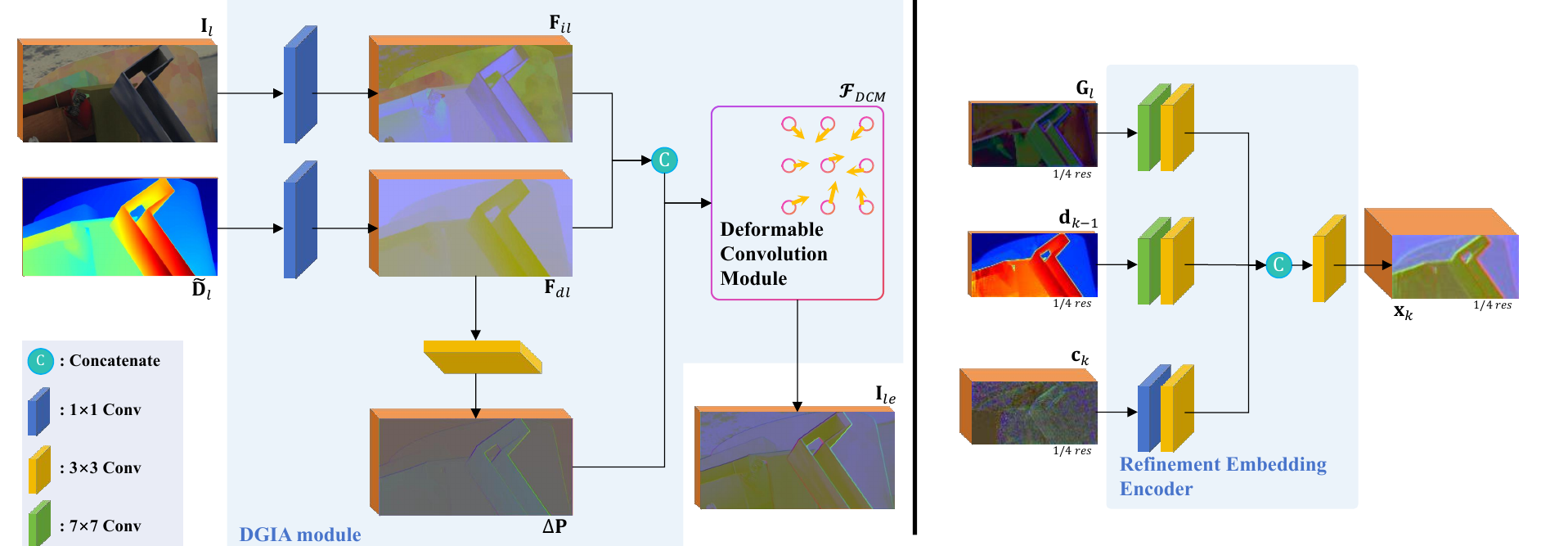}
    \caption{Architectures of the DGIA module (left) and the refinement embedding encoder (right).}
    \label{fig:dgia-ree}
\end{figure*}

Let $\mathbf{I}_l$ and $\mathbf{I}_r$ denote RGB images of the left and right views, respectively. After being processed by Depth Anything V2, their monocular depth maps, $\mathbf{\tilde{D}}_l$ and $\mathbf{\tilde{D}}_r$, are obtained. As illustrated in the left panel of Fig. \ref{fig:dgia-ree} (using the left view as an example), \(1\times1\) convolutions are first applied to $\mathbf{I}_l$ and $\mathbf{\tilde{D}}_l$ to obtain 8-channel feature maps $\mathbf{F}_{il}$ and $\mathbf{F}_{dl}$. The enhanced left view $\mathbf{I}_{le} \in \mathbb{R}^{3 \times H \times W}$  is then estimated as:
\begin{equation}
\begin{split}
\label{image-fusion}
    &\mathbf{I}_{le} = \mathcal{F}_\text{DCM}(\Delta\emph{\textbf{P}} ,\text{Concat}(\mathbf{F}_{il},\mathbf{F}_{dl})) \\
    & \Delta\emph{\textbf{P}} = \text{Conv}_{3\times3}(\mathbf{F}_{dl})\\
\end{split}
\end{equation}
where $\mathcal{F}_\text{DCM}$ denote the deformable convolution operation. The offset $\Delta\emph{\textbf{P}}$ denotes the spatial displacement adjusting each sampling position in the standard convolution kernel. $\text{Concat}(\mathbf{F}_{il},\mathbf{F}_{dl})$ performs channel-wise concatenation. Similarly, the enhanced right view $\mathbf{I}_{re} \in \mathbb{R}^{3 \times H \times W}$  is obtained.

\subsection{Gradient-Guided GRU Refinement}
\textbf{Matching feature Extraction}: Given the enhanced $\mathbf{I}_{le}$ and $\mathbf{I}_{re}$, a $7\times7$ convolutional layer first increases the channel dimension from 3 to 64. Cascaded residual blocks then expand the channels to 128 while reducing the spatial resolution to 1/4. Finally, $1\times1$ convolutional layer is used to output the feature
$\mathbf{F}_{l} \in \mathbb{R}^{ C\times H/4 \times W/4 }$ and $\mathbf{F}_{r} \in \mathbb{R}^{C\times H/4 \times W/4 }$, where $C=256$.

\textbf{Context feature Extraction}:
The context feature is extracted from the enhanced left-view image $\mathbf{I}_{le}$, which encodes global geometric cues and is used to guide the update of the hidden state in GRU. The main architecture includes a $7\times7$ convolutional layer followed by multiple residual blocks. The resulting features are further processed by output heads composed of residual blocks and $3\times3$ convolutional layers, yielding multi-level features \(\{\mathbf{F}_{\text{net}}^{i}, \mathbf{F}_{\text{inp}}^{i}\}\) $(i = 1, 2, 3)$ at 1/4, 1/8, 1/16 resolutions. Then \(\mathbf{F}_{\text{net}}^{i}\) is used to initialize the hidden state of GRU, and \(\mathbf{F}_{\text{inp}}^{i}\) serves as the bias for the update gate, reset gate, and hidden state in GRU.

\textbf{Cost Volume Construction}:
Given the left and the right features $\mathbf{F}_{l} \in \mathbb{R}^{ C\times H/4 \times W/4 }$ and $\mathbf{F}_{r} \in \mathbb{R}^{C\times H/4 \times W/4 }$, we first construct an all-pairs correlate cost volume. The cost for points
$(i,j)$ (left view) and $(i,k)$ (right view) is computed as follows:
\begin{equation}
\label{costVolume}
   \mathbf{C}_{ijk} = \frac{1}{\sqrt{C}}\sum_h {(\mathbf{F}_{l})}_{hij} \cdot {(\mathbf{F}_{r})}_{hik}, \quad \mathbf{C} \in \mathbb{R}^{{H/4} \times {W/4} \times {W/4}}
\end{equation}
Then, a 4-level cost pyramid $\mathbf{C}_{i} (i = 0, 1, 2, 3)$ is constructed by performing average pooling with a stride of 2 along the last dimension.

\textbf{GRU update structure}:
The GRU estimates the current hidden state $\mathbf{h}_k$ using the current refinement embedding $\mathbf{x}_{k}$ and the previous hidden state $\mathbf{h}_{k-1}$. Based on the hidden state $\mathbf{h}_k$, we decode a residual disparity $\Delta\mathbf{d}_k$
through two convolutional layers, then we update the current disparity $\mathbf{d}_{k+1} = \mathbf{d}_k + \Delta\mathbf{d}_k$. The vector $\mathbf{d}_0$ is initialized to all zeros. Finally, the disparity estimated at 1/4 resolution is up-sampled to full resolution using the up-sampling method in RAFT-Stereo. At 1/4 resolution, the specific steps of the GRU are as follows:
\begin{equation}
 \begin{split}
 \label{GRU-update}
&\mathbf{z}_k = \sigma\big( \text{Conv}([\mathbf{h}_{k-1}, \mathbf{x}_k], W_z) + \mathbf{b}_z \big) \\
&\mathbf{r}_k = \sigma\big( \text{Conv}([\mathbf{h}_{k-1}, \mathbf{x}_k], W_r) + \mathbf{b}_r \big) \\
&\tilde{\mathbf{h}}_k = \tanh\big( \text{Conv}([\mathbf{r}_k \odot \mathbf{h}_{k-1}, \mathbf{x}_k], W_h) + \mathbf{b}_h \big) \\
&\mathbf{h}_k = (1 - \mathbf{z}_k) \odot \mathbf{h}_{k-1} + \mathbf{z}_k \odot \tilde{\mathbf{h}}_k
\end{split}
\end{equation}
where $\mathbf{z}_k$, $\mathbf{r}_k$ and $\tilde{\mathbf{h}}_k$ are the update gate, reset gate, and candidate hidden state, respectively. $\mathbf{h}_0$ is initialized as $\mathbf{F}_{\text{net}}^1$. The bias $\mathbf{b}_z$, $\mathbf{b}_r$ and $\mathbf{b}_h$ are initialized as $\mathbf{F}_{\text{inp}}^1$.
Extracted from the enhanced left image $\mathbf{I}_{le}$, $\mathbf{F}_{\text{inp}}^1$ encodes the global geometric information. Acting as bias in GRU to govern disparity updates, it like a geometry-guided filtering mechanism that prevents the disparity map from being perturbed by local noise.

\begin{figure*}[htp]
    \centering
    \includegraphics[trim=10 20 10 10, clip, width=\textwidth, page=3]{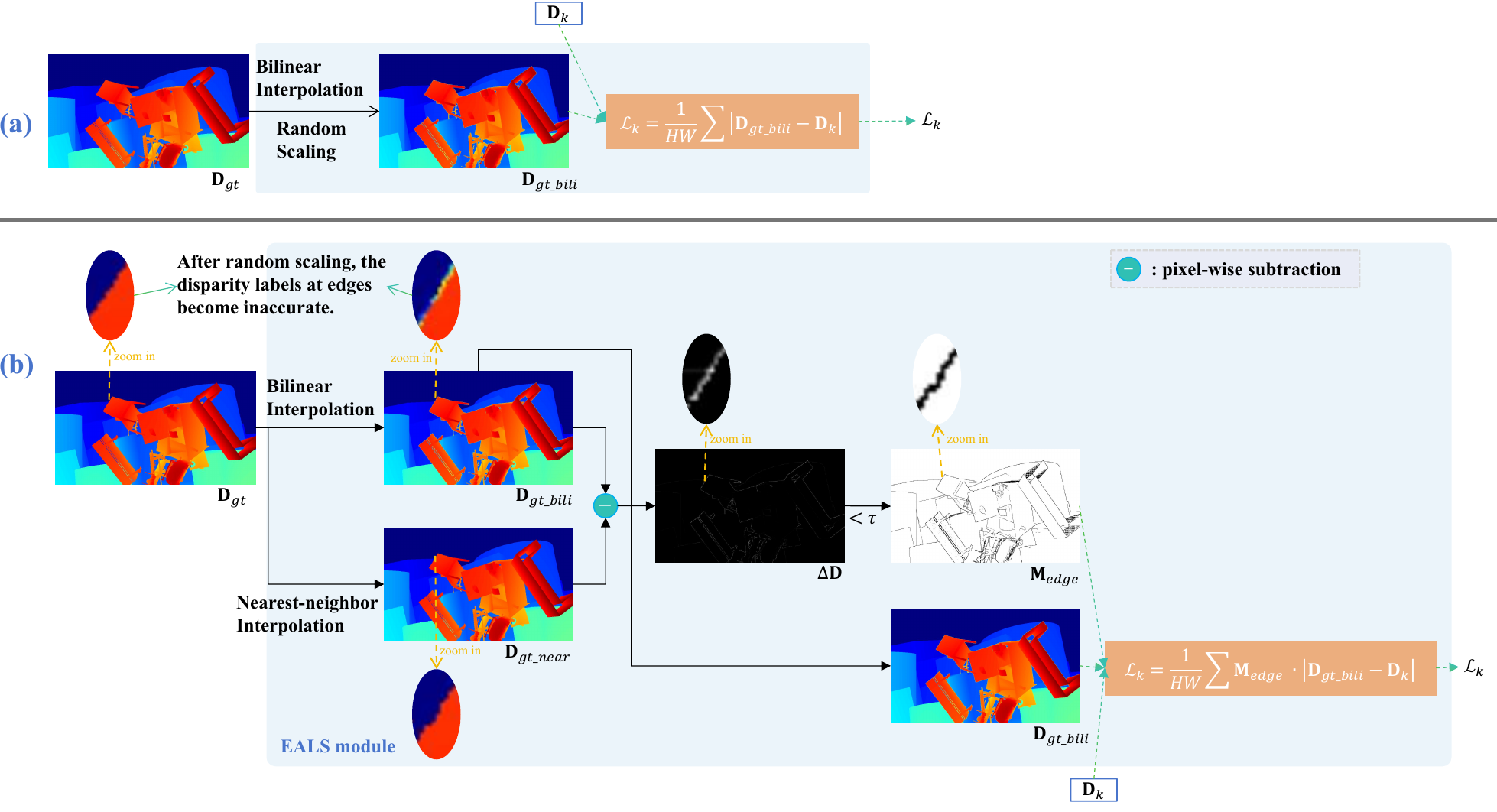}
    \caption{(a) Existing methods directly use the inaccurate disparity maps caused by random scaling for supervision. (b) The proposed edge-aware loss.}
    \label{fig:Edge-Aware Loss Suppression Module}
\end{figure*}

At each GRU step, the input $\mathbf{x}_{k}$ is computed with disparity $\mathbf{d}_{k-1}$, the local correlation volume $\mathbf{c}_{k-1}$ indexed by disparity $\mathbf{d}_{k-1}$, and the gradient information $\mathbf{G}_{l}$ as shown in the right panel of Fig. \ref{fig:dgia-ree}. The local cost volume is derived from the all-pairs correlation pyramid in the same way as RAFT-Stereo. The gradient information $\mathbf{G}_{l} \in \mathbb{R}^{3\times H/4 \times W/4 }$, including horizontal, vertical gradients of monocular depth, and gradient magnitude map. To reduce inter-sample scale variation in the gradient magnitude, instance normalization is applied. Specifically, $\tilde{\mathbf{D}}_l$ is first downsampled to $1/4$ resolution using bilinear interpolation and then normalized by instance normalization to obtain $\tilde{\mathbf{D}}_{l\_low} \in \mathbb{R}^{1 \times H/4 \times W/4}$. The following operations are applied to $\tilde{\mathbf{D}}_{l\_low}$ to obtain $\mathbf{G}_l$:

\begin{equation}
\begin{split}
&\mathbf{G}_x = \tilde{\mathbf{D}}_{l\_low} * \mathbf{K}_x, \text{  where } \mathbf{K}_x = \begin{bmatrix} -1 & 0 & 1 \\ -2 & 0 & 2 \\ -1 & 0 & 1 \end{bmatrix},
\\
&\mathbf{G}_y = \tilde{\mathbf{D}}_{l\_low} * \mathbf{K}_y, \text{  where } \mathbf{K}_y = \begin{bmatrix} -1 & -2 & -1 \\ 0 & 0 & 0 \\ 1 & 2 & 1 \end{bmatrix},
\\
&\mathbf{M} = \sqrt{\mathbf{G}_x^2 + \mathbf{G}_y^2 + \varepsilon},
\\
&\mathbf{G}_{l,0} = \frac{\mathbf{G}_x}{\mathbf{M} + \varepsilon}, \text{  }
\mathbf{G}_{l,1} = \frac{\mathbf{G}_y}{\mathbf{M} + \varepsilon}, \text{  }
\mathbf{G}_{l,2} = \mathrm{IN}(\mathbf{M}),
\\
&\mathbf{G}_l = \mathrm{Concat}\bigl( \mathbf{G}_{l,0},\; \mathbf{G}_{l,1},\; \mathbf{G}_{l,2} \bigr)
\end{split}
\end{equation}

where IN denotes  instance normalization. This paper incorporates gradient features to guide hidden-state updates. This design is motivated by the observation that gradient directions in monocular depth are relatively reliable and capture local relative structures between a pixel and its neighbors, making them particularly useful for updating GRU hidden states in hard-to-match regions.

Instead of directly using the gradient features to guide disparity updates, we employ them in hidden-state updates for two reasons. First, as discussed in the "Insights for SM Network Architecture" section, directly guiding disparity updates with gradient features would make disparity estimation jointly dependent on the hidden state and monocular gradients, leading to excessive coupling during training and potentially harming generalization. Second, when monocular gradient features are unreliable, such strong dependence may further degrade performance.

\subsection{Edge-Aware Loss Suppression}
In the training process of SM, random scaling operations such as bilinear interpolation and nearest-neighbor interpolation are often adopted as data augmentation techniques to improve the model's robustness to scale variations. However, random scaling renders disparity labels unreliable near boundaries. Existing methods directly use these inaccurate disparity maps for supervision, as illustrated in Fig. \ref{fig:Edge-Aware Loss Suppression Module}(a).

To address this issue, we propose an edge-aware loss. The key insight is that at disparity boundaries, bilinear interpolation produces blurred values by averaging neighboring pixels, whereas nearest-neighbor interpolation maintains sharp discontinuities by directly selecting a neighboring pixel value. We therefore approximate the edge mask by exploiting the discrepancy between these two interpolation results. As shown in Fig. \ref{fig:Edge-Aware Loss Suppression Module}(b), we first generate
$\mathbf{D}_{gt\_bili}$ and $\mathbf{D}_{gt\_near}$ by applying bilinear and nearest-neighbor interpolation to the ground-truth disparity map $\mathbf{D}_{gt}$, respectively, and then compute their absolute difference
$\Delta \mathbf{D}=|\mathbf{D}_{gt\_bili} - \mathbf{D}_{gt\_near}|$. Then, the binary edge mask is defined as:
\begin{equation}
 \label{edge}
\mathbf{M}_\text{edge} = \mathbf{1}_{\{\Delta \mathbf{D} \le \tau\}} \\
\end{equation}
where $\mathbf{1}_{\{.\}}$ is the indicator function, which equals 1 when the condition inside the braces holds, and 0. otherwise. Then the loss function is defined as:

\begin{equation}
 \begin{split}
 \label{loss}
 &\mathcal{L}_k = \frac{1}{HW} \sum \mathbf{M}_\text{edge} \cdot |\mathbf{D}_{gt\_bili} - \mathbf{D}_{k}| \\
 &\mathcal{L} = \sum_{k=1}^{N} \gamma^{N-k}\mathcal{L}_k, \text{  } \gamma=0.9 \\
\end{split}
\end{equation}
$\mathcal{L}_k$ denotes the loss function defined for the $k-th$ GRU.

\section{Experiments and Results}\label{sec:exp}
\subsection{Datasets and Evaluation Metrics}

We train all models on the synthetic SceneFlow dataset, which contains about 35,000 stereo image pairs with dense disparity ground truth. SceneFlow provides three subsets: FlyingThings3D, Driving, and Monkaa. To evaluate the generalization ability of our method, we test on five real-world benchmarks that cover diverse scenes and challenging conditions:

\textbf{Middlebury:} A high-resolution indoor stereo dataset providing subpixel-accurate disparity ground truth, which contains fifteen samples. We evaluate our method via endpoint error (EPE) and the percentage of pixels with disparity error exceeding 2 pixels (D2).

\textbf{KITTI 2015:} A real-world driving stereo benchmark featuring dynamic street scenes. It is officially split into 200 training stereo pairs with LiDAR-derived sparse disparity labels and 200 unannotated test pairs. Only the training split is adopted for our experiments, and we report EPE as well as the percentage of pixels with disparity error exceeding 3 pixels (D3).

\textbf{ETH3D:} A grayscale stereo benchmark covering indoor and outdoor low-texture scenarios, with an official split of 27 training pairs and 20 test pairs. We only utilize its training subset and measure EPE and the percentage of pixels with disparity error exceeding 1 pixel (D1).

\textbf{DrivingStereo:} A large-scale autonomous driving stereo dataset with an auxiliary weather subset consisting of 2000 stereo image pairs evenly distributed over four weather conditions (sunny, cloudy, foggy, rainy), with 500 samples per category. This weather subset is entirely extracted from the original training data, and all experiments are conducted on this subset. For each weather category, we report EPE and D3.

For all datasets, all evaluation metrics are calculated on complete disparity maps without pre-filtering or occlusion masking unless otherwise specified.

\subsection{Implementation Details}

All experiments (including generalization comparisons and ablation studies) adopt the same training configuration as RAFT‑Stereo. We use the AdamW optimizer with an initial learning rate of $0.0002$ and a weight decay of $1\times10^{-5}$. The learning rate is scheduled by OneCycleLR, where the warmup phase occupies $1\%$ of the total steps and the learning rate is linearly annealed to near zero. The data augmentation strategy includes random color jitter (brightness, contrast, saturation, and hue), random rectangular occlusions added to the right image, as well as spatial transformations such as random scaling, cropping, and vertical jitter. These augmentations help improve the generalization capability of our model across diverse scenes. The only difference is that we set batch size to 4. All models are trained on a single NVIDIA GeForce RTX 4090 GPU with $24$ GB of memory, and the code is implemented in PyTorch.

\subsection{Comparison with State-of-the-Art Methods}

We compare our method with two groups of state-of-the-art stereo matching approaches. The first group consists of classic iterative refinement methods, including RAFT-Stereo \cite{Raft-stereo}, IGEV-Stereo \cite{iterative_Geometry}, and Selective-IGEV \cite{wang2024selective}. The second group leverages monocular depth priors from Depth Anything V2, including DEFOM-Stereo \cite{Defom}, Monster \cite{Monster} and MGStereo \cite{Feng2024MCStereoML}. All models are trained on the SceneFlow dataset. 

\begin{figure*}[htp]
\centering
\small
\setlength{\tabcolsep}{1.4pt}
\begin{tabular}{p{3.9cm} p{3.9cm} p{3.9cm} p{3.9cm}}
\myimage{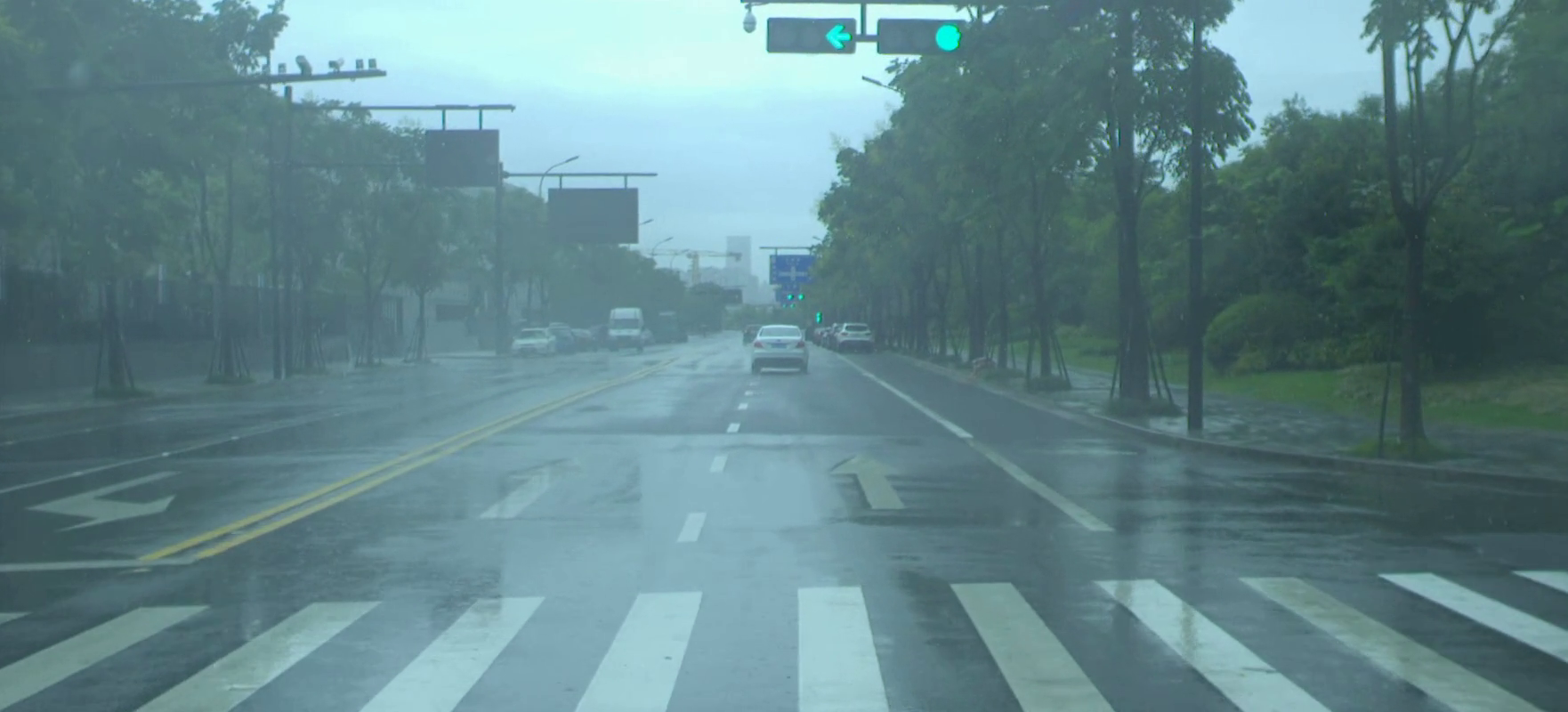}{./all-models-generalization/rainny/raw/13-35-659.png}{271 699 1313 62} &
\myimage{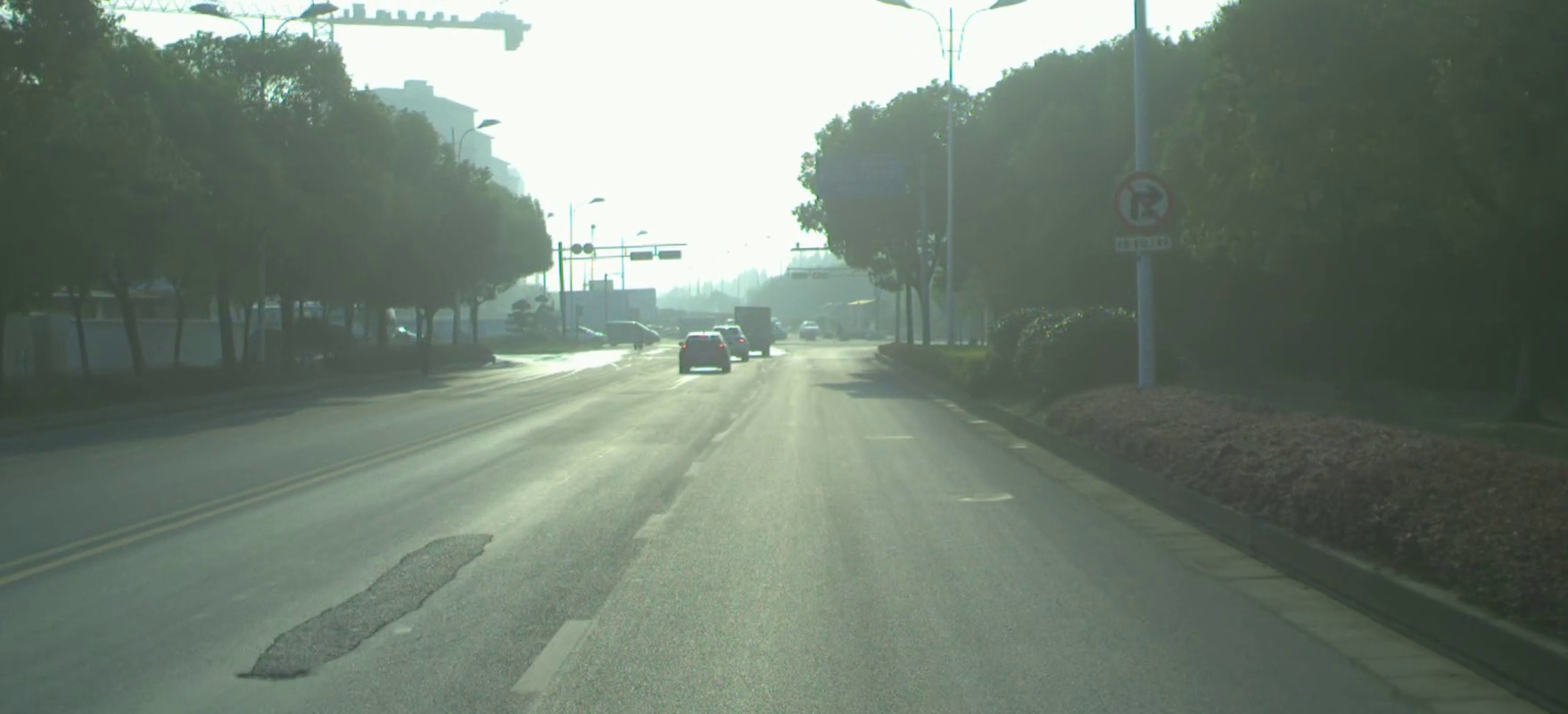}{./all-models-generalization/foggy/raw/06-39-171.png}{1239 438 426 108} &
\myimage{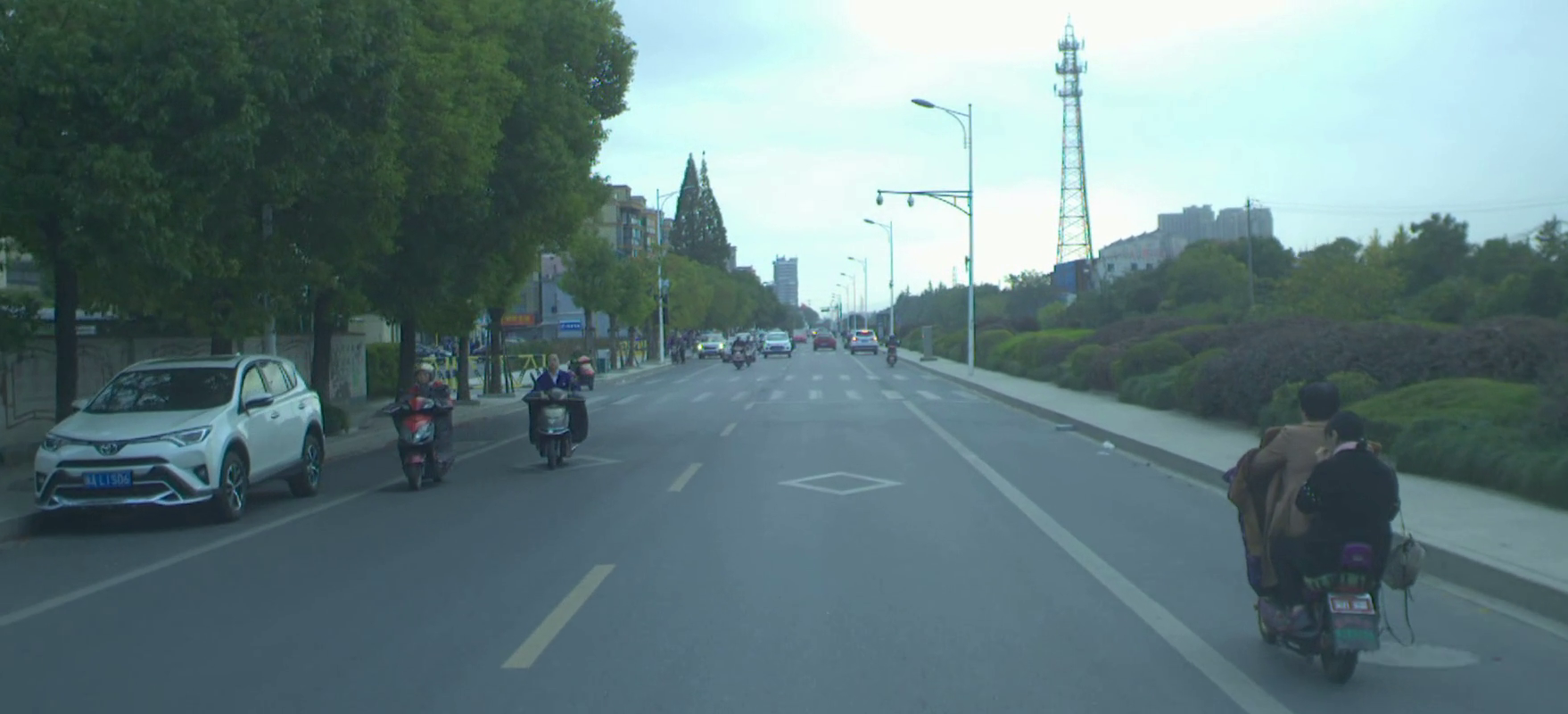}{./all-models-generalization/cloudy/raw/03-15-641.png}{1360 35 152 425} &
\myimage{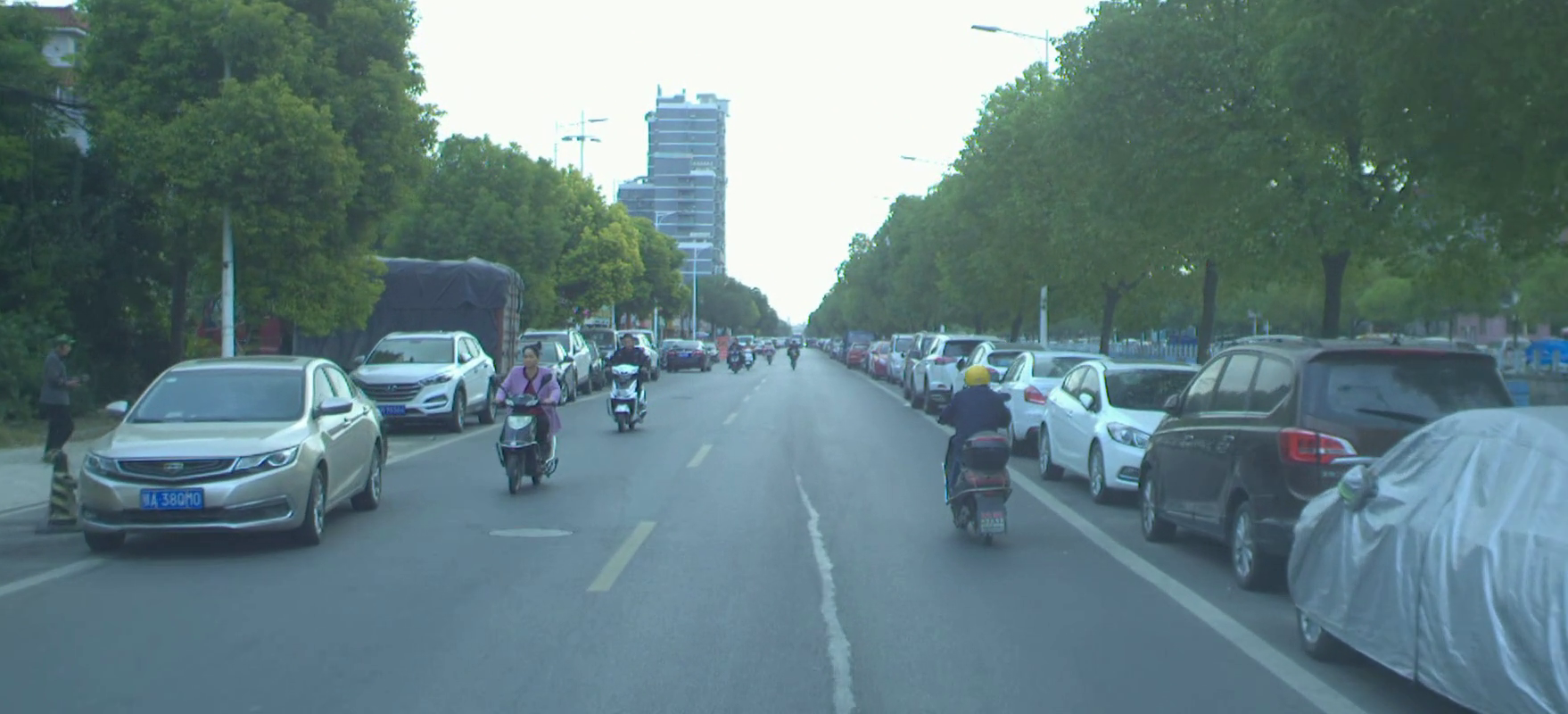}{./all-models-generalization/cloudy/raw/05-39-923.png}{34 282 1646 364} \\
\hline
\myimage{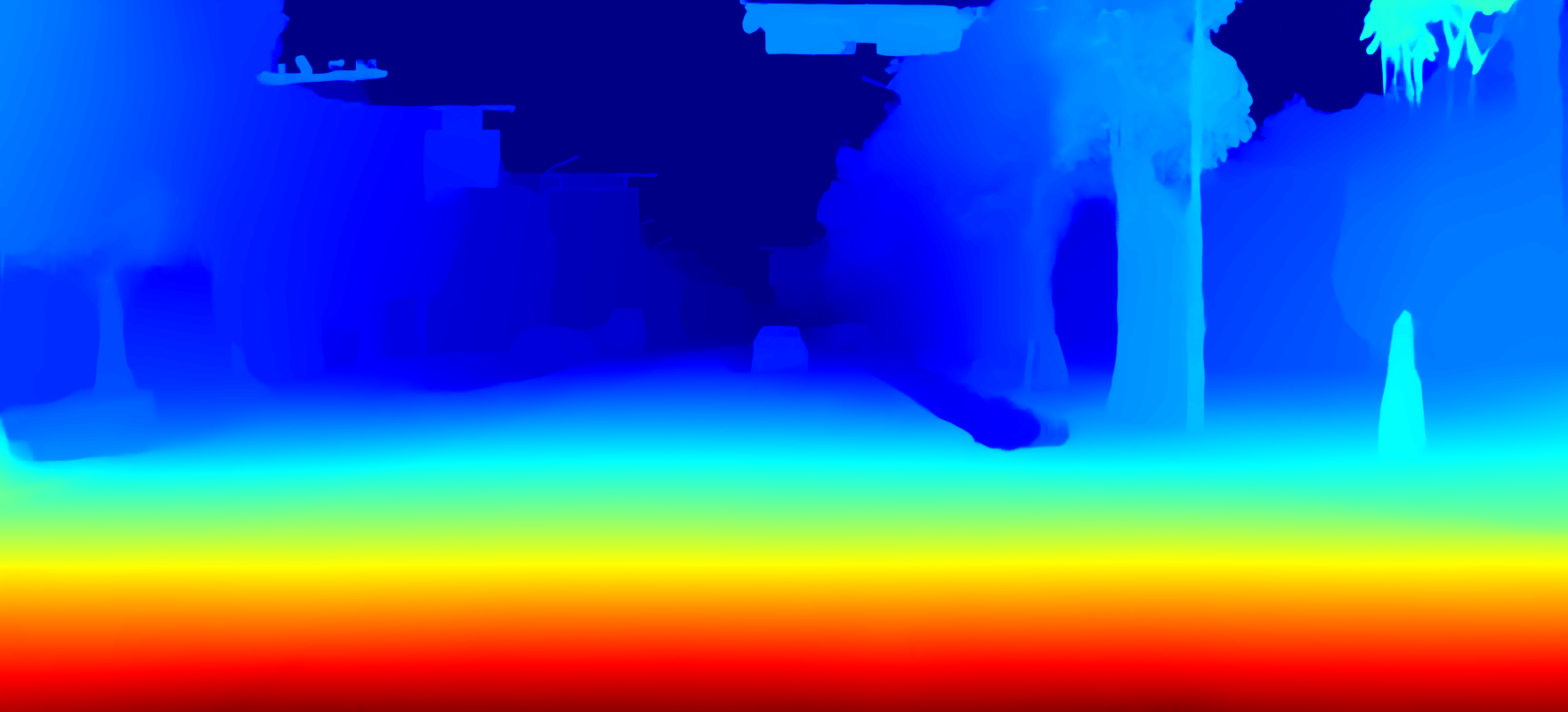}{./all-models-generalization/rainny/RAFT-Stereo-main-1/13-35-659.png_disp_pred.png}{271 699 1313 62} &
\myimage{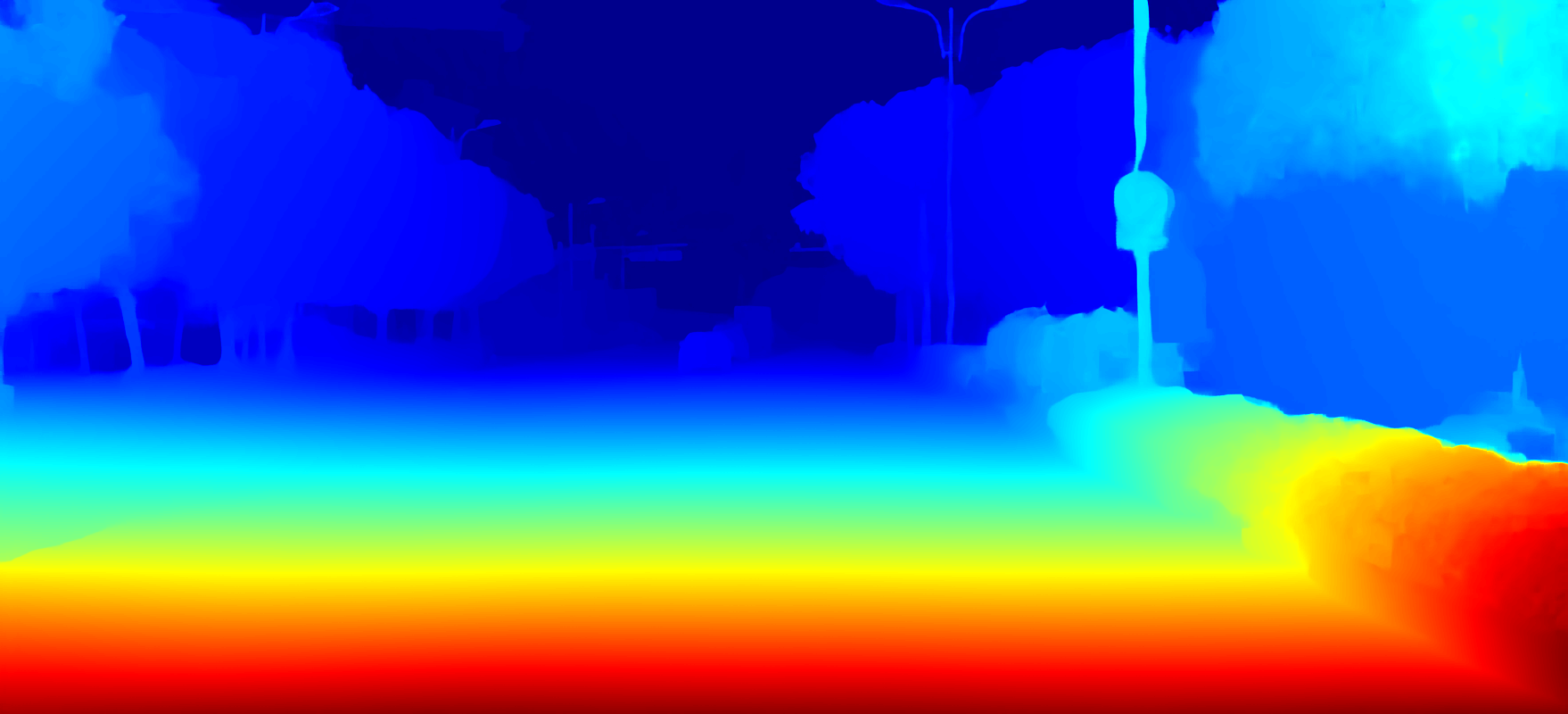}{./all-models-generalization/foggy/RAFT-Stereo-main-1/06-39-171.png_disp_pred.png}{1239 438 426 108} &
\myimage{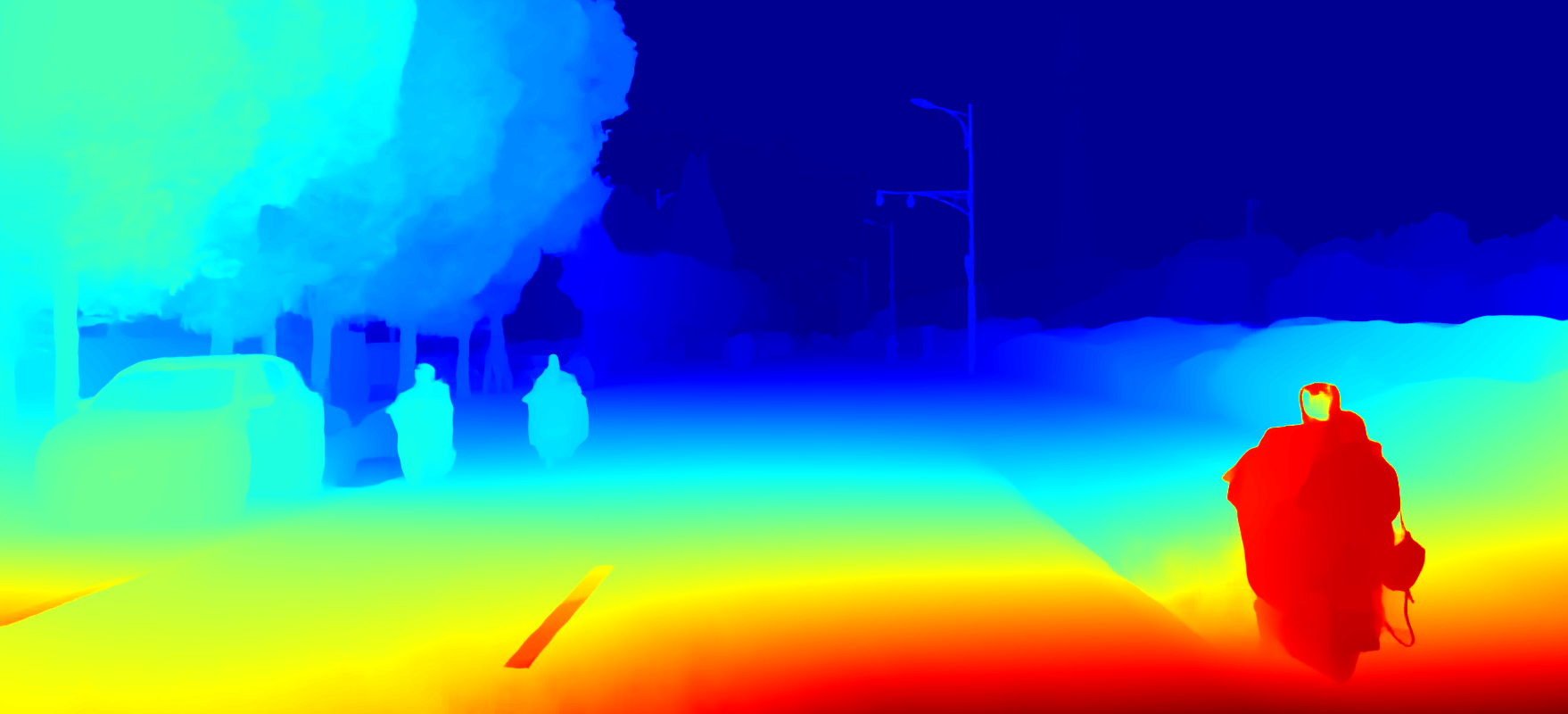}{./all-models-generalization/cloudy/RAFT-Stereo-main-1/03-15-641.png_disp_pred.png}{1360 35 152 425} &
\myimage{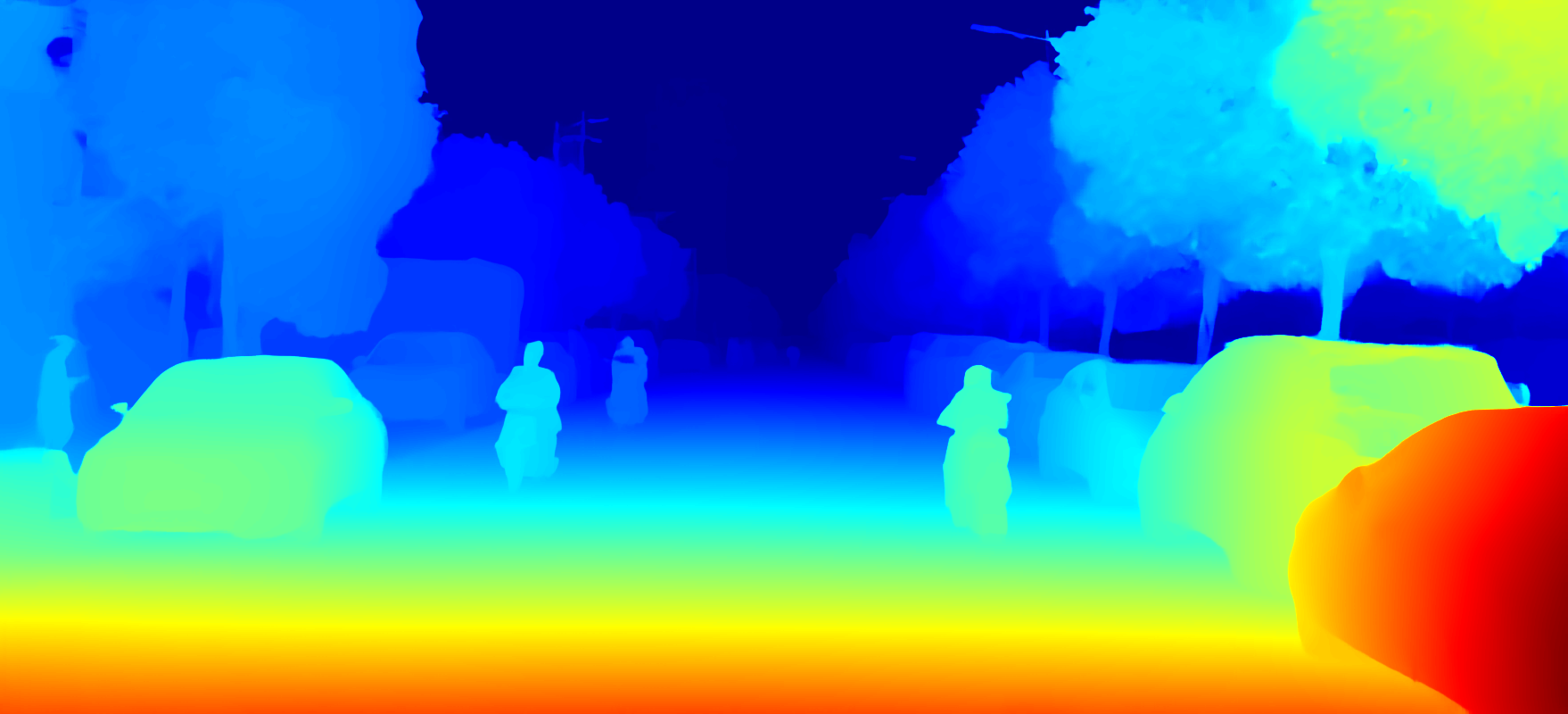}{./all-models-generalization/cloudy/RAFT-Stereo-main-1/05-39-923.png_disp_pred.png}{34 282 1646 364} \\
\myimage{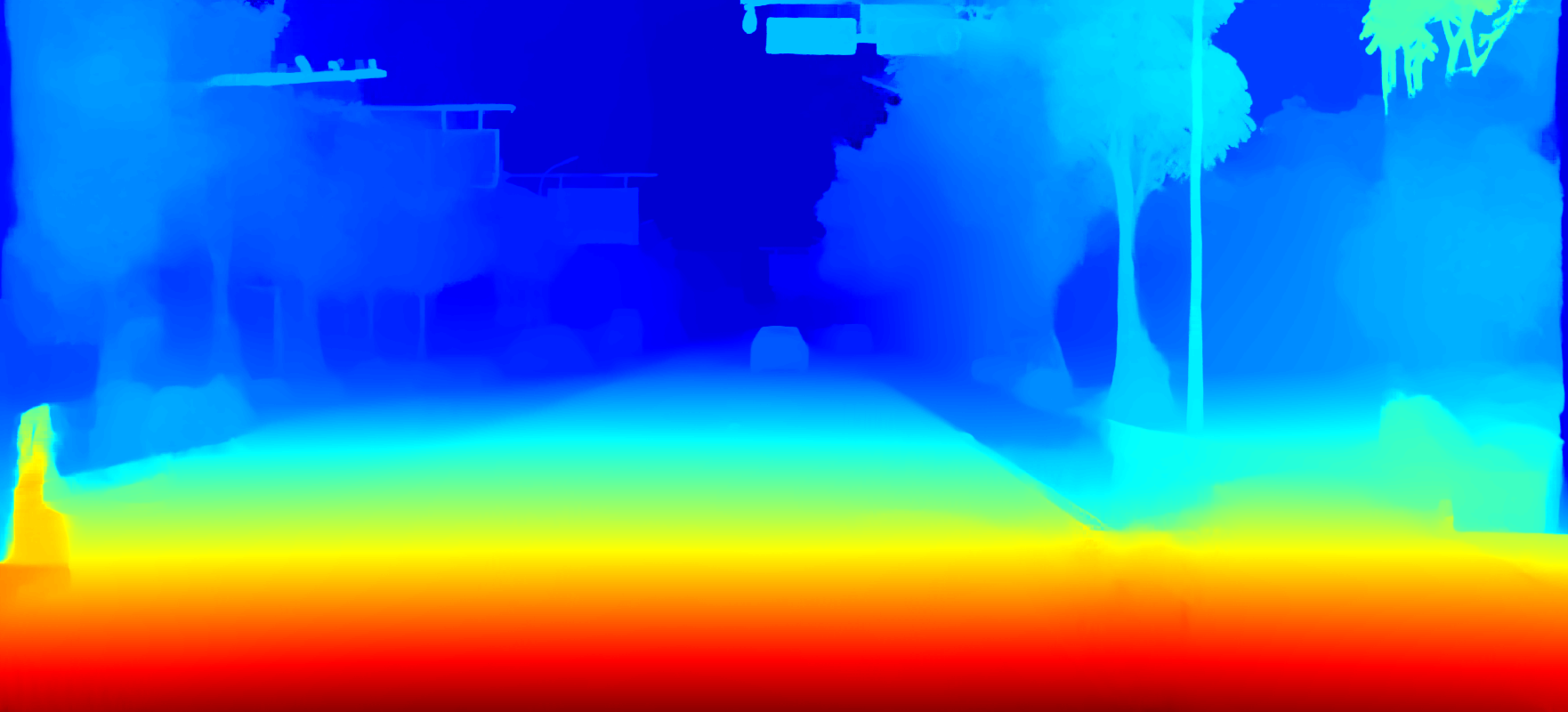}{./all-models-generalization/rainny/IGEV-Stereo/13-35-659.png_disp_pred.png}{271 699 1313 62} &
\myimage{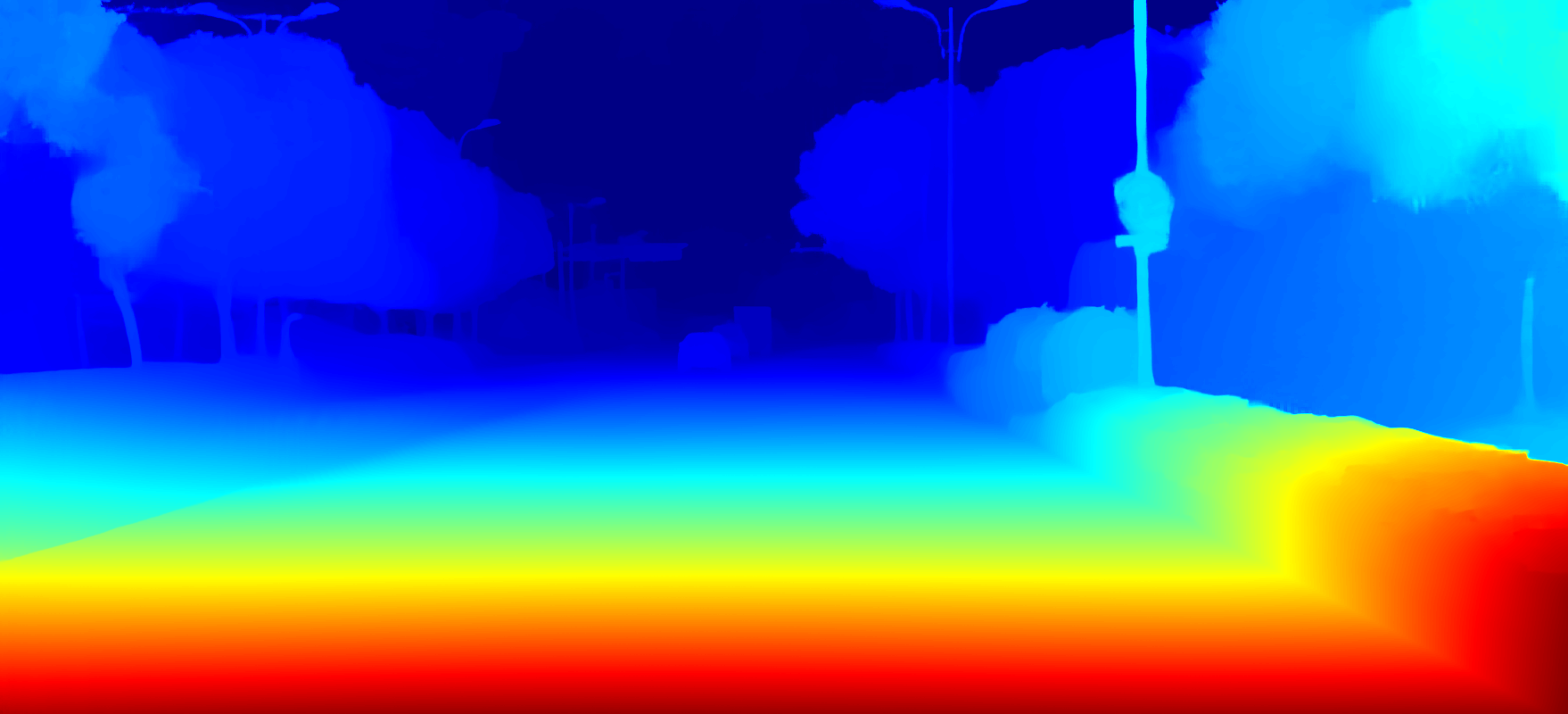}{./all-models-generalization/foggy/IGEV-Stereo/06-39-171.png_disp_pred.png}{1239 438 426 108} &
\myimage{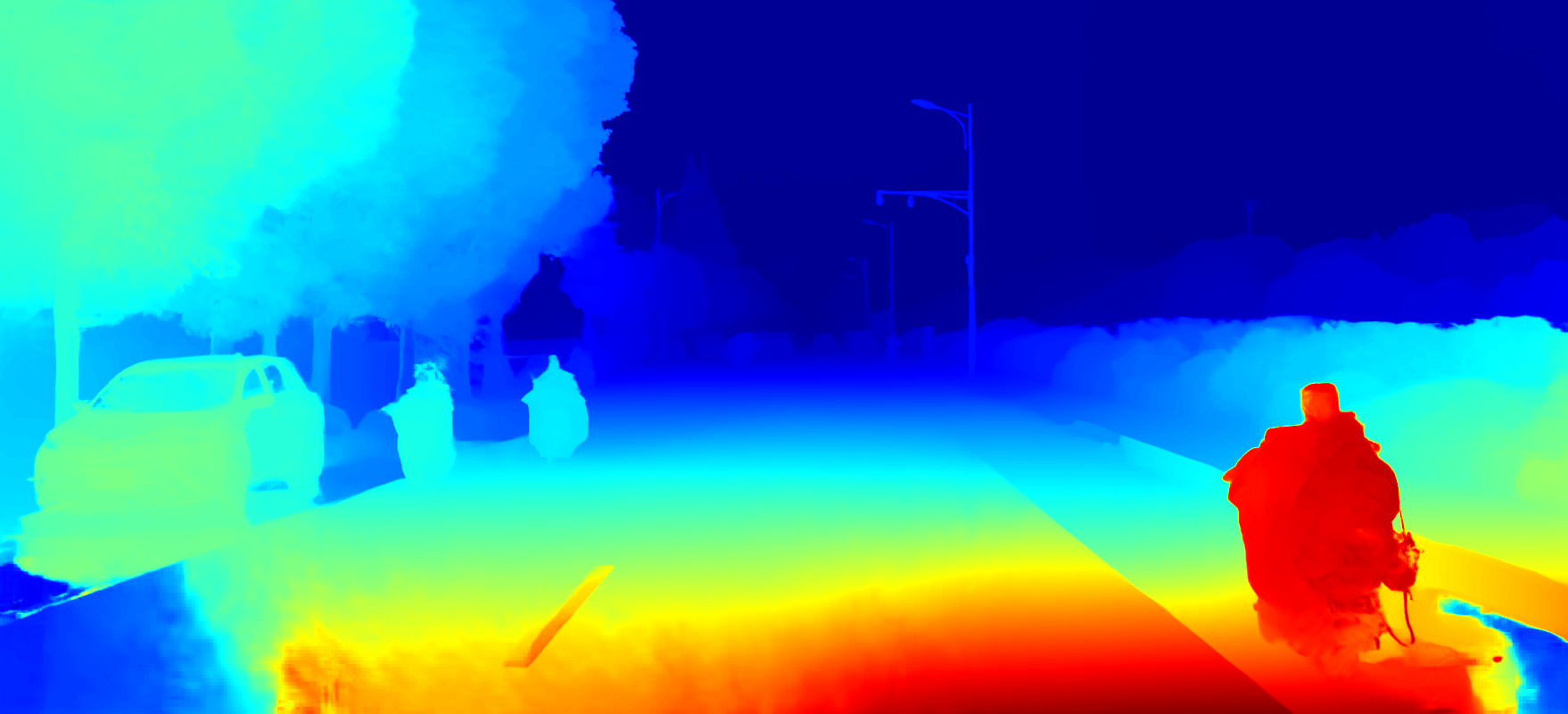}{./all-models-generalization/cloudy/IGEV-Stereo/03-15-641.png_disp_pred.png}{1360 35 152 425} &
\myimage{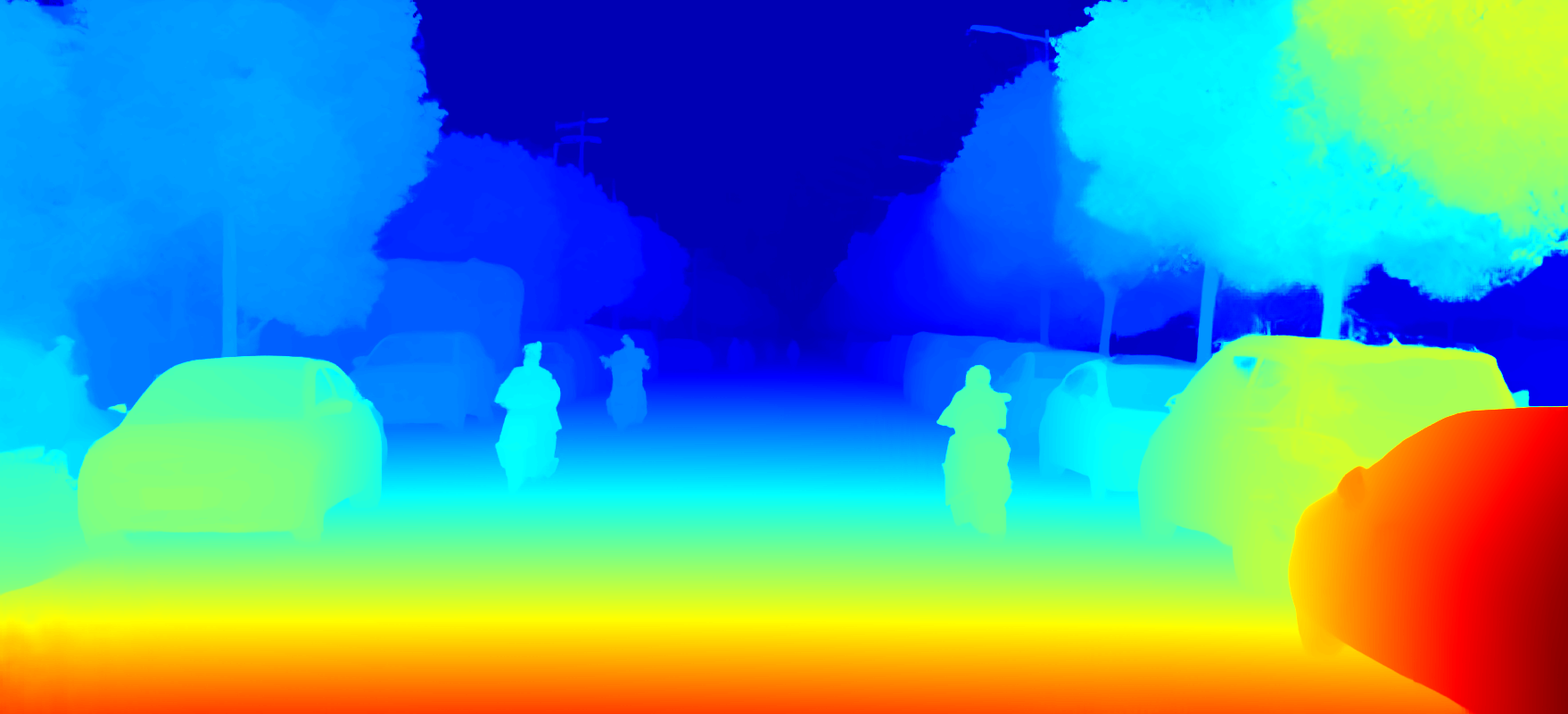}{./all-models-generalization/cloudy/IGEV-Stereo/05-39-923.png_disp_pred.png}{34 282 1646 364} \\
\myimage{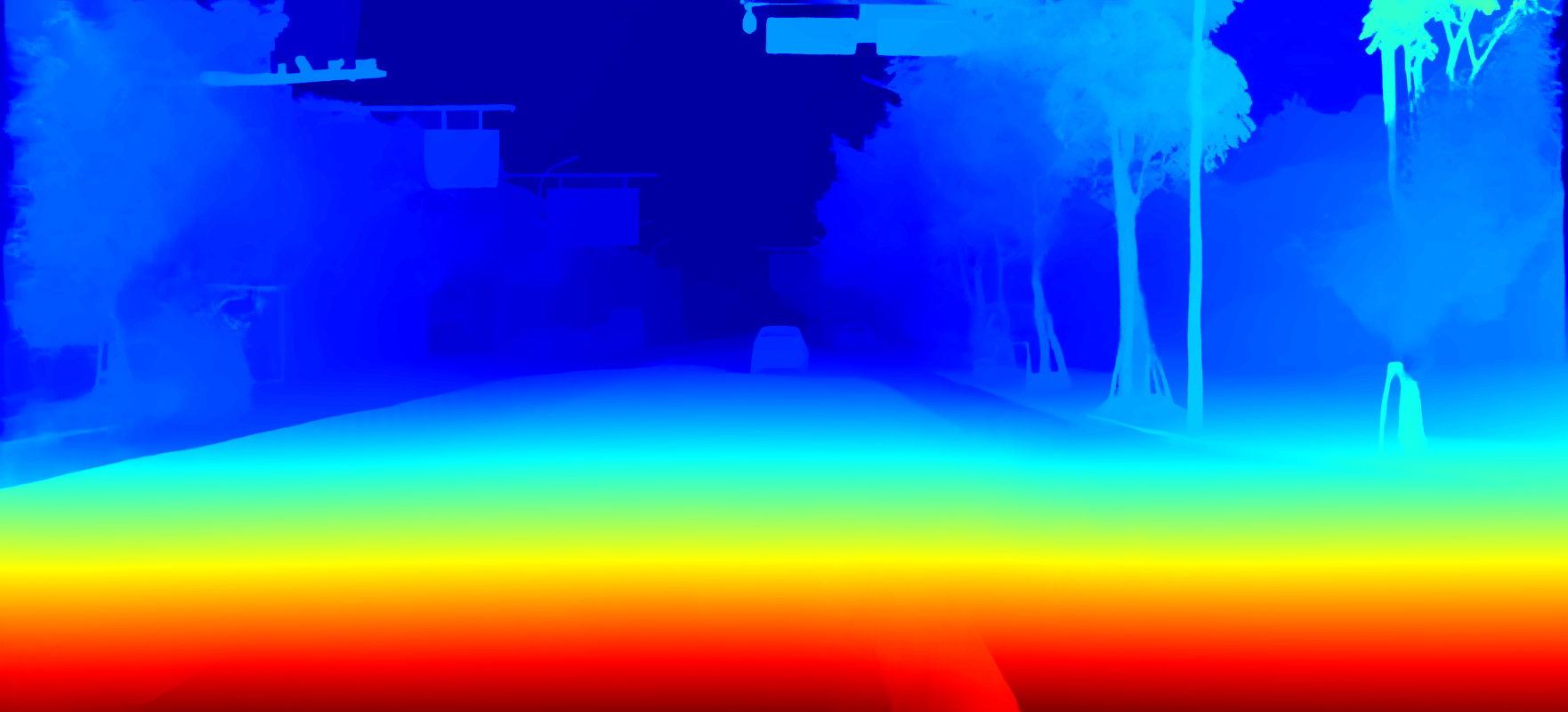}{./all-models-generalization/rainny/Selective-IGEV/13-35-659.png_disp_pred.png}{271 699 1313 62} &
\myimage{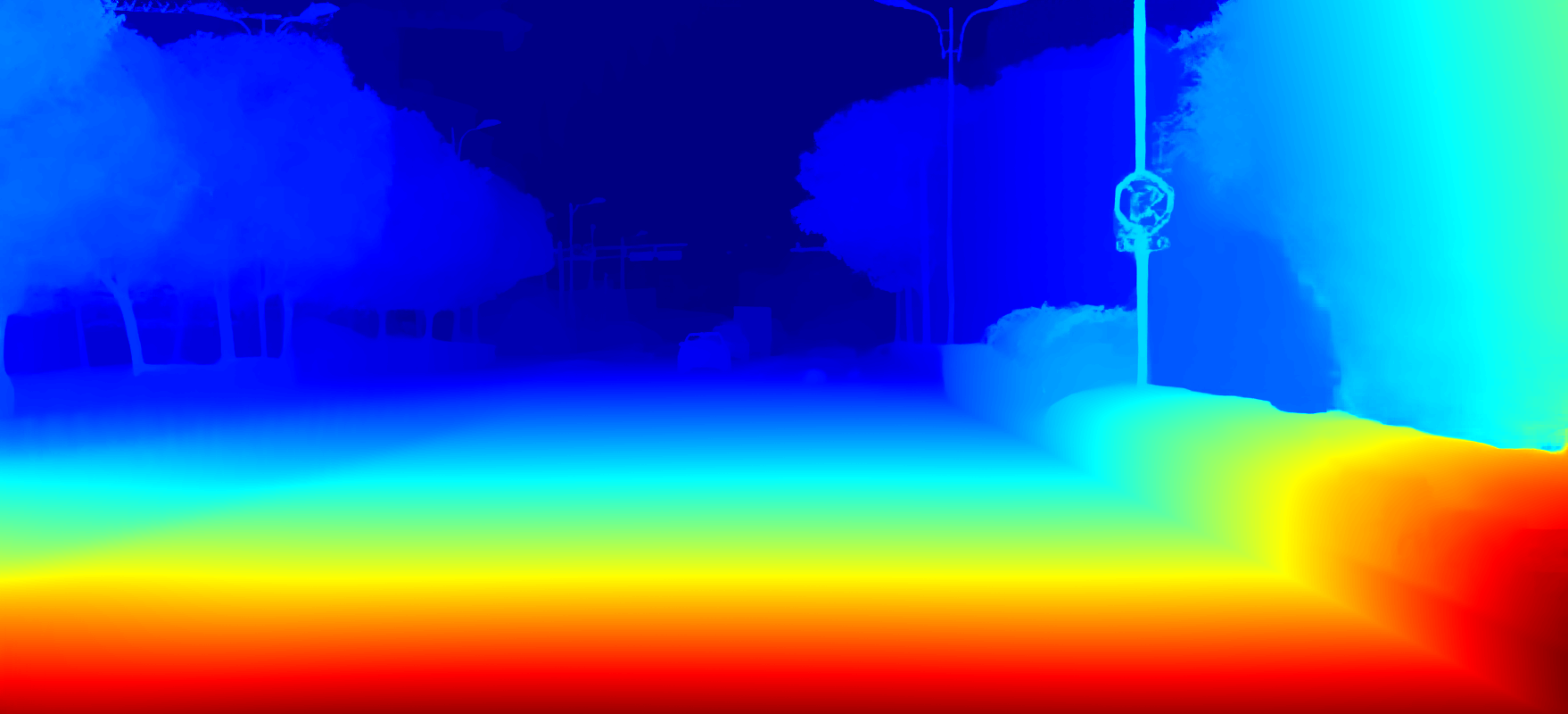}{./all-models-generalization/foggy/Selective-IGEV/06-39-171.png_disp_pred.png}{1239 438 426 108} &
\myimage{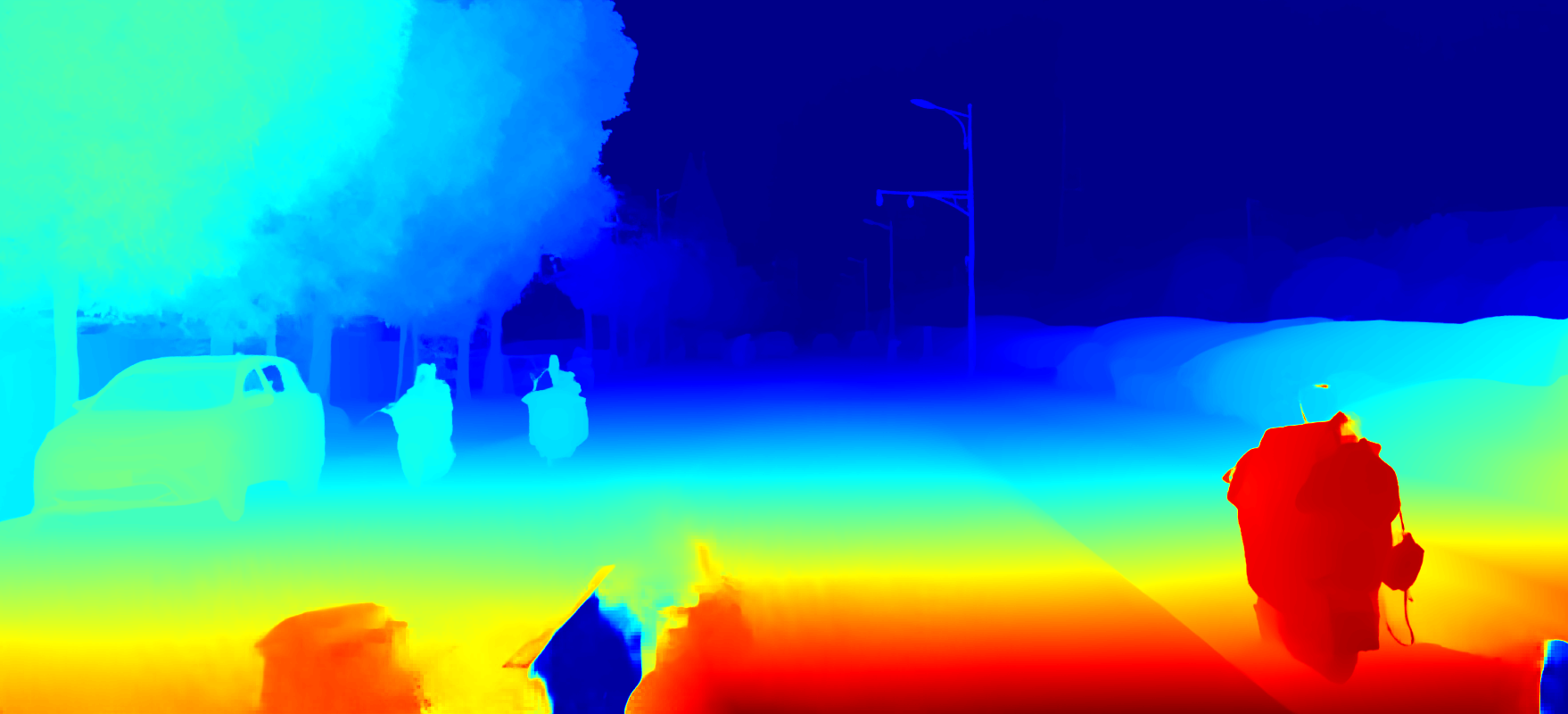}{./all-models-generalization/cloudy/Selective-IGEV/03-15-641.png_disp_pred.png}{1360 35 152 425} &
\myimage{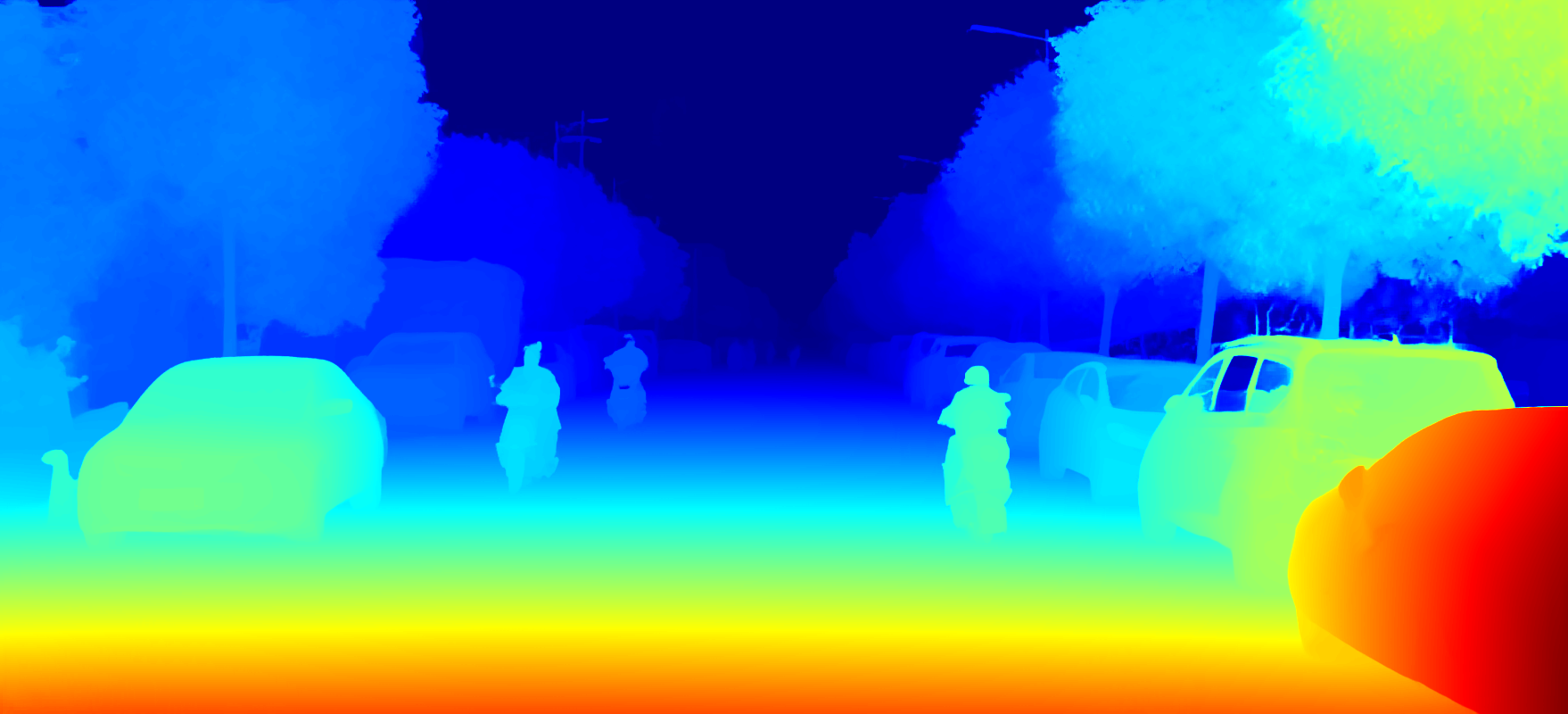}{./all-models-generalization/cloudy/Selective-IGEV/05-39-923.png_disp_pred.png}{34 282 1646 364} \\
\myimage{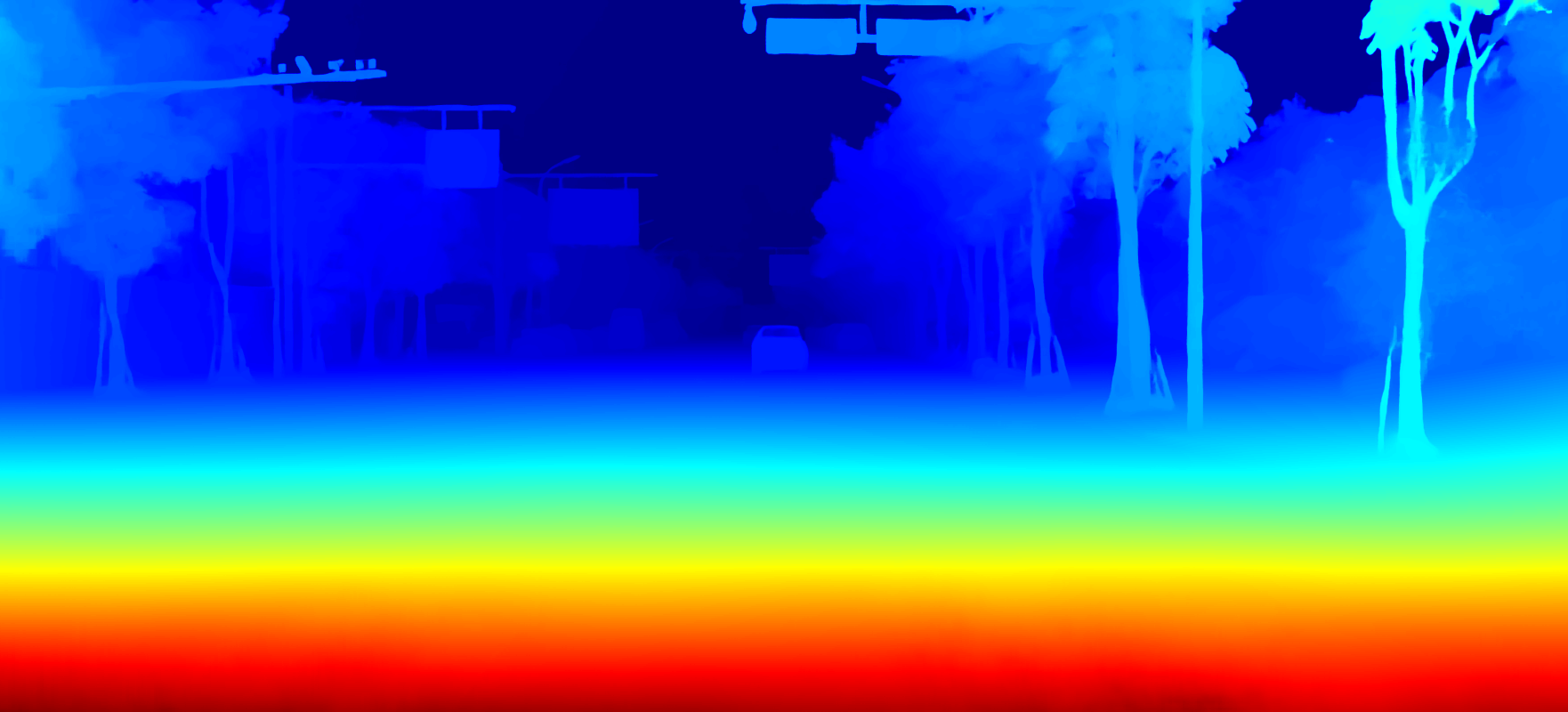}{./all-models-generalization/rainny/DEFOM-Stereo/13-35-659.png_disp_pred.png}{271 699 1313 62} &
\myimage{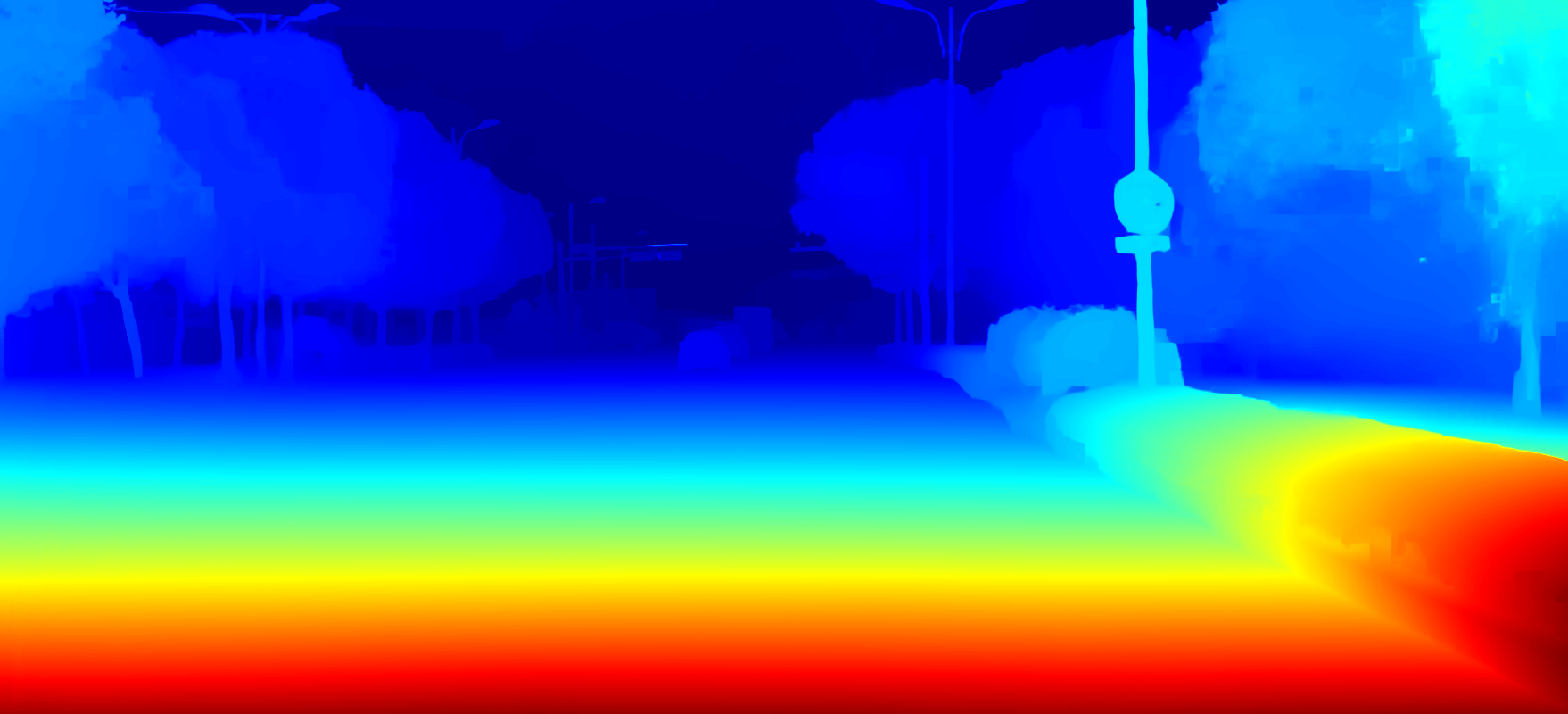}{./all-models-generalization/foggy/DEFOM-Stereo/06-39-171.png_disp_pred.png}{1239 438 426 108} &
\myimage{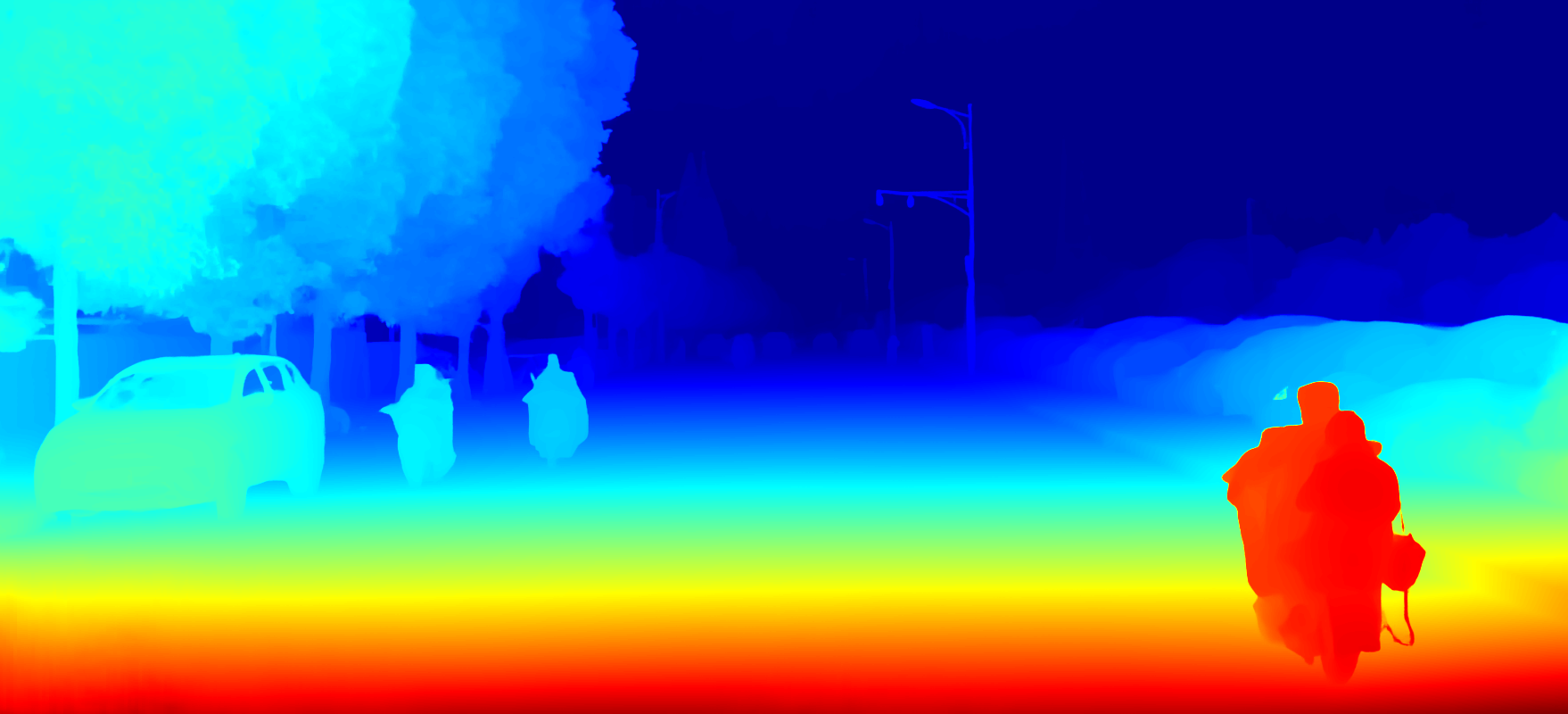}{./all-models-generalization/cloudy/DEFOM-Stereo/03-15-641.png_disp_pred.png}{1360 35 152 425} &
\myimage{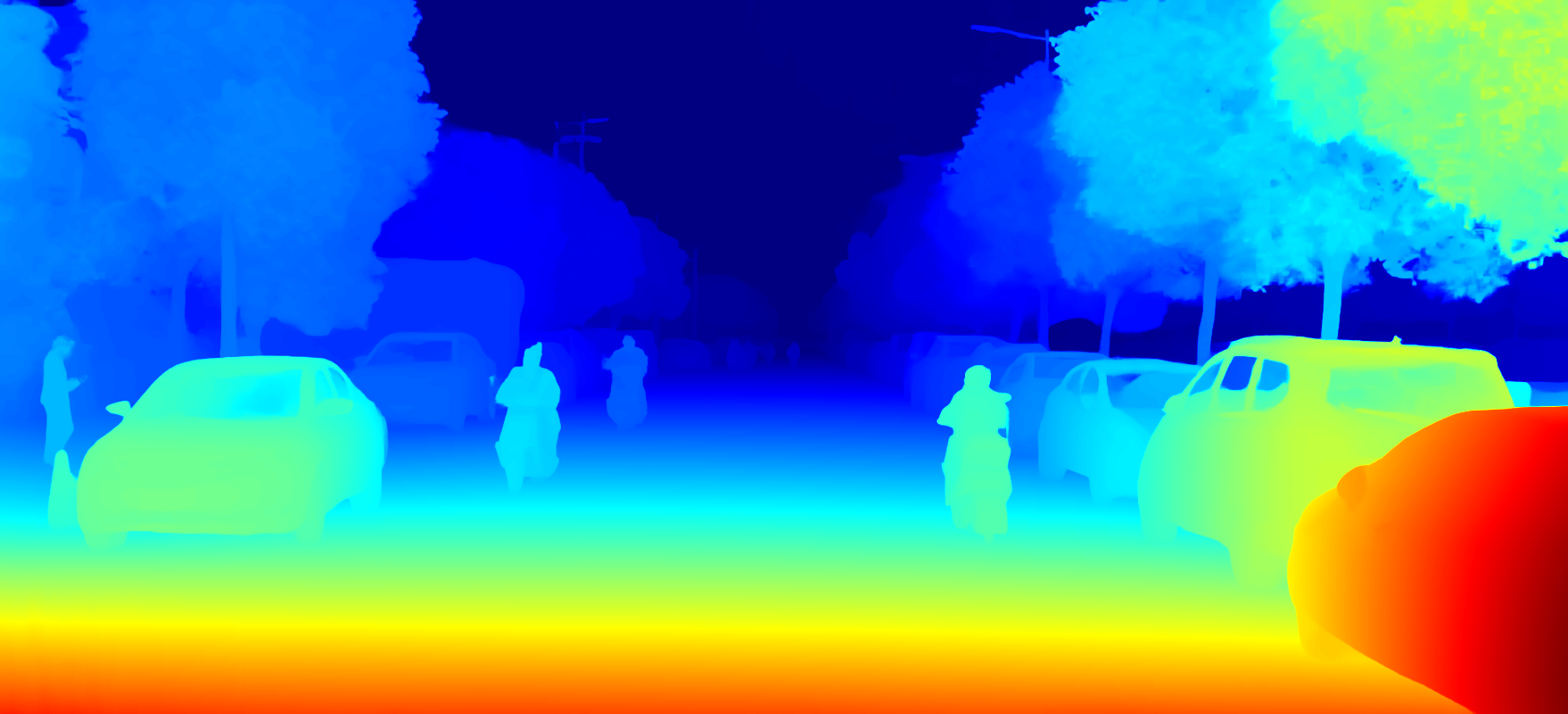}{./all-models-generalization/cloudy/DEFOM-Stereo/05-39-923.png_disp_pred.png}{34 282 1646 364} \\
\myimage{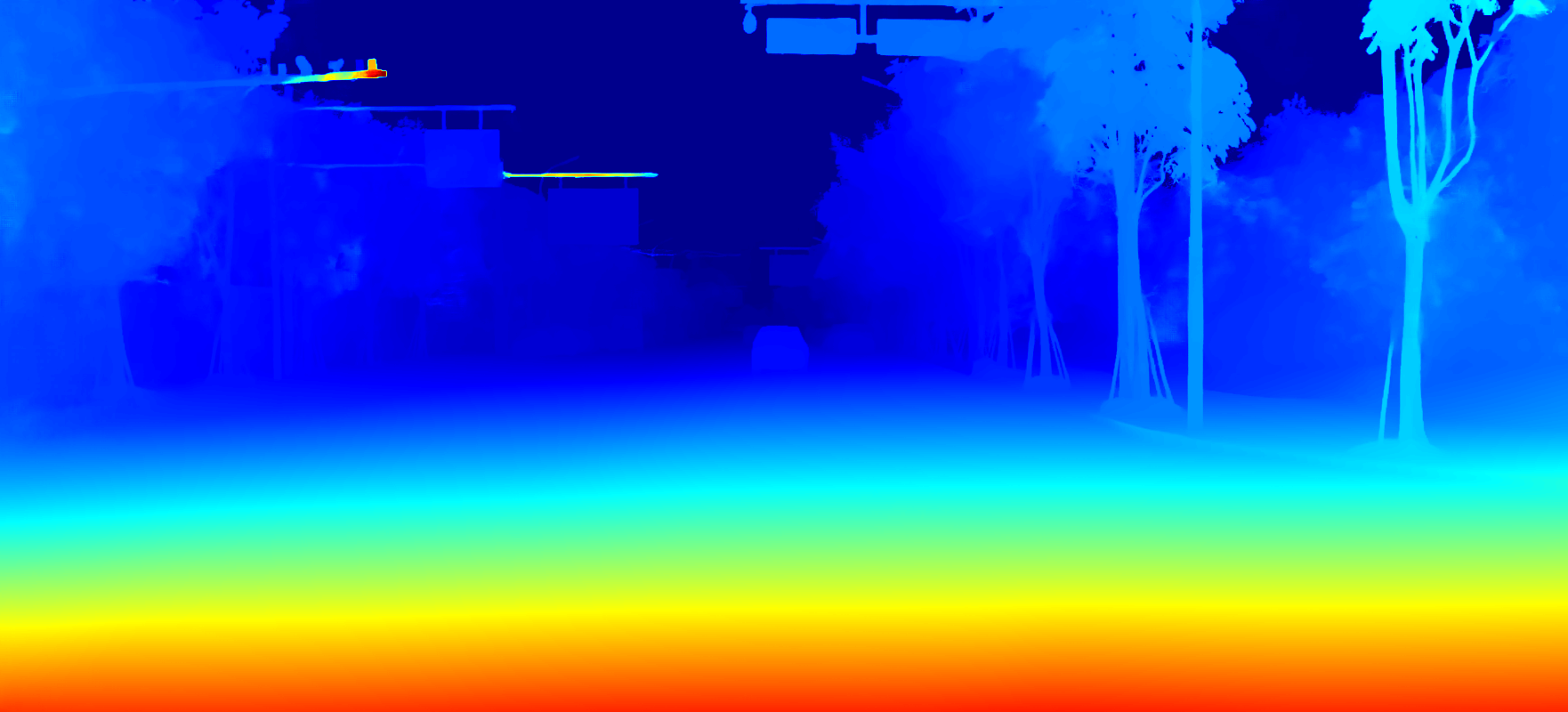}{./all-models-generalization/rainny/MonSter/13-35-659.png_disp_pred.png}{271 699 1313 62} &
\myimage{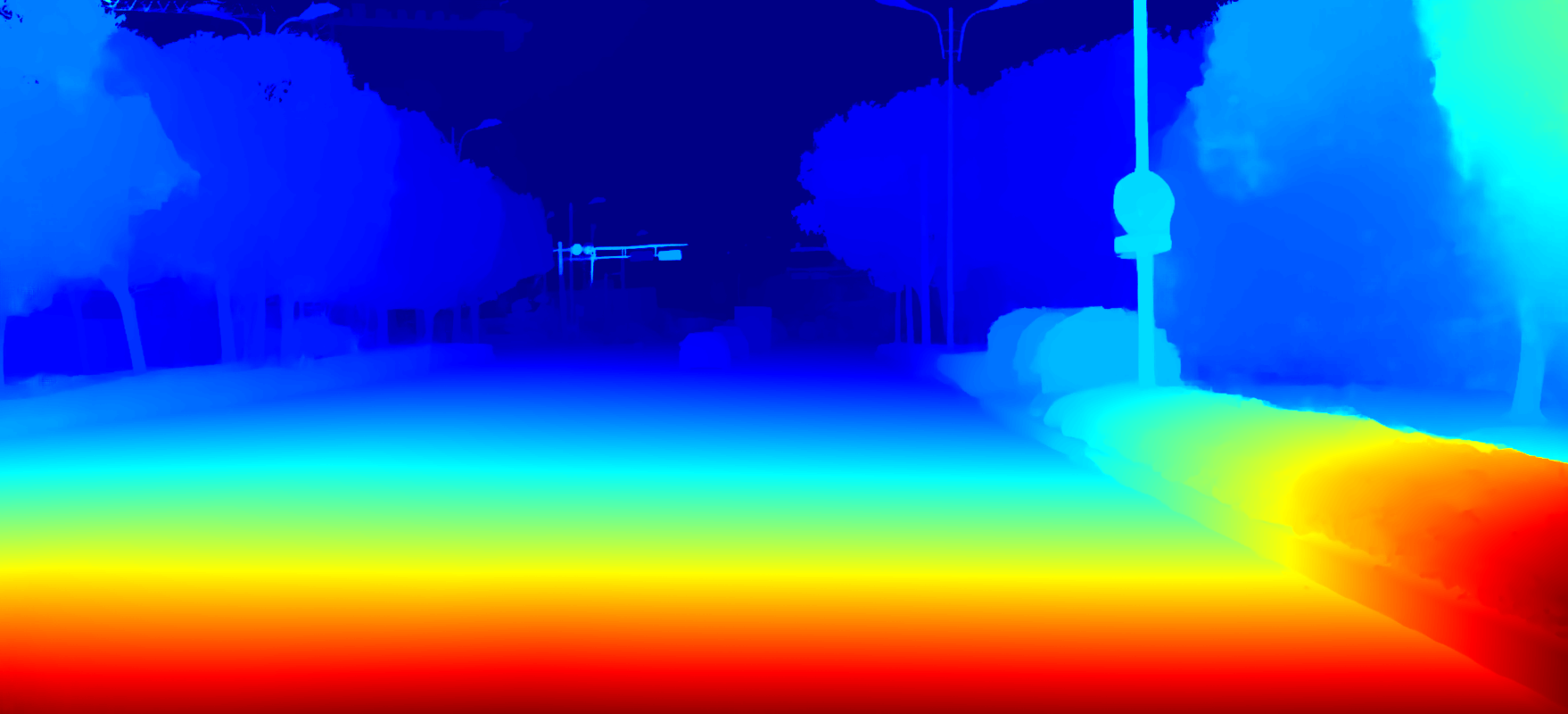}{./all-models-generalization/foggy/MonSter/06-39-171.png_disp_pred.png}{1239 438 426 108} &
\myimage{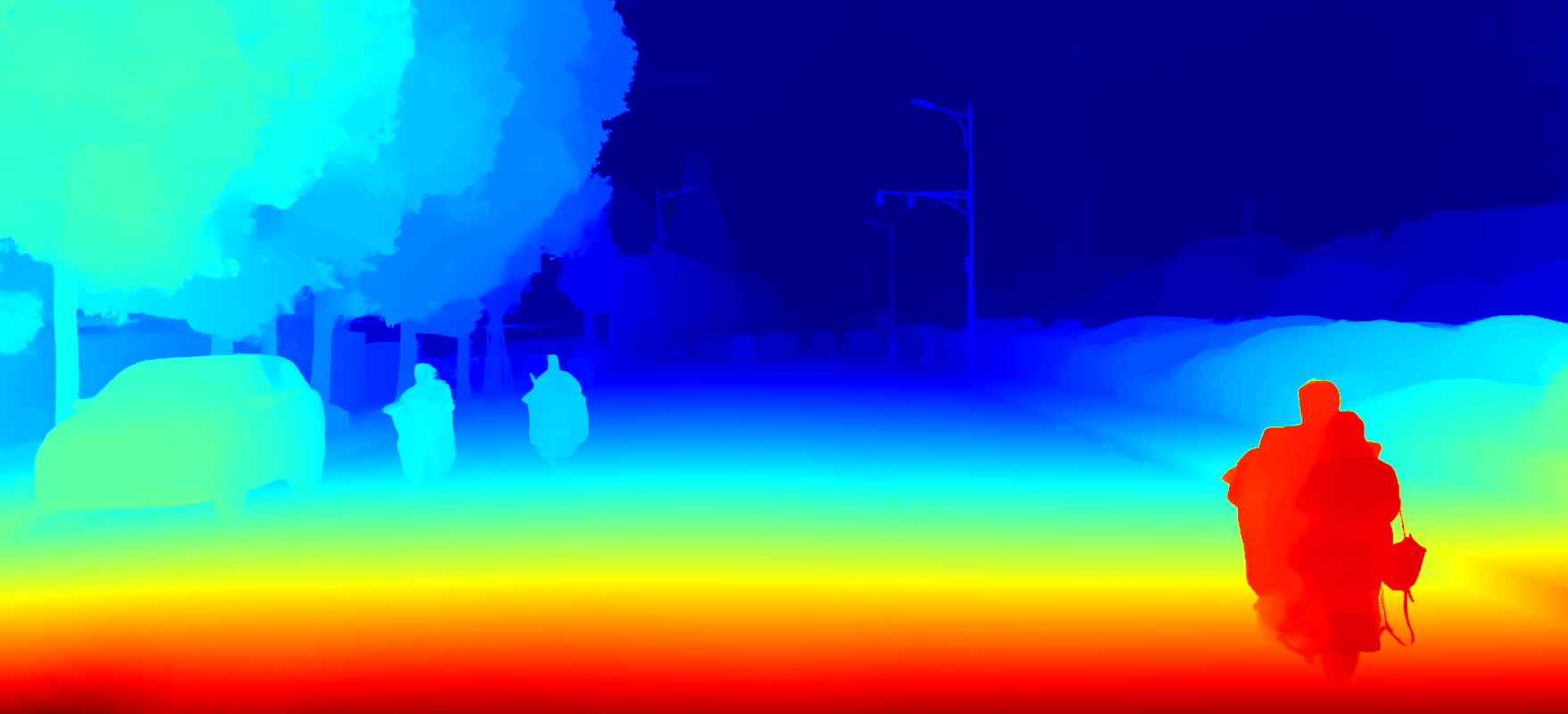}{./all-models-generalization/cloudy/MonSter/03-15-641.png_disp_pred.png}{1360 35 152 425} &
\myimage{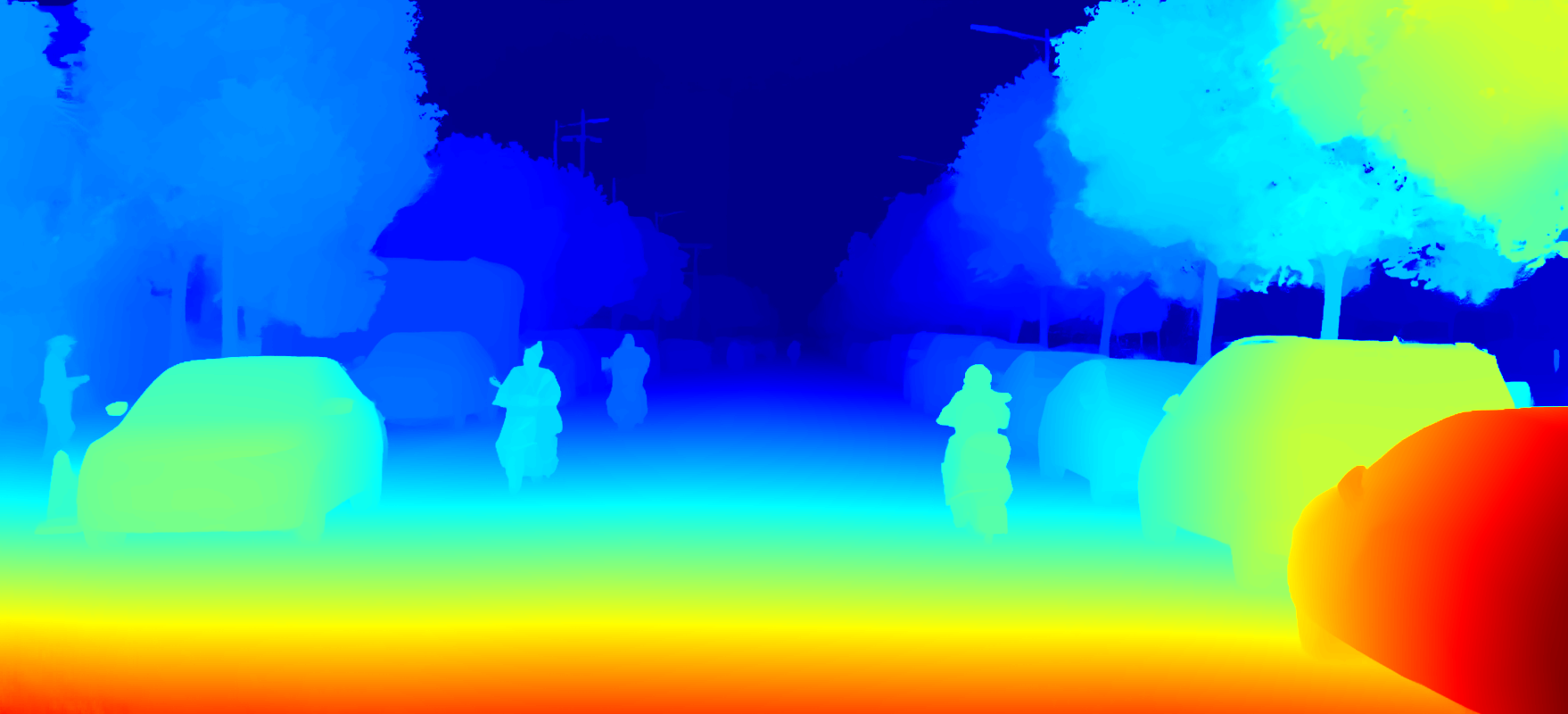}{./all-models-generalization/cloudy/MonSter/05-39-923.png_disp_pred.png}{34 282 1646 364} \\
\myimage{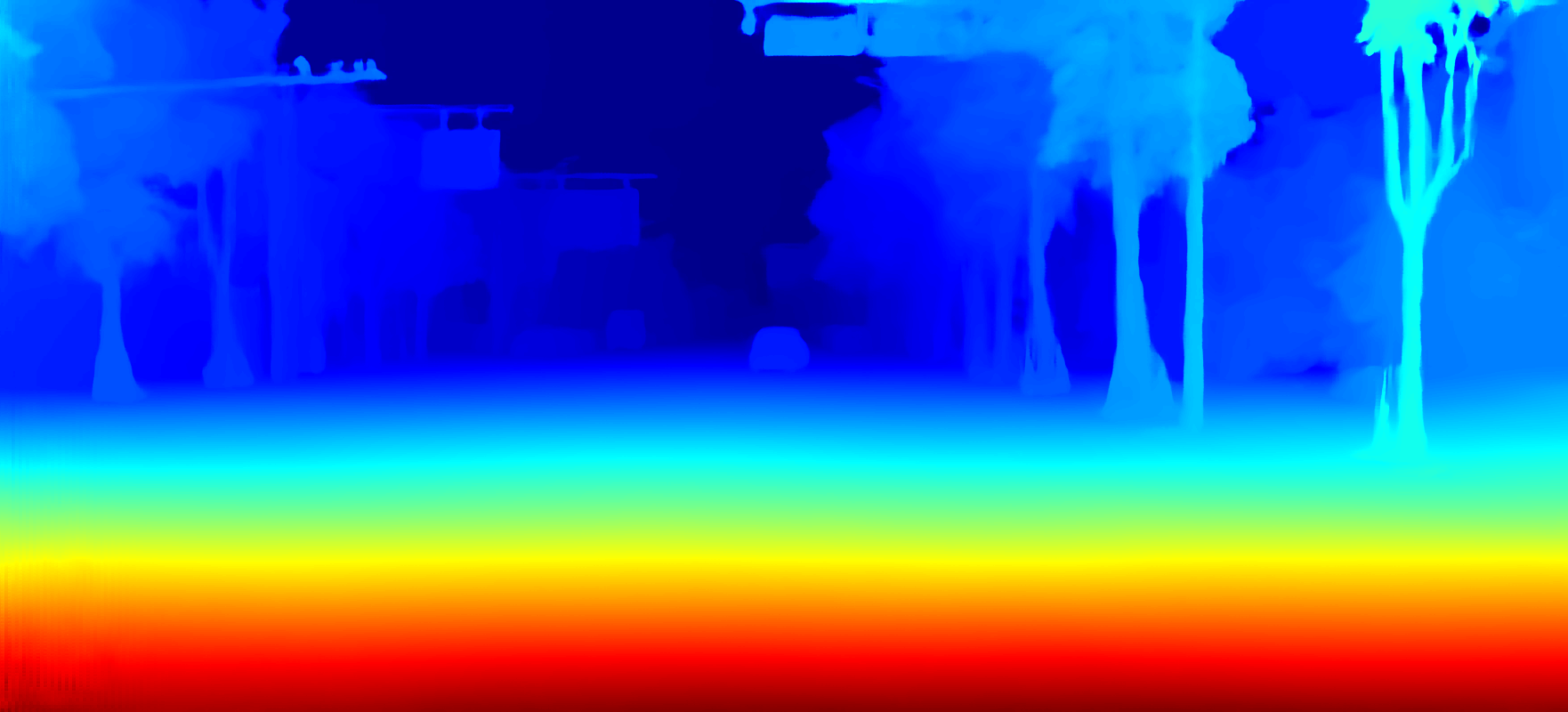}{./all-models-generalization/rainny/Diving-into-the-Fusion-of-Monocular-Priors-for-Generalized-Stereo-Matching/13-35-659.png_disp_pred.png}{271 699 1313 62} &
\myimage{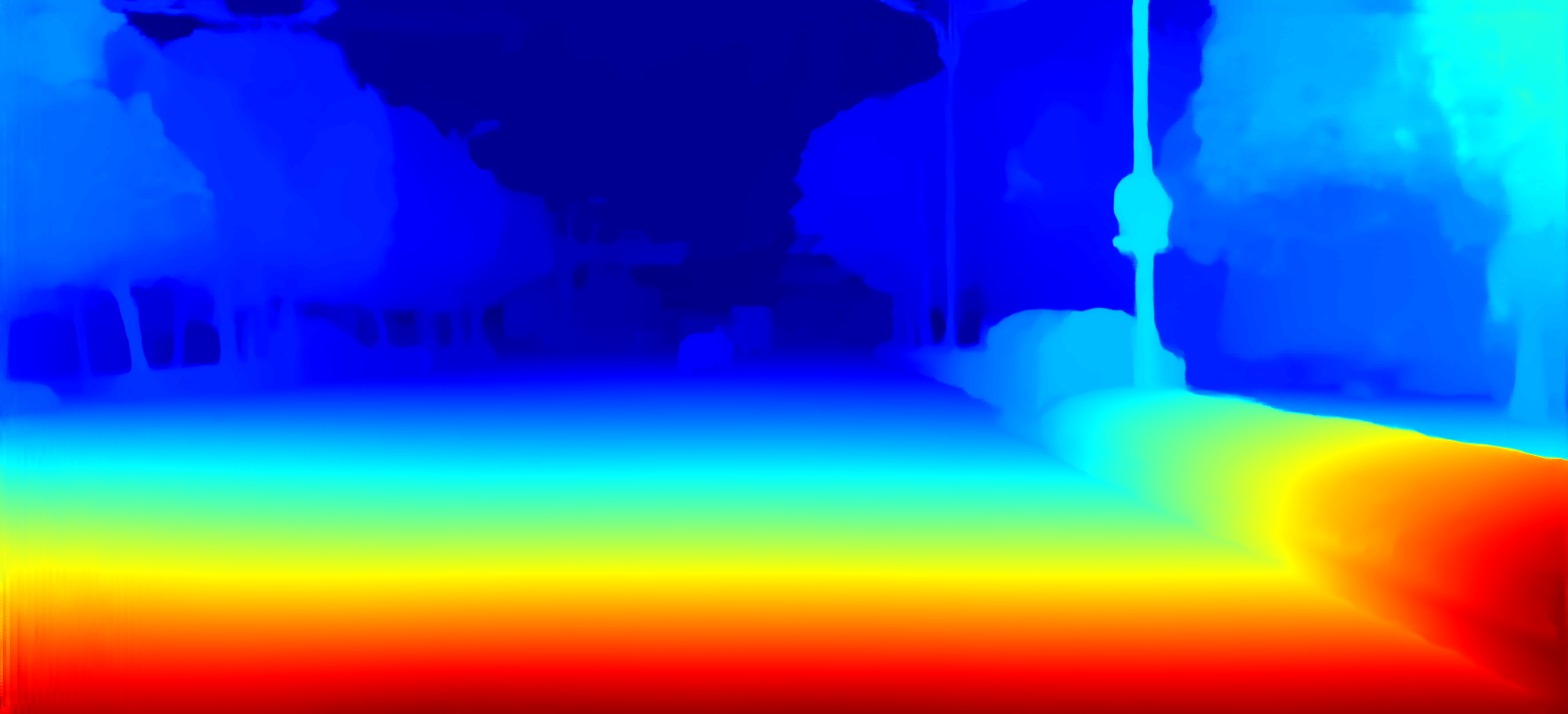}{./all-models-generalization/foggy/Diving-into-the-Fusion-of-Monocular-Priors-for-Generalized-Stereo-Matching/06-39-171.png_disp_pred.png}{1239 438 426 108} &
\myimage{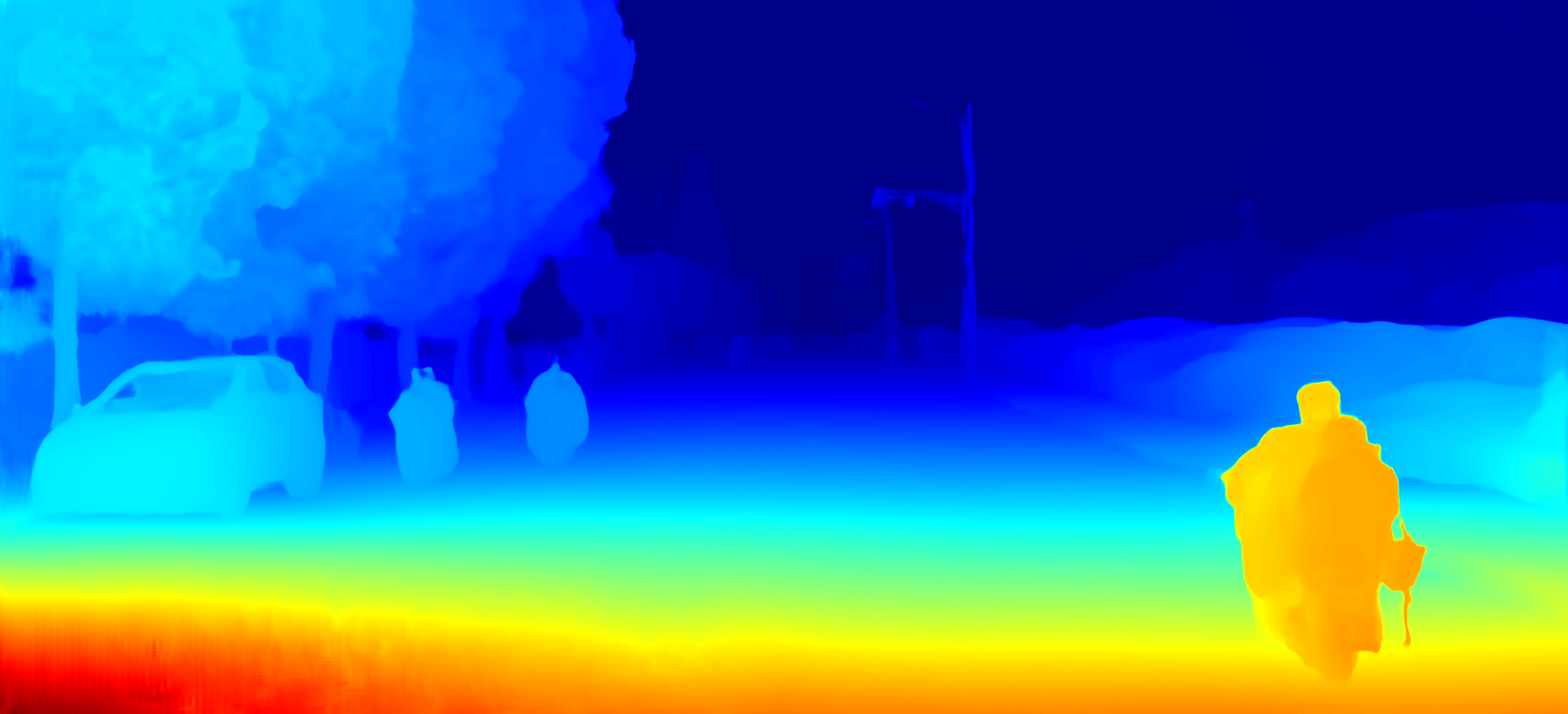}{./all-models-generalization/cloudy/Diving-into-the-Fusion-of-Monocular-Priors-for-Generalized-Stereo-Matching/03-15-641.png_disp_pred.png}{1360 35 152 425} &
\myimage{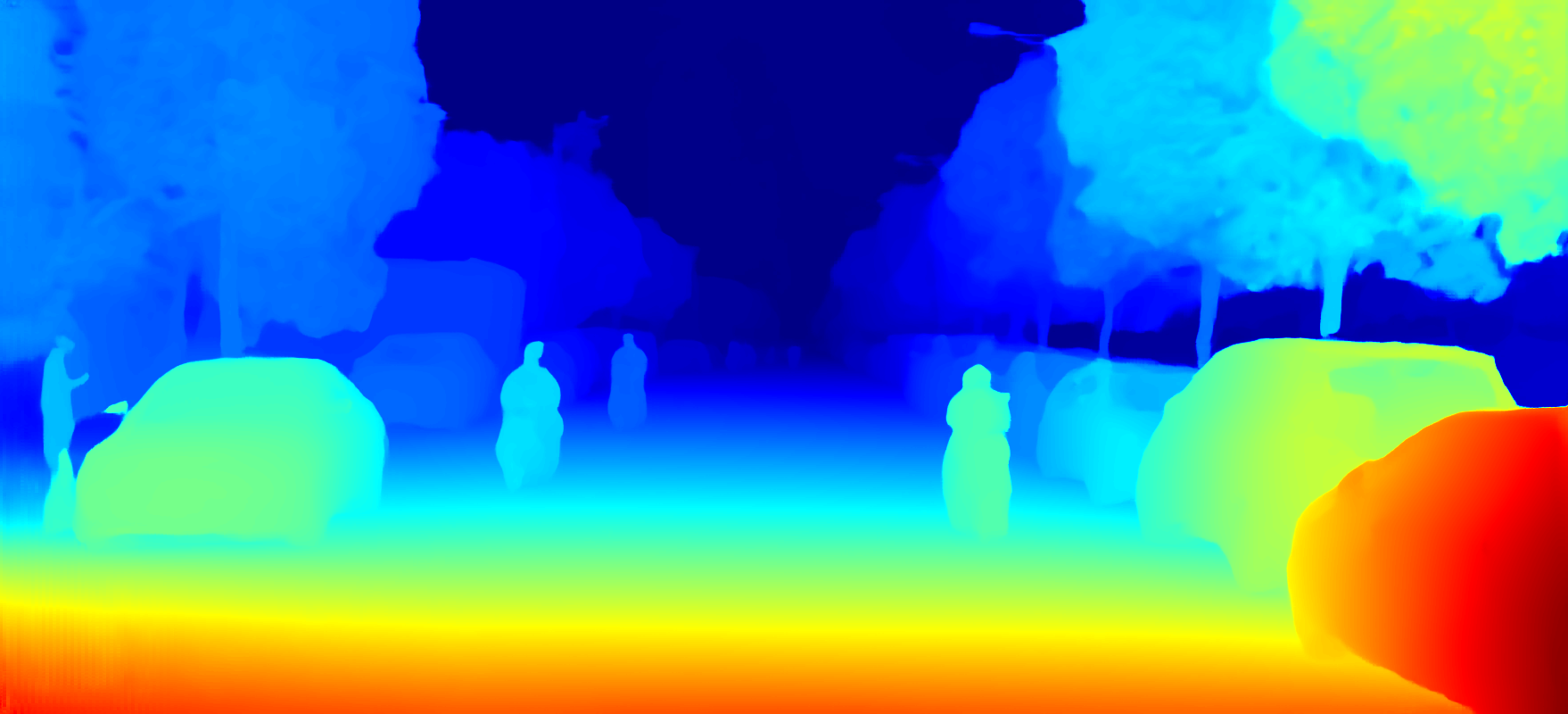}{./all-models-generalization/cloudy/Diving-into-the-Fusion-of-Monocular-Priors-for-Generalized-Stereo-Matching/05-39-923.png_disp_pred.png}{34 282 1646 364} \\
\myimage{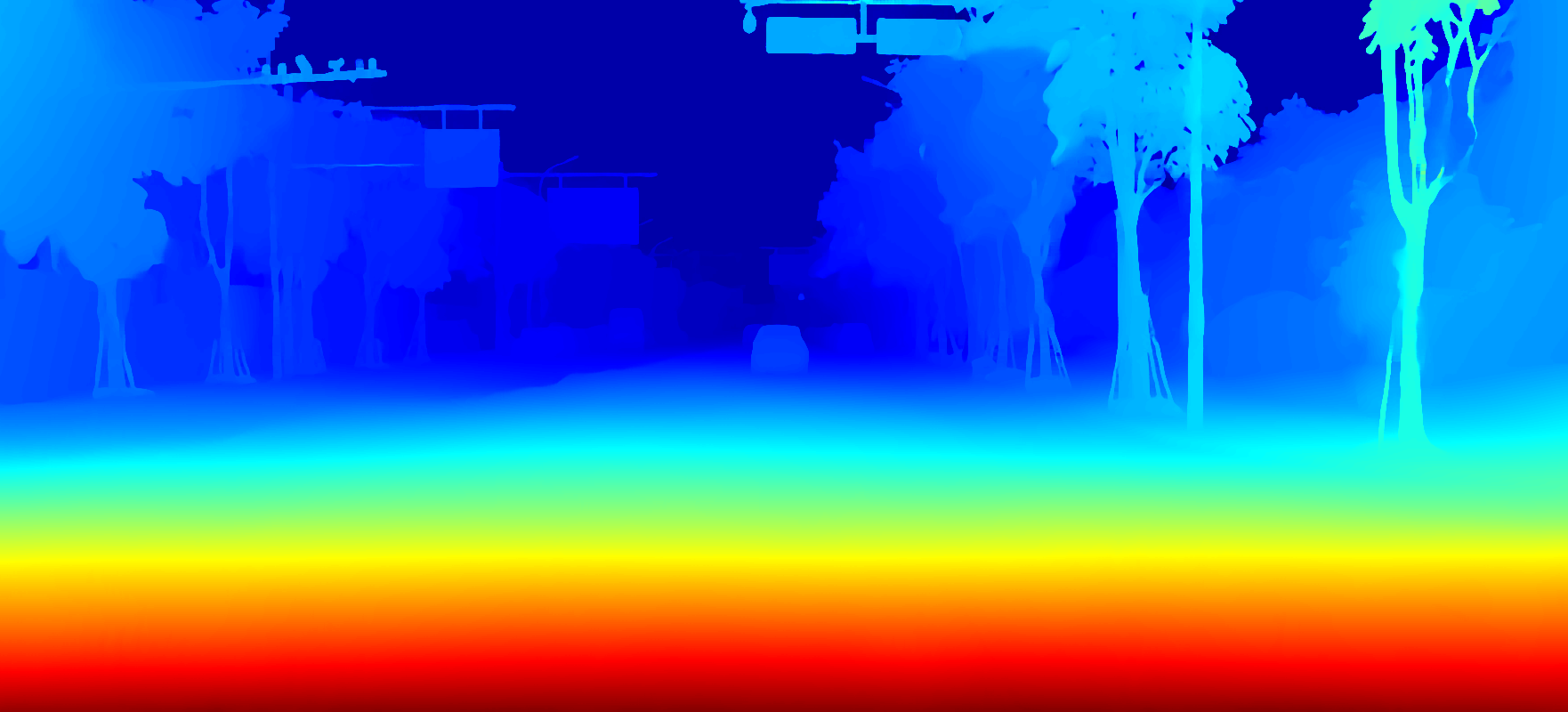}{./all-models-generalization/rainny/debug/13-35-659.png_disp_pred.png}{271 699 1313 62} &
\myimage{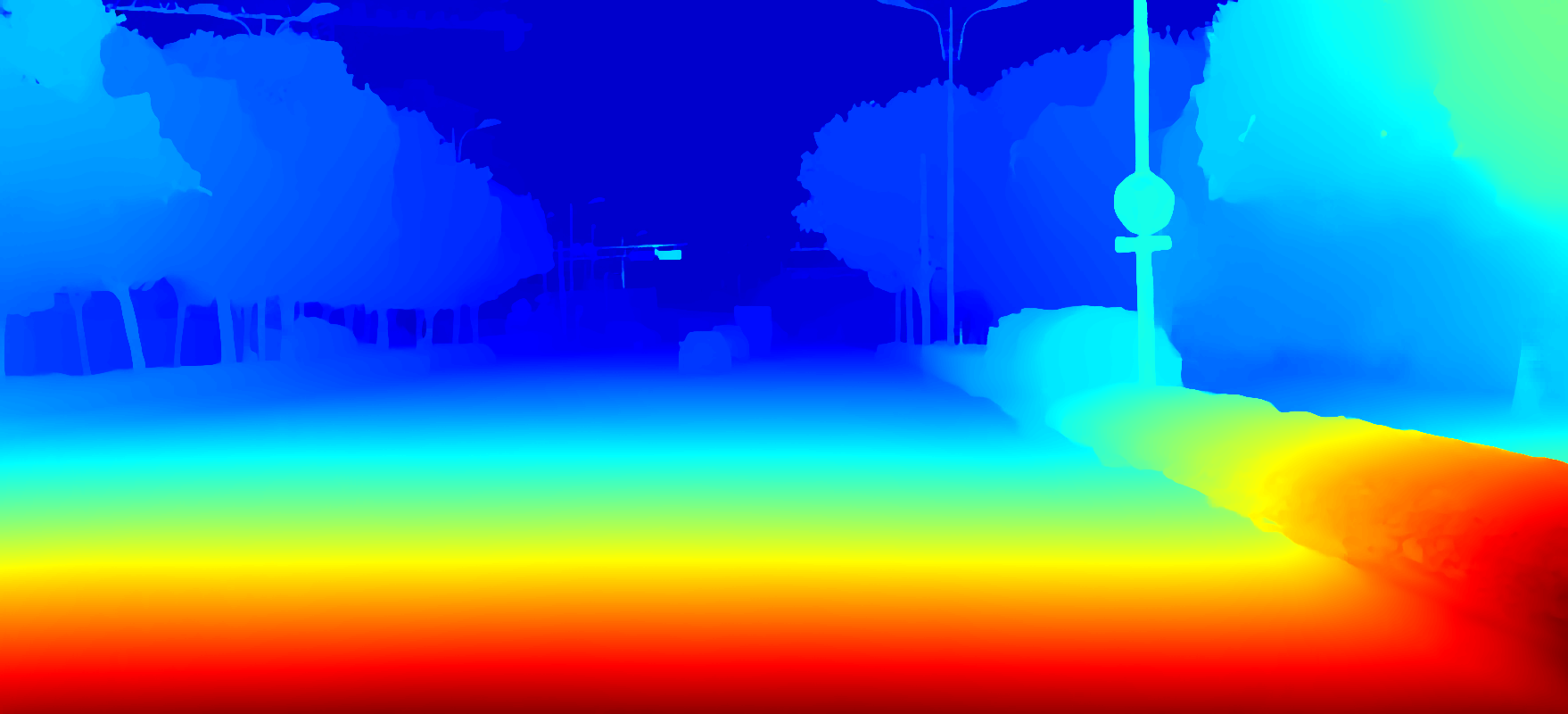}{./all-models-generalization/foggy/debug/06-39-171.png_disp_pred.png}{1239 438 426 108} &
\myimage{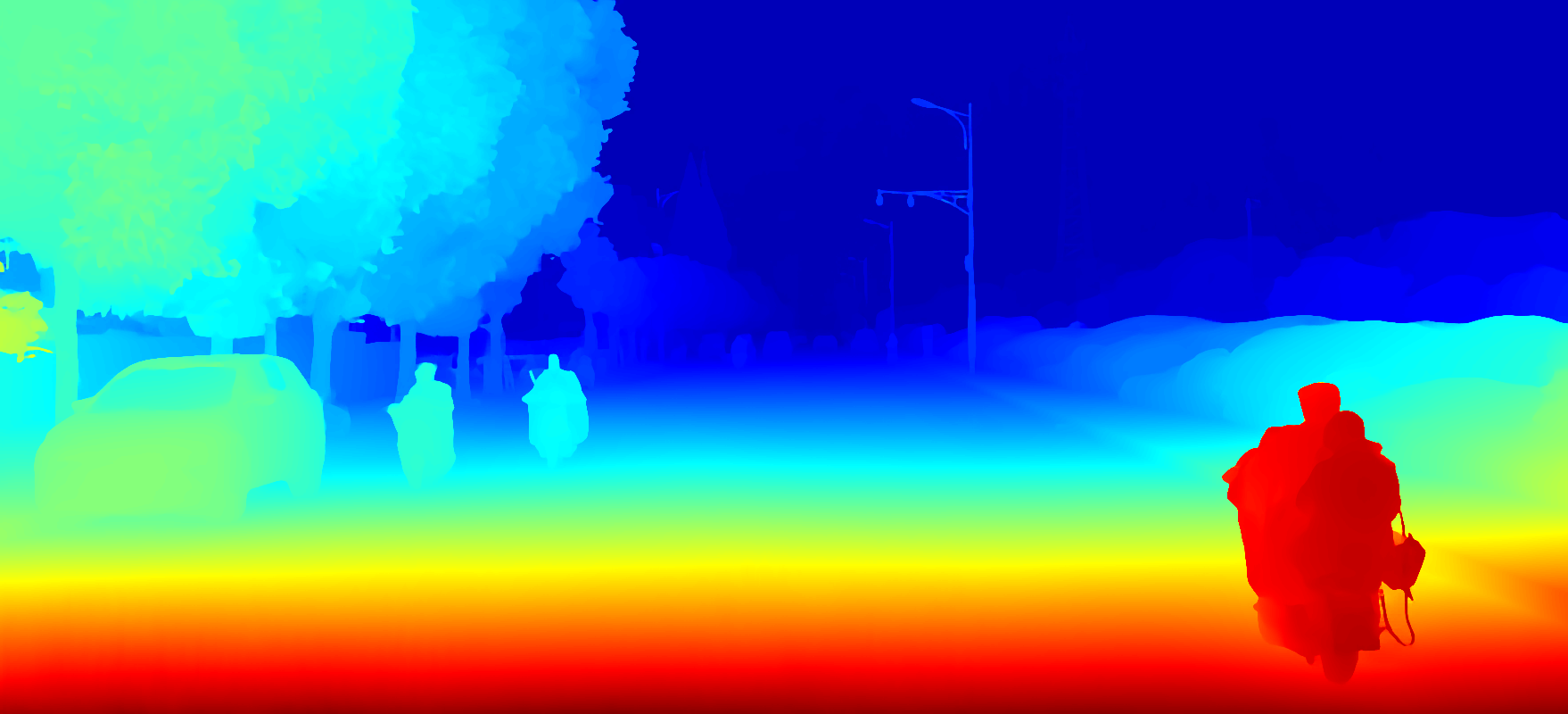}{./all-models-generalization/cloudy/debug/03-15-641.png_disp_pred.png}{1360 35 152 425} &
\myimage{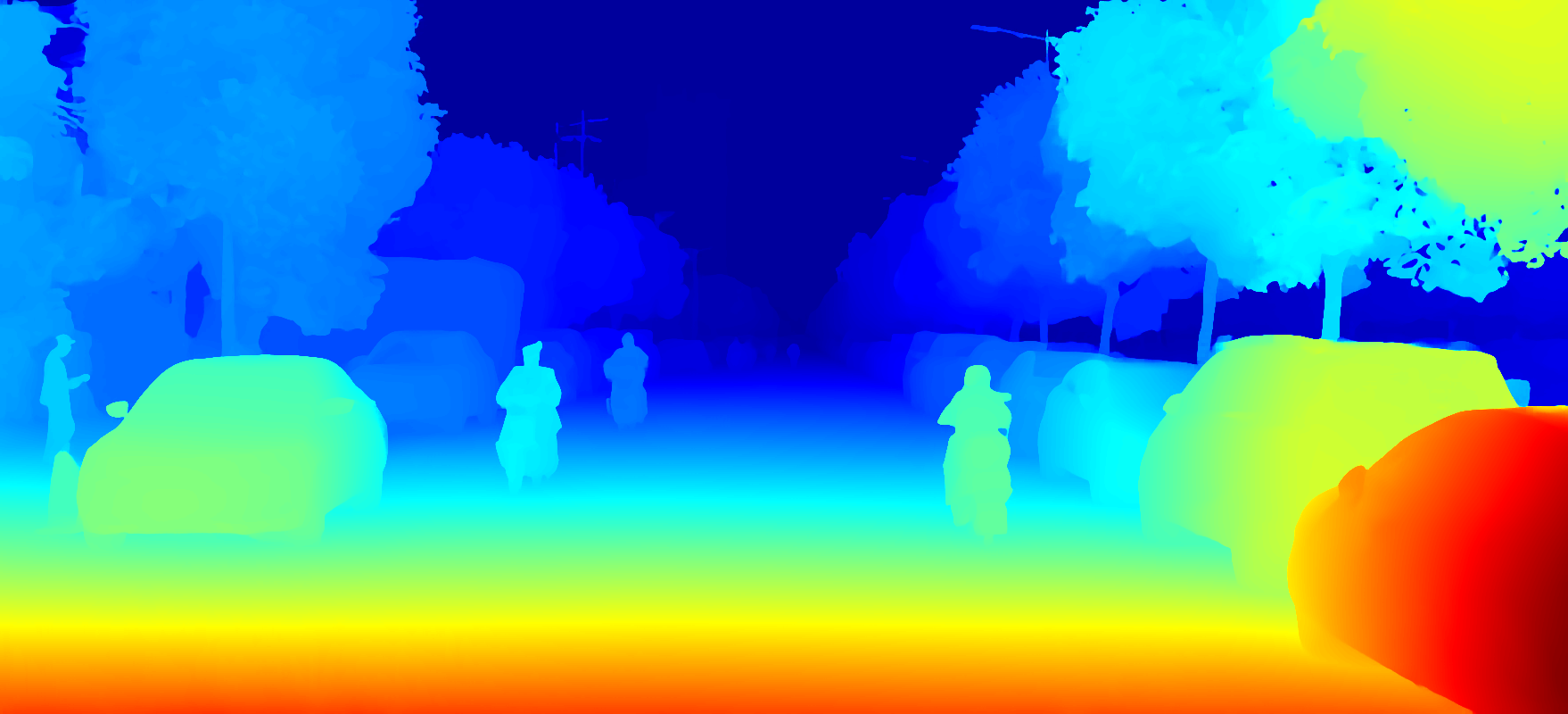}{./all-models-generalization/cloudy/debug/05-39-923.png_disp_pred.png}{34 282 1646 364} \\
\end{tabular}
\caption{Comparison of disparity maps across four different scenes (columns). The rows correspond to (from top to bottom): raw image, RAFT-Stereo, IGEV-Stereo, Selective-IGEV, DEFOM-Stereo, MonSter, MGStereo, and Ours. The bottom-right corner of each image shows a zoomed-in region.}
\label{fig:vis_compara_1}
\end{figure*}

\begin{table*}[htp]
\centering
\caption{Zero-shot generalization performance on multiple benchmarks. All models are trained only on SceneFlow. Best results are in \textbf{bold}, second best are underlined. DS-Sunny, DS-Cloudy, DS-Foggy, and DS-Rainy denote the four weather subsets of the DrivingStereo dataset. The same abbreviations for DrivingStereo subsets are used throughout the following tables.}
\label{tab:zero_comparison}
\small
\setlength{\tabcolsep}{5pt}
\begin{tabular}{lcccccccccccccc}
\toprule
 & \multicolumn{2}{c}{Midd-H} & \multicolumn{2}{c}{DS-Sunny} & \multicolumn{2}{c}{DS-Cloudy} & \multicolumn{2}{c}{DS-Foggy} & \multicolumn{2}{c}{DS-Rainy} & \multicolumn{2}{c}{KITTI15} & \multicolumn{2}{c}{ETH3D} \\
\cmidrule(lr){2-15}
Method & EPE & D2 & EPE & D3 & EPE & D3 & EPE & D3 & EPE & D3 & EPE & D3 & EPE & D1 \\
\midrule
RAFT-Stereo & 1.43 & 11.21 & 1.89 & 9.87 & 1.89 & 10.30 & 1.93 & 13.64 & 4.31 & 29.44 & 1.13 & 5.68 & 0.27 & 2.61 \\
IGEV-Stereo & 2.63 & 11.92 & 2.60 & 10.73 & 3.29 & 12.64 & 2.32 & 16.27 & 6.87 & 33.83 & 1.21 & 6.03 & 0.33 & 4.05 \\
Selective-IGEV & 2.58 & 11.79 & 2.61 & 11.73 & 3.05 & 11.51 & 2.38 & 15.20 & 5.50 & 32.95 & 1.25 & 6.05 & 1.41 & 6.08 \\
DEFOM-Stereo & \textbf{0.84} & \textbf{6.03} & \underline{1.75} & \underline{8.73} & \underline{1.66} & 9.11 & \underline{1.70} & \underline{10.78} & 3.68 & 33.68 & 1.06 & 4.99 & 0.35 & 2.38 \\
Moster & 1.92 & 10.21 & 1.98 & 9.18 & 2.15 & \underline{8.60} & 2.12 & 17.08 & \underline{3.60} & \textbf{20.38} & \textbf{0.89} & \textbf{3.48} & \textbf{0.22} & \textbf{1.34} \\
MGStereo & 1.16 & 8.60 & 1.83 & 9.04 & 2.07 & 9.67 & 3.19 & 14.26 & 3.99 & 28.69 & 1.12 & 5.64 & 0.25 & 1.88 \\
Ours & \underline{1.01} & \underline{7.22} & \textbf{1.61} & \textbf{6.81} & \textbf{1.51} & \textbf{6.63} & \textbf{1.63} & \textbf{9.31} & \textbf{2.50} & \underline{26.77} & \underline{1.01} & \underline{4.62} & \underline{0.23} & \underline{1.81} \\
\bottomrule
\end{tabular}
\end{table*}

\textbf{Generalization Analysis:}
Table \ref{tab:zero_comparison} reports the zero-shot generalization performance on four challenging real-world benchmarks. For fair comparison, all competing methods use their officially released pre-trained models (trained on the SceneFlow dataset), and all models, including ours, are evaluated with 32 GRU iterations. From Table \ref{tab:zero_comparison}, we can observe the following points: 1) Algorithms incorporating the Depth Anything prior significantly outperform those without it, revealing the untapped potential of bridging geometric correspondence with monocular foundation models; 2) Our algorithm achieves the top or second-best performance. Compared to RAFT-Stereo, the EPE is improved by $10.63\%$ to $41.86\%$ across various datasets. The results demonstrate that the generalization capabilities of Depth Anything V2 can be effectively transferred to the RAFT-Stereo model through the designed mechanism proposed in this paper.
Fig. \ref{fig:vis_compara_1} presents the visual comparison of disparity predictions on four examples. Compared to iterative methods that do not use Depth Anything V2, our method achieves more accurate predictions in large textureless areas. Moreover, compared to other iterative methods that employ the large version of Depth Anything V2, our method obtains more accurate predictions on fine‑structured objects, benefiting from the proposed Edge-Aware Loss Suppression strategy.

\begin{table}[htbp]
	\centering
	\setlength{\tabcolsep}{1pt}
	\caption{Computational Complexity Analysis for Different Methods.}
	\label{tab:runtime}
	\begin{tabular}{lccc}
	\toprule
	Model & FLOPs (GMac) & run time(s) & para num (M) \\
	\midrule
	RAFT-Stereo & 188.20 & 0.181 & 11.12 \\
	IGEV-Stereo & 328.82 & 0.261 & 12.60 \\
	Selective-IGEV & 406.20 & 0.219 & 13.14 \\
	DEFOM-Stereo & 843.42 & 0.294 & 47.30 \\
	Moster & 963.92 & 0.369 & 53.38 \\
	MGStereo & 659.65 & 0.215 & 0.68 \\
	Ours & 274.32 & 0.284 & 11.23 \\
	\bottomrule
	\end{tabular}
\end{table}

\begin{figure}[htp]
    \centering
    \includegraphics[trim=10 10 10 10, clip, width=0.95\columnwidth]{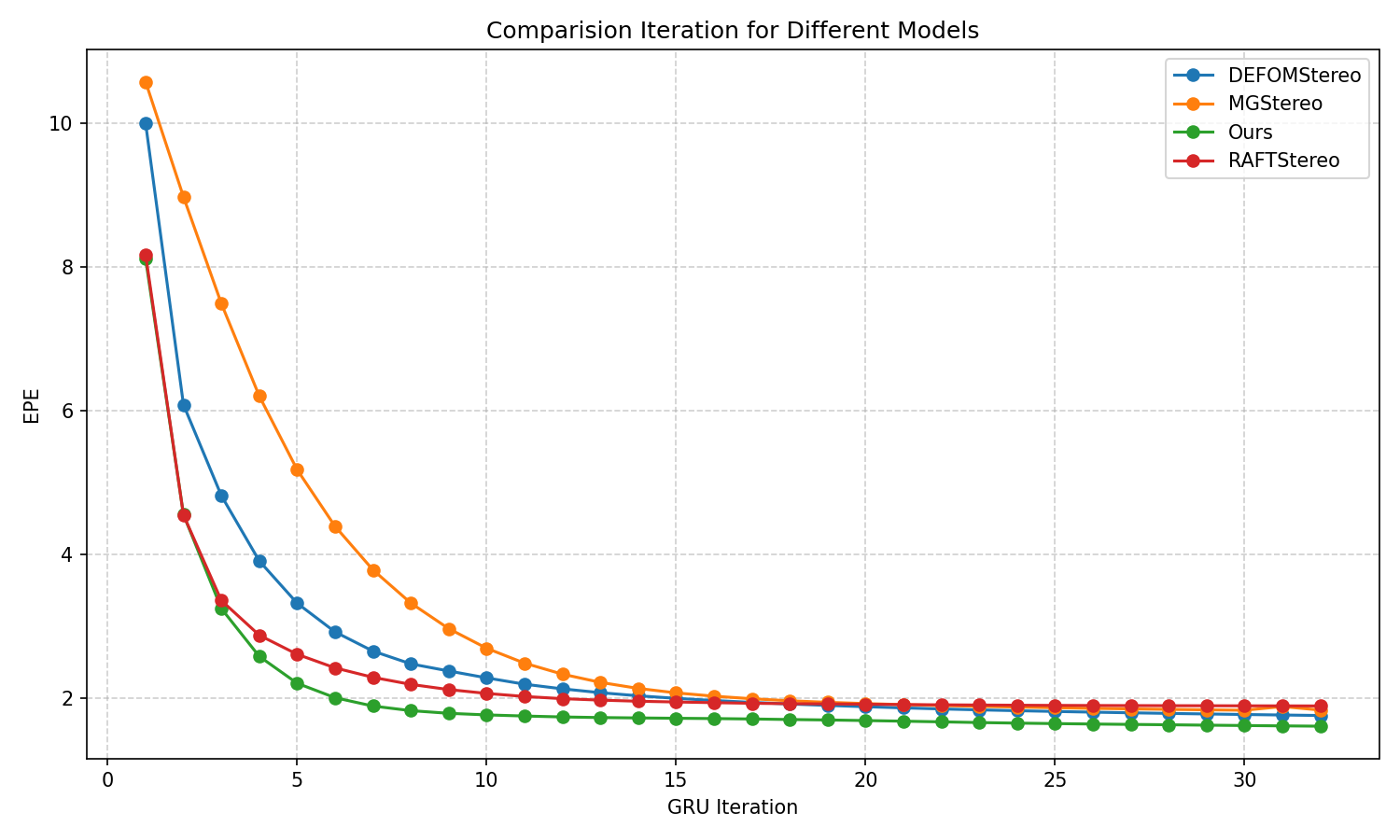}
    \caption{Comparison of EPE performance across GRU iterations for iterative stereo matching models initialized with zero disparity.}
    \label{fig:GRU-Iteration-compare}
\end{figure}

\textbf{Computational Complexity Analysis:}
Table \ref{tab:runtime} lists the runtime, number of trainable parameters, and FLOPs of the seven models. In terms of training parameters, MGStereo requires the fewest, as its feature extraction modules directly leverage the frozen parameters of Depth Anything V2 without retraining. Our method adds only a limited number of extra parameters to RAFT-Stereo. Compared to other algorithms that incorporate monocular priors, our method achieves significant advantages in both FLOPs and parameter count. The models that outperform our method on certain metrics in Table \ref{tab:zero_comparison} all incur longer runtimes and higher FLOPs. This indicates that our method attains greater efficiency while remaining competitive, striking a better balance between performance and computational cost. Furthermore, unlike the other models adopting the large version of Depth Anything V2 ( 335.3M parameters), our method uses only its base version (97.5M parameters), further demonstrating the superiority of the proposed mechanism and its robustness against errors in monocular depth. 
Fig. \ref{fig:GRU-Iteration-compare} shows per-iteration EPE performance comparison of iterative SM models initialized with zero disparity. Our method consistently outperforms the other three after the initial iterations. Reaching comparable performance by the 7th GRU iteration versus over 20 for DEFOM-Stereo, MGStereo, and RAFT-Stereo. It demonstrates stronger GRU convergence stability in our method.

\begin{figure*}[htp]
\centering
\setlength{\tabcolsep}{0.03cm}
\newlength{\imageheight}
\setlength{\imageheight}{2.7cm}

\begin{tabular}{cccc}

\includegraphics[height=\imageheight, trim=1579 340 20 337, clip]{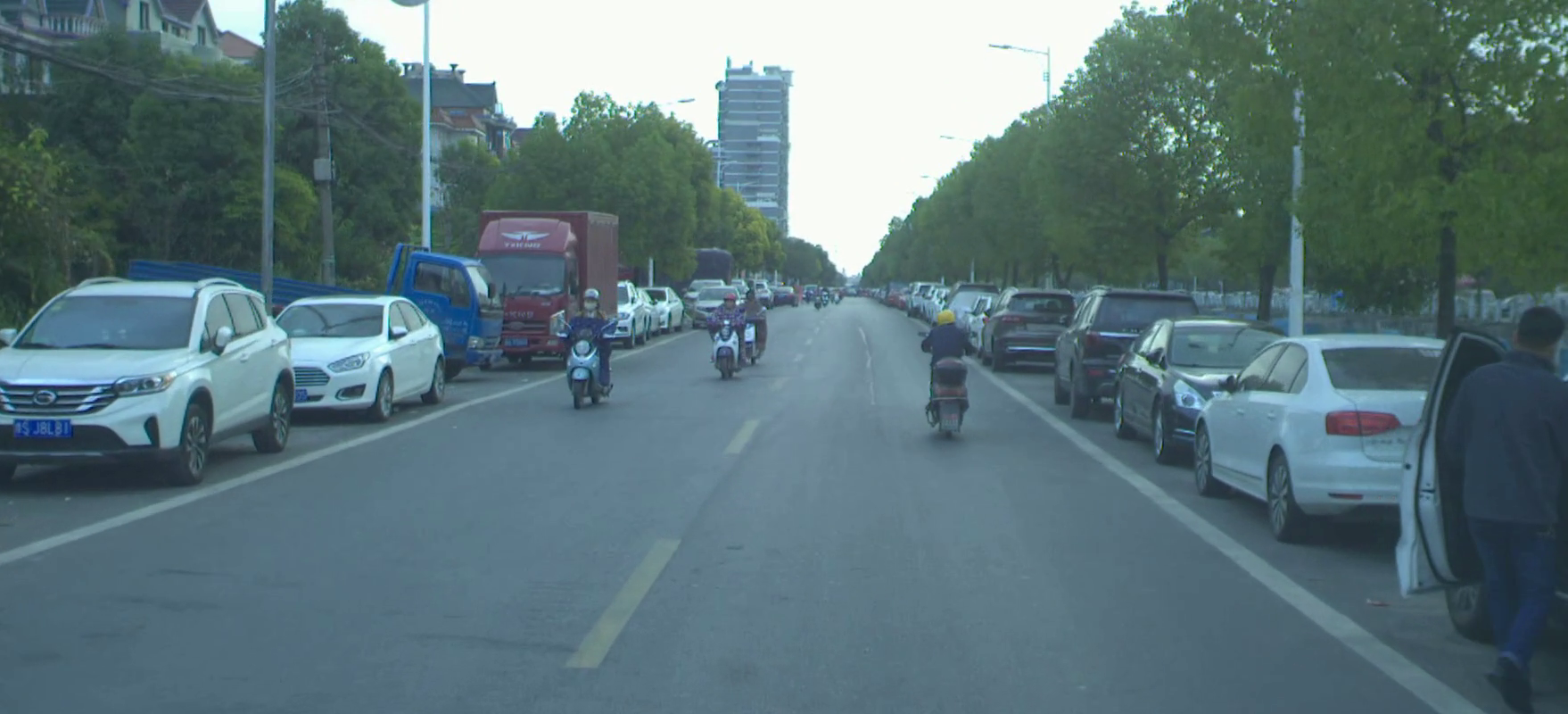} &
\includegraphics[height=\imageheight, trim=1579 340 20 337, clip]{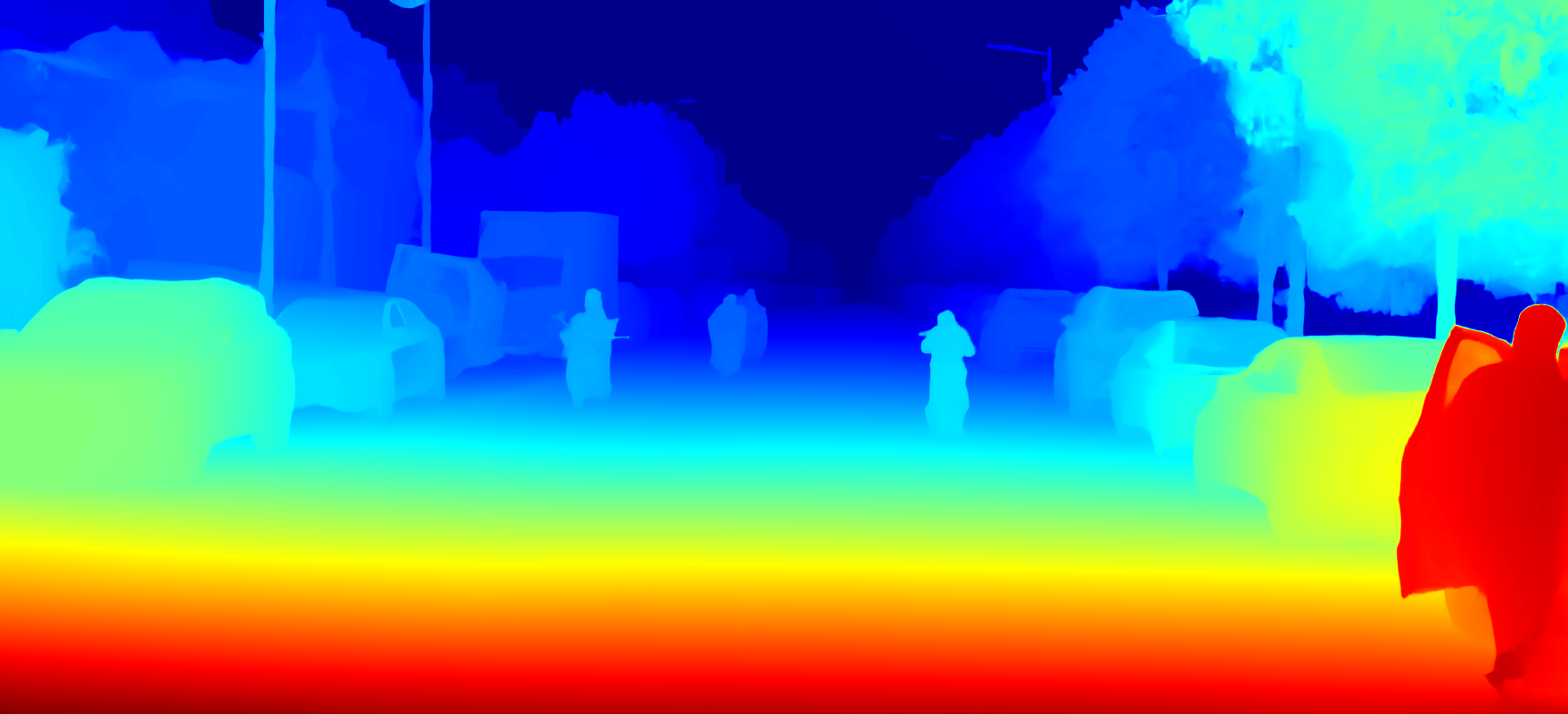} &
\includegraphics[height=\imageheight, trim=1579 340 20 337, clip]{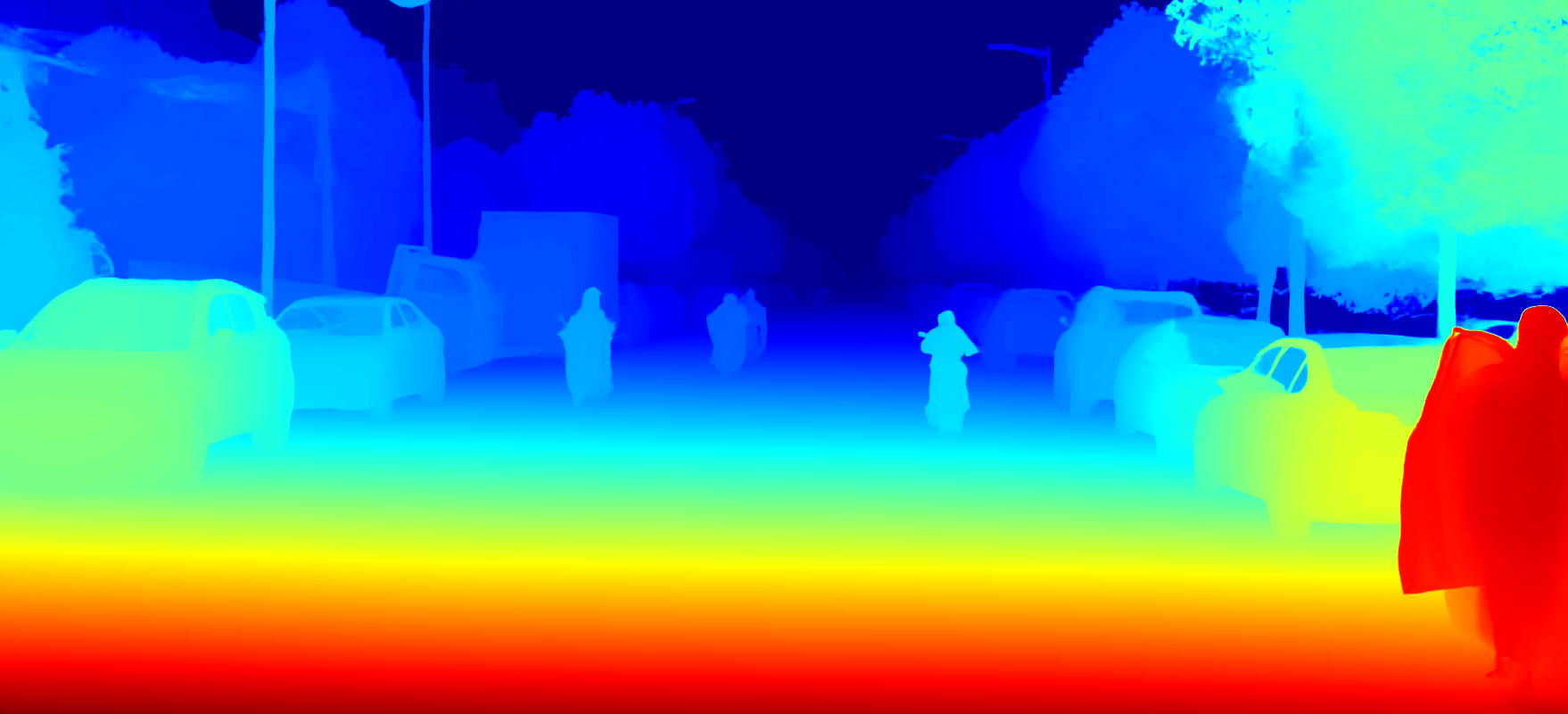} &
\includegraphics[height=\imageheight, trim=1579 340 20 337, clip]{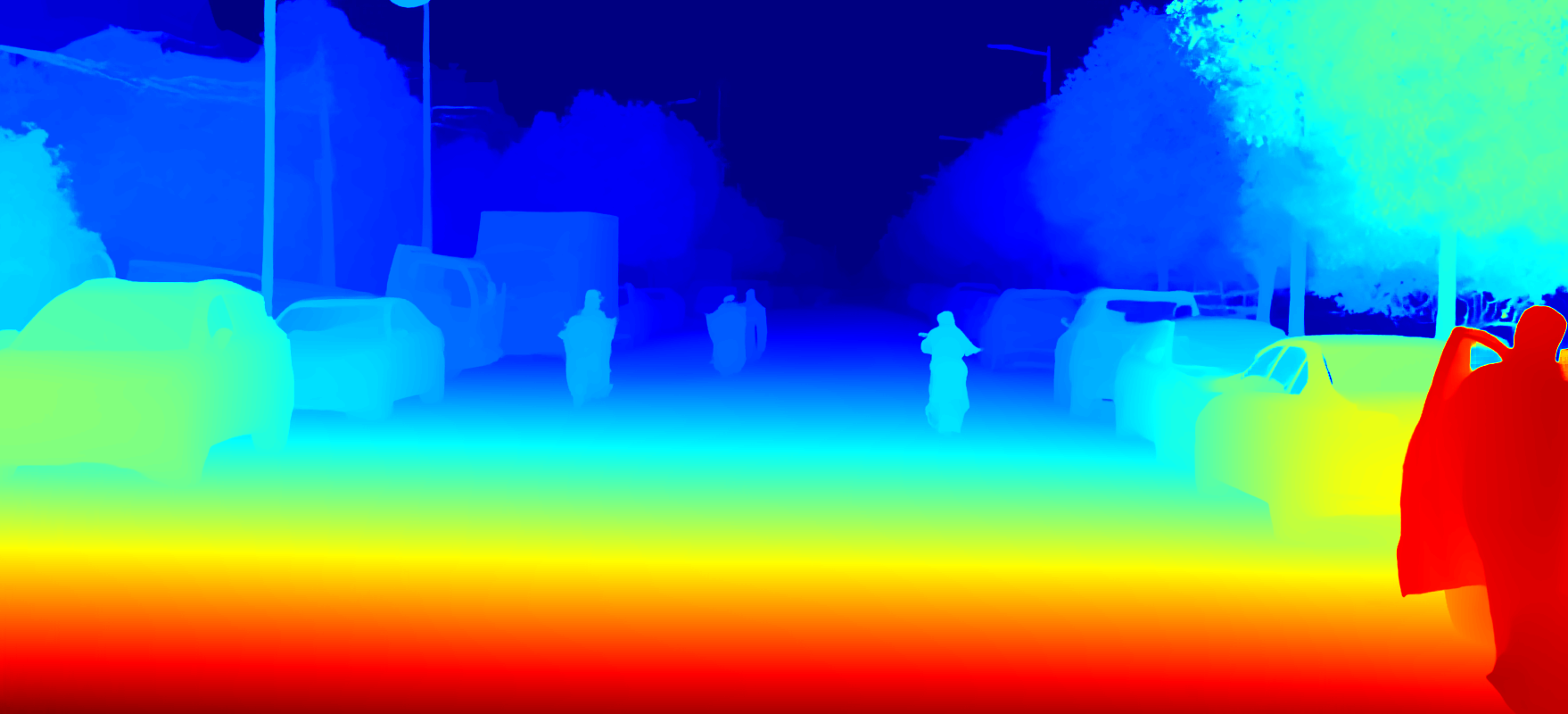} \\
\parbox{3cm}{\centering Raw Image} & \parbox{3cm}{\centering RAFT-Stereo} & \parbox{3cm}{\centering IGEV-Stereo} & \parbox{3cm}{\centering Selective-IGEV} \\

\includegraphics[height=\imageheight, trim=1579 340 20 337, clip]{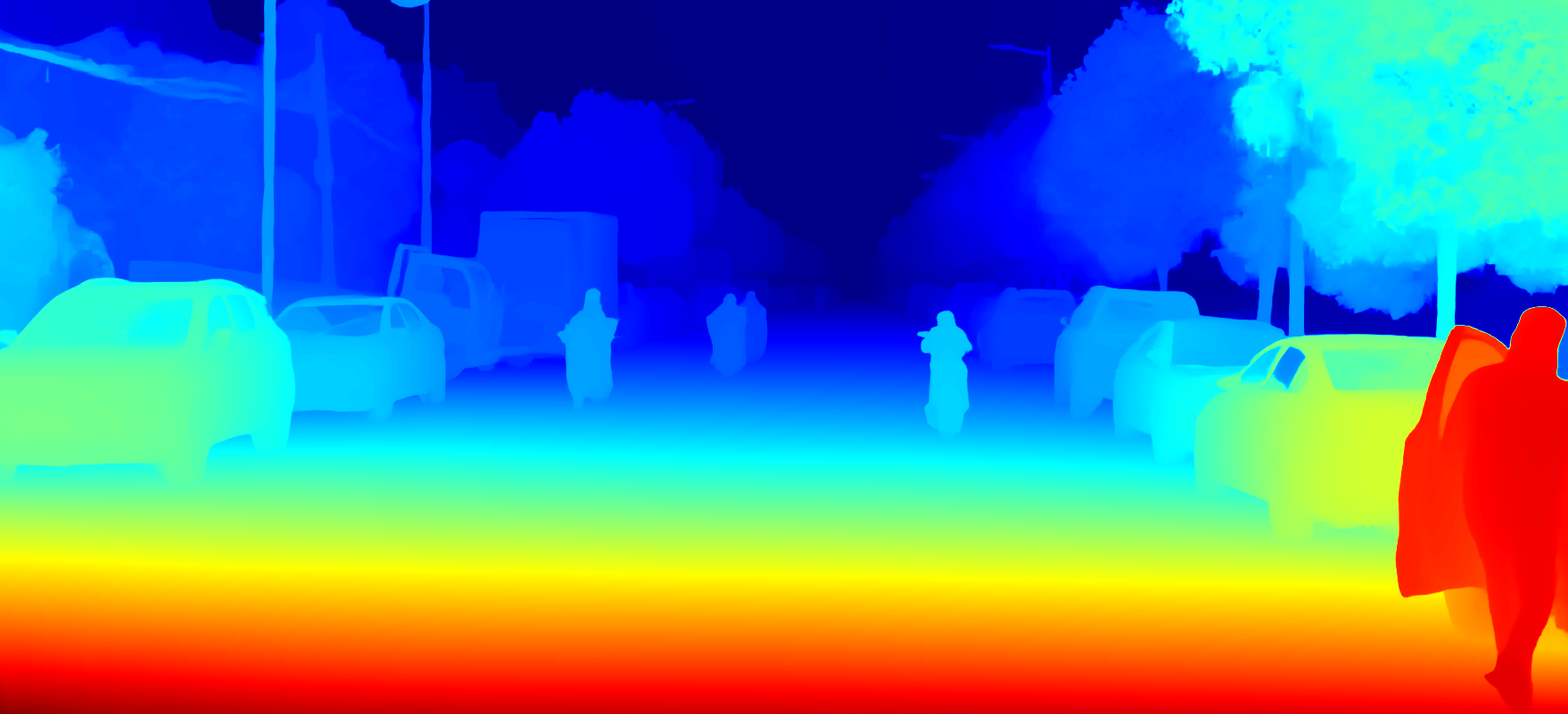} &
\includegraphics[height=\imageheight, trim=1579 340 20 337, clip]{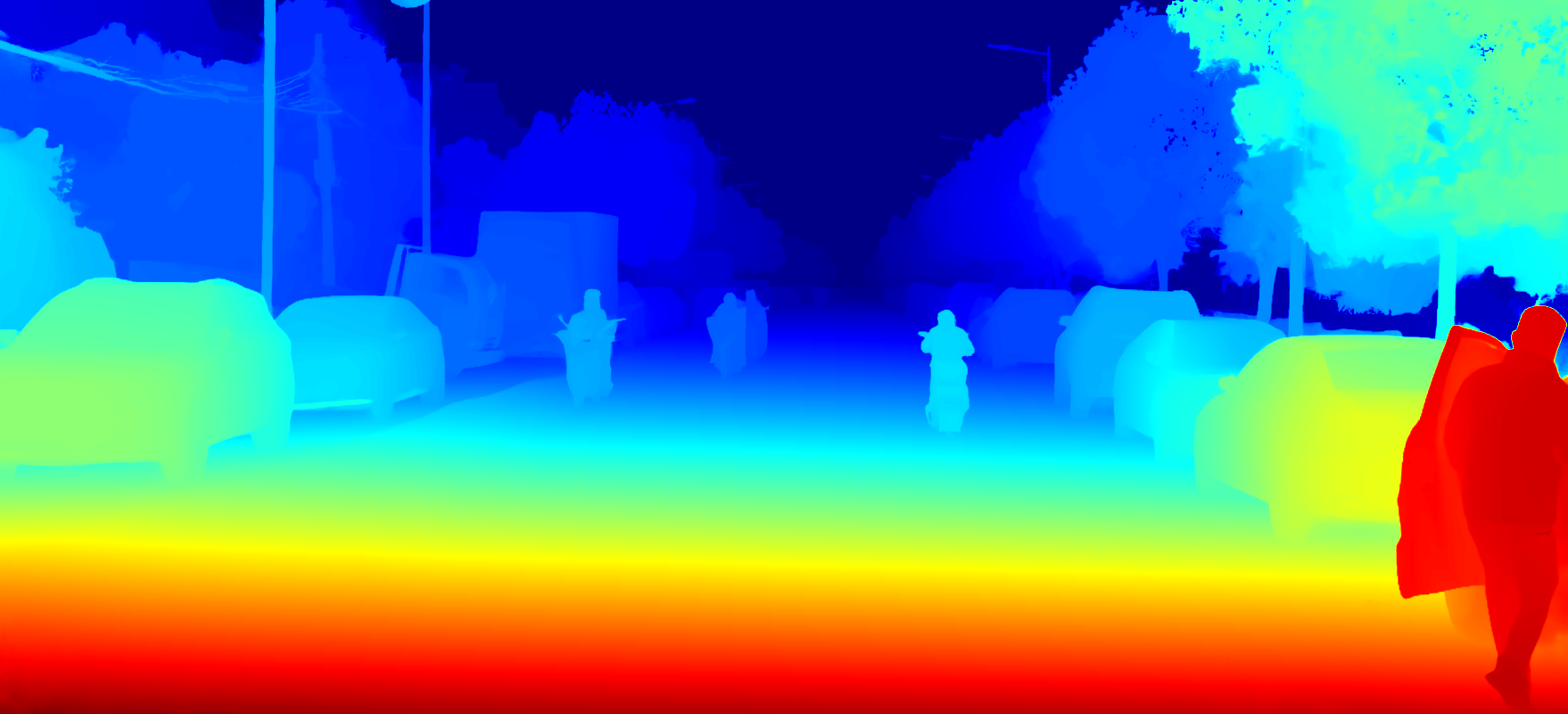} &
\includegraphics[height=\imageheight, trim=1579 340 20 337, clip]{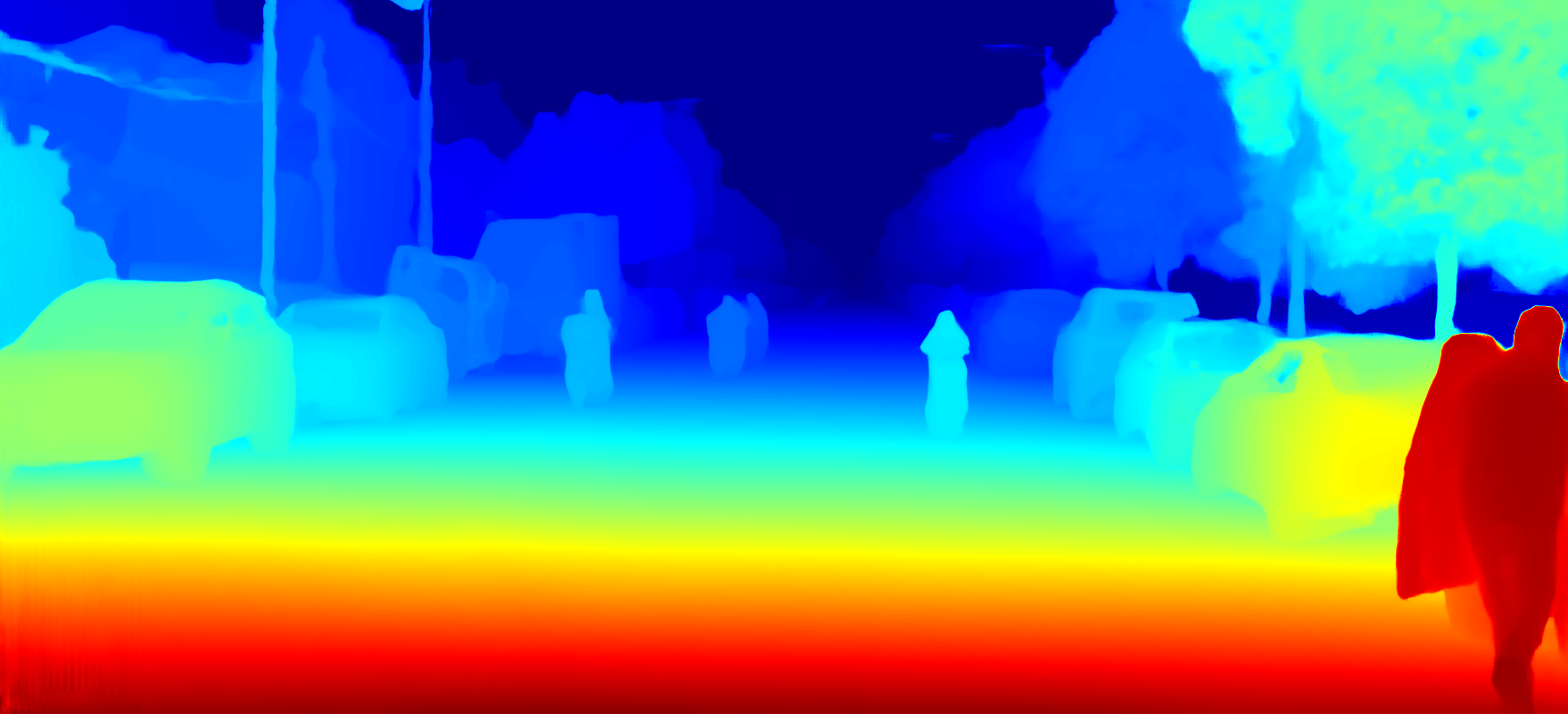} &
\includegraphics[height=\imageheight, trim=1579 340 20 337, clip]{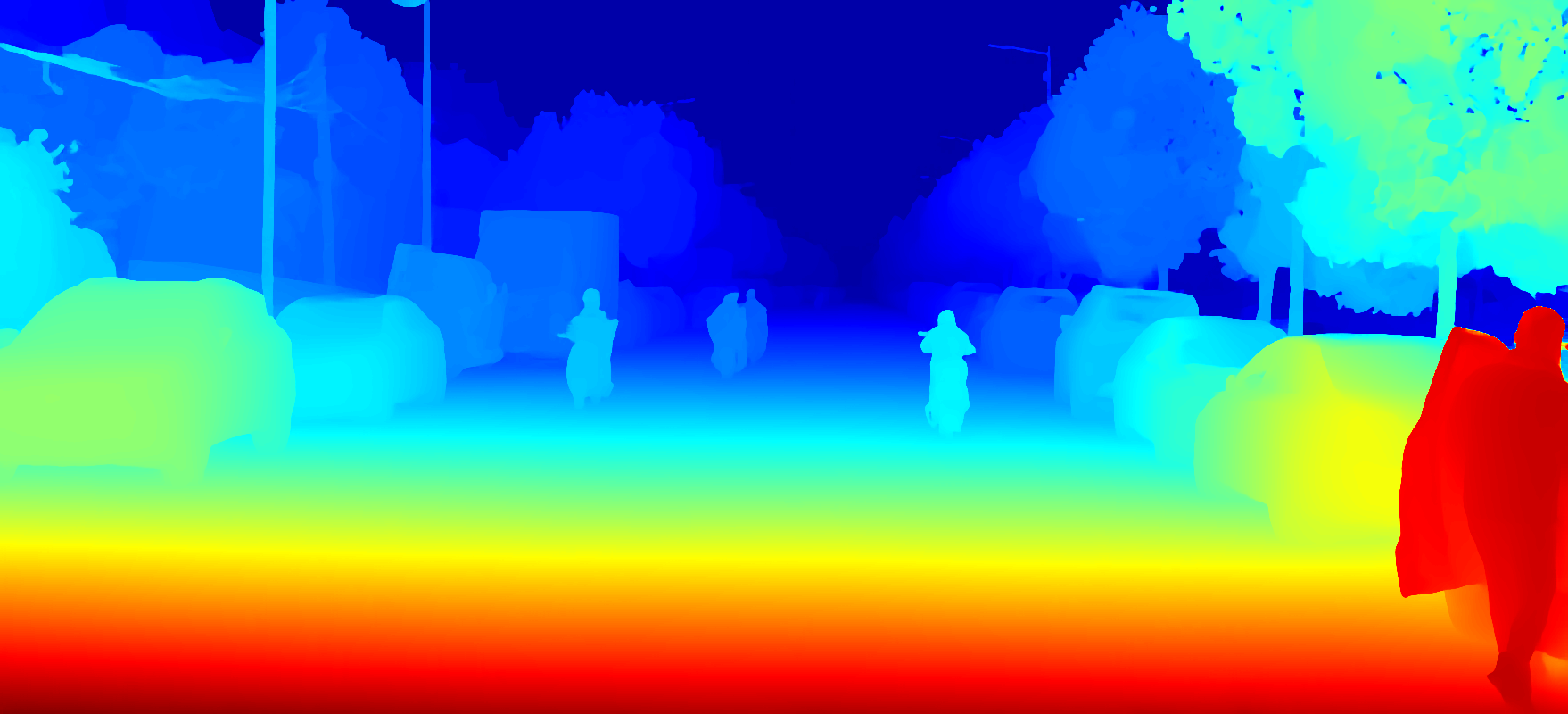} \\
\parbox{3cm}{\centering DEFOM-Stereo} & \parbox{3cm}{\centering MonSter} & \parbox{3cm}{\centering MGStereo} & \parbox{3cm}{\centering Ours} \\

\end{tabular}

\caption{Disparity edge comparison with all images cropped to the same region. Our algorithm can accurately locates disparity boundaries.}
\label{fig:vis_compara_2}
\end{figure*}

\textbf{Disparity Edge Prediction:}
Fig. \ref{fig:vis_compara_2} illustrates an expected behavior of our Edge‑Aware Loss Suppression (EALS) module. By masking the supervision on edge regions during training, the model cannot directly learn disparity from edges; instead, it is encouraged to generalize disparity knowledge learned from non‑edge areas to the edges. The results in Fig. \ref{fig:vis_compara_2} confirm that this expectation is met. Although the predicted object shapes are not completely smooth, the edge artifacts in the disparity maps produced by our method are significantly reduced compared to other models.

These results validate that the proposed depth-guided Image augmentation, gradient-guided GRU refinement, and  edge-aware loss suppression modules collectively improve the generalization and accuracy of iterative SM, particularly in challenging regions.

\subsection{Ablation Studies}

We conduct four ablation studies. The first evaluates the contribution of the three proposed core components: DGIA, gradient-GRU, and EALS. The remaining three perform ablation analyses on the internal architecture design of each component.

\begin{table*}[htp]
	\centering
	\caption{Ablation study on DrivingStereo (sunny, cloudy, foggy, rainy). Best results are in \textbf{bold}, second best are underlined.}
	\label{tab:ablation_1}
	\small
	\setlength{\tabcolsep}{9pt}
	\begin{tabular}{lcccccccc}
	\toprule
	 & \multicolumn{2}{c}{DS-Sunny} & \multicolumn{2}{c}{DS-Cloudy} & \multicolumn{2}{c}{DS-Foggy} & \multicolumn{2}{c}{DS-Rainy} \\
	\cmidrule(lr){2-9}
	Method & EPE & D3 & EPE & D3 & EPE & D3 & EPE & D3 \\
	\midrule
	Baseline & 1.81 & 8.34 & 1.68 & 8.85 & 1.83 & 12.73 & 3.79 & 32.36 \\
	Baseline+DGIA & 1.82 & 8.58 & 1.64 & 8.25 & 1.74 & 11.18 & 2.79 & \underline{28.37} \\
	Baseline+DGIA+gradient-GRU & \underline{1.72} & \underline{7.96} & \underline{1.57} & \underline{7.91} & \underline{1.69} & \underline{10.77} & \underline{2.61} & 29.13 \\
	Baseline+DGIA+gradient-GRU+EALS & \textbf{1.60} & \textbf{6.81} & \textbf{1.51} & \textbf{6.63} & \textbf{1.63} & \textbf{9.31} & \textbf{2.50} & \textbf{26.76} \\
	\bottomrule
	\end{tabular}
\end{table*}

\begin{table*}[htp]
	\centering
	\caption{Ablation study on Depth-Guided Image Augmentation module. Best results are in \textbf{bold}, second best are underlined.}
	\label{tab:ablation_2}
	\small
	\setlength{\tabcolsep}{5pt}
	\begin{tabular}{lcccccccccccccc}
	\toprule
	 & \multicolumn{2}{c}{Midd-H} & \multicolumn{2}{c}{DS-Sunny} & \multicolumn{2}{c}{DS-Cloudy} & \multicolumn{2}{c}{DS-Foggy} & \multicolumn{2}{c}{DS-Rainy} & \multicolumn{2}{c}{KITTI15} & \multicolumn{2}{c}{ETH3D} \\
	\cmidrule(lr){2-15}
	Method & EPE & D2 & EPE & D3 & EPE & D3 & EPE & D3 & EPE & D3 & EPE & D3 & EPE & D1 \\
	\midrule
	Depth+CNN+both & \underline{1.47} & \underline{9.35} & \textbf{1.68} & \textbf{8.32} & \textbf{1.63} & 9.05 & \underline{1.76} & 12.37 & \underline{2.76} & 29.70 & \underline{1.05} & \underline{5.24} & \underline{0.22} & 2.41 \\
	Depth+DCN+only & 1.53 & 9.49 & 1.85 & \underline{8.57} & 1.67 & \underline{8.76} & \underline{1.76} & \underline{12.14} & \textbf{2.68} & \underline{29.31} & 1.11 & 5.57 & \underline{0.22} & \underline{1.81} \\
	Depth+DCN+both & \textbf{0.99} & \textbf{7.55} & \underline{1.82} & 8.58 & \underline{1.64} & \textbf{8.25} & \textbf{1.74} & \textbf{11.18} & 2.79 & \textbf{28.37} & \textbf{1.02} & \textbf{4.87} & \textbf{0.21} & \textbf{1.75} \\
	\bottomrule
	\end{tabular}
\end{table*}

\begin{table*}[htp]
	\centering
	\caption{Ablation study on Gradient-Guided GRU Refinement module. Best results are in \textbf{bold}, second best are underlined.}
	\label{tab:ablation_3}
	\small
	\setlength{\tabcolsep}{5pt}
	\begin{tabular}{lcccccccccccccc}
	\toprule
	 & \multicolumn{2}{c}{Midd-H} & \multicolumn{2}{c}{DS-Sunny} & \multicolumn{2}{c}{DS-Cloudy} & \multicolumn{2}{c}{DS-Foggy} & \multicolumn{2}{c}{DS-Rainy} & \multicolumn{2}{c}{KITTI15} & \multicolumn{2}{c}{ETH3D} \\
	\cmidrule(lr){2-15}
	Method & EPE & D2 & EPE & D3 & EPE & D3 & EPE & D3 & EPE & D3 & EPE & D3 & EPE & D1 \\
	\midrule
	Depth-Guided & \underline{1.15} & \underline{8.28} & \underline{1.80} & \underline{8.35} & \underline{1.60} & \underline{8.29} & \underline{1.72} & \underline{11.32} & \underline{2.72} & \textbf{29.13} & \textbf{1.03} & \textbf{5.01} & \textbf{0.21} & \textbf{1.95} \\
	Gradient-Guided & \textbf{1.07} & \textbf{8.19} & \textbf{1.72} & \textbf{7.96} & \textbf{1.57} & \textbf{7.91} & \textbf{1.69} & \textbf{10.77} & \textbf{2.61} & \textbf{29.13} & \textbf{1.03} & \underline{5.15} & \textbf{0.21} & \underline{1.97} \\
	\bottomrule
	\end{tabular}
\end{table*}

\begin{table*}[htp]
\centering
\caption{Ablation study on Edge-Aware Loss Suppression module. Best results are in \textbf{bold}, second best are underlined.}
\label{tab:ablation_4}
\small
\setlength{\tabcolsep}{7pt}
\begin{tabular}{lcccccccccccccc}
\toprule
 & \multicolumn{2}{c}{Midd-H} & \multicolumn{2}{c}{DS-Sunny} & \multicolumn{2}{c}{DS-Cloudy} & \multicolumn{2}{c}{DS-Foggy} & \multicolumn{2}{c}{DS-Rainy} & \multicolumn{2}{c}{KITTI15} & \multicolumn{2}{c}{ETH3D} \\
\cmidrule(lr){2-15}
Method & EPE & D1 & EPE & D1 & EPE & D1 & EPE & D1 & EPE & D1 & EPE & D1 & EPE & D1 \\
\midrule
$\tau = 0.8$ & 1.61 & 9.72 & 1.63 & \underline{6.90} & \underline{1.53} & \underline{6.96} & \underline{1.66} & \underline{9.82} & \underline{2.56} & \underline{27.34} & 1.03 & 4.83 & \textbf{0.22} & 1.91 \\
$\tau = 0.9$ & \textbf{1.00} & \textbf{7.21} & \textbf{1.60} & \textbf{6.81} & \textbf{1.51} & \textbf{6.63} & \textbf{1.63} & \textbf{9.31} & \textbf{2.50} & \textbf{26.76} & \textbf{1.00} & \textbf{4.61} & \textbf{0.22} & \textbf{1.81} \\
$\tau = 1$ & \underline{1.09} & \underline{8.66} & \underline{1.62} & 6.96 & 1.55 & 7.44 & 1.69 & 10.78 & 2.63 & 28.76 & \underline{1.01} & \underline{4.65} & \textbf{0.22} & \underline{1.86} \\
\bottomrule
\end{tabular}
\end{table*}

\textbf{Analysis of DGIA, gradient-GRU, and EALS:}
As shown in the Table \ref{tab:ablation_1}, the DGIA module significantly improves the model's robustness under adverse weather conditions, incurring only a minor performance drop on sunny scenes, thereby enhancing overall generalization. The gradient-GRU module improves performance under sunny, foggy and cloudy conditions but degrades under rainy conditions, indicating that it is sensitive to degradation types and mainly benefits contrast‑reducing degradations. In contrast, the EALS module achieves consistent performance gains across all four weather conditions, demonstrating its robustness and generalizability to various types of weather degradation. Overall, these results indicate that while DGIA and gradient-GRU offer targeted improvements, EALS provides the most reliable and universal enhancement across diverse weather scenarios.

\textbf{Internal Architecture Analysis of DGIA:} Table \ref{tab:ablation_2} compares three fusion strategies in DGIA module: “Depth+\allowbreak CNN+\allowbreak both” fuses the depth map and RGB image using standard convolution for both feature and context extraction. “Depth+\allowbreak DCN+\allowbreak only” uses deformable convolution (DCN), applying the enhanced RGB image only to context extraction, while the original RGB is used for feature extraction. “Depth+\allowbreak DCN+\allowbreak both” employs DCN and uses the enhanced RGB image for both modules. Experimental results indicate that using DCN is crucial for fusing depth maps with RGB images.  Furthermore, both matching feature extraction and context extraction benefit from depth enhanced RGB images.

\begin{figure}[htp]
\begin{flushleft}
\begin{minipage}{0.48\columnwidth}
    \includegraphics[width=\linewidth]{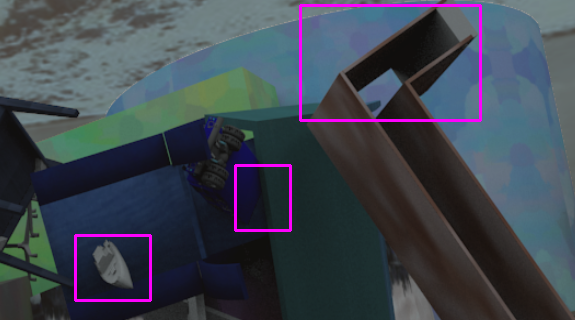}
    \medskip
    \small (a) RGB Image
\end{minipage}
\vspace{0.3\baselineskip}
\begin{minipage}{0.48\columnwidth}
    \includegraphics[width=\linewidth]{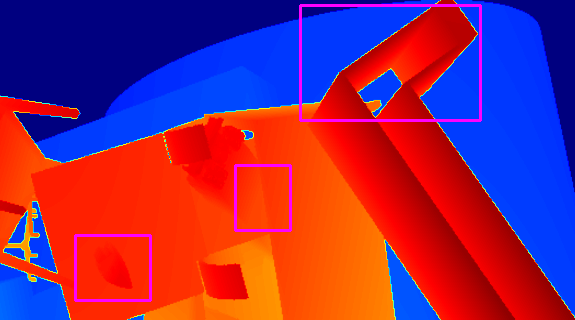}
    \medskip
    \small (b) GT disparity
\end{minipage}
\hspace{0.02\columnwidth}
\begin{minipage}{0.48\columnwidth}
    \includegraphics[width=\linewidth]{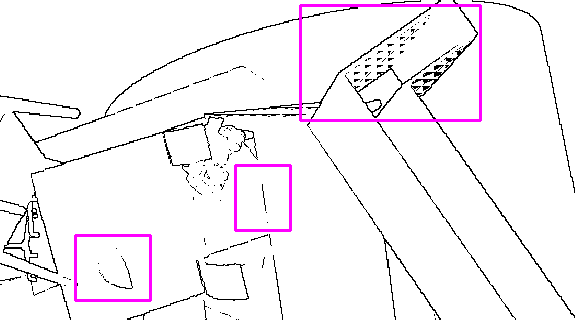}
    \medskip
    \small (c) \(\tau=0.8\)
\end{minipage}
\vspace{0.3\baselineskip}
\begin{minipage}{0.48\columnwidth}
    \includegraphics[width=\linewidth]{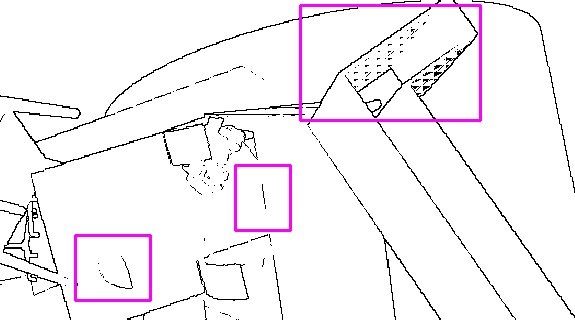}
    \medskip
    \small (d) \(\tau=0.9\)
\end{minipage}
\hspace{0.02\columnwidth}
\begin{minipage}{0.48\columnwidth}
    \includegraphics[width=\linewidth]{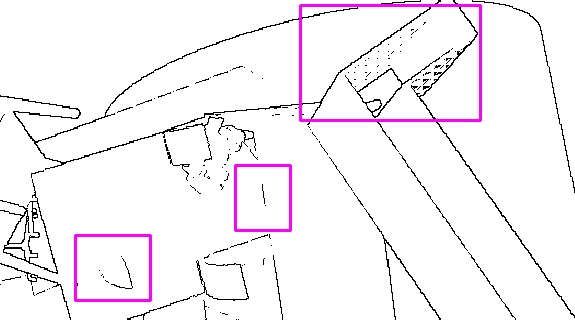}
    \medskip
    \small (e) \(\tau=1.0\)
\end{minipage}
\end{flushleft}
\caption{Visual comparison of masks with different thresholds in the EALS module. When $\tau$ is set too low, more pixels in non-edge regions are mistakenly treated as edges and their disparity supervision is masked; when $\tau$ is set too high, the suppression effect on edge pixels weakens, leading to insufficient masking of edge supervision.}
\label{fig:edge-mask}
\end{figure}

\textbf{Internal Architecture Analysis of gradient-GRU:} Table  \ref{tab:ablation_3} compares two strategies: “Depth‑Guided” uses the depth map as additional context to guide the GRU updates, while “Gradient‑Guided” replaces the depth map with its gradient map. The results show that gradient guidance consistently outperforms direct depth guidance. We hypothesize that the GRU implicitly acquires depth information from the cost volume and initial hidden state (derived from DGIA's depth-enhanced images). Thus, an additional depth map is redundant, while the gradient map provides complementary structural cues absent in the GRU's context.

\textbf{Analysis of EALS:} In the design of the EALS module, we adopt a hard‑mask strategy based on the raw edge mask \({\Delta \mathbf{D}} < \tau\) ; the results for different \(\tau\) are shown in Table \ref{tab:ablation_4}. A large threshold \(\tau\) makes the mask too permissive, missing edges with small disparity differences, while a small \(\tau\) makes it overly strict, suppressing supervision in slanted internal object regions and hindering contextual learning for accurate disparity prediction (see Fig. \ref{fig:edge-mask}). Based on experiments, \(\tau\) = 0.9 was selected as the optimal setting.


\section{Conclusion}\label{sec:concl}

In this paper, to fully exploit the monocular prior knowledge from Depth Anything V2 and enhance the generalization capability of stereo matching, we propose three complementary modules: DGIA, gradient-GRU, and EALS. The DGIA module fuses the monocular depth map with the RGB image via deformable convolution, injecting geometric priors into both matching and context features, thereby significantly improving robustness in ill-posed regions such as texture-less and occluded areas. The gradient-GRU module converts the depth map into a gradient map, providing complementary structural information for the GRU iterations and helping the model escape local oscillations. The EALS module constructs a binary edge mask, effectively suppressing unreliable edge supervision caused by random scaling in data augmentation and reducing training interference. Together, these three modules form a coherent framework that bridges monocular and stereo depth representations.

Extensive zero-shot generalization experiments on KITTI 2015, Middlebury, ETH3D, and DrivingStereo demonstrate that our method achieves state-of-the-art performance, validating the synergistic effectiveness of the three proposed modules. Moreover, compared to other state-of-the-art models, our approach maintains a lightweight architecture with only 11.23M parameters and competitive computational cost, making it practical for real-world deployment.

For future work, we plan to explore adaptive soft weights for edge disparity supervision to better preserve fine-grained structures while suppressing noise. We also intend to extend our method to more challenging cross-domain scenarios, including indoor and outdoor environments with significant domain gaps, and investigate the integration of stronger monocular foundation models to further boost generalization performance.

\sloppy
\bibliographystyle{elsarticle-num} 
\bibliography{refs}



\end{document}